\documentclass{article}

\usepackage{iclr2027_conference,times}
\iclrfinalcopy

\usepackage[utf8]{inputenc}
\usepackage[T1]{fontenc}
\usepackage{kotex}
\usepackage{url}
\usepackage{microtype}
\usepackage{xcolor}
\usepackage{nicefrac}

\usepackage{amsmath,amssymb,amsfonts}
\usepackage{amsthm}
\usepackage{enumerate}
\usepackage{graphicx}
\usepackage{booktabs}
\usepackage{multirow,hhline}
\usepackage{algorithm}
\usepackage{algpseudocode}
\usepackage{tabularx}
\usepackage{tabulary}
\usepackage{caption}
\usepackage{subcaption}
\usepackage{framed}

\usepackage{hyperref}
\hypersetup{breaklinks=true,colorlinks=true,linkcolor=blue,citecolor=blue,urlcolor=blue}
\usepackage[noabbrev,capitalise,nameinlink]{cleveref}

\newtheorem{theorem}{Theorem}

\newtheorem{assumption}{Assumption}

\newtheorem{lemma}{Lemma}
\newtheorem*{unnumberedlemma}{Lemma}
\newtheorem{definition}{Definition}

\DeclareMathOperator{\argmax}{arg\,max}

\title{A Smooth Polynomial Lyapunov Certificate for Convergence of Q-Learning and Its Smooth Variants}

\author{%
Donghwan Lee \quad Hyunjun Na\\
Department of Electrical Engineering\\
Korea Advanced Institute of Science and Technology\\
Daejeon, 34141, South Korea\\
\texttt{donghwan@kaist.ac.kr}
}

\begin{document}
\maketitle

\begin{abstract}
Classical convergence analyses of Q-learning rely on the $\infty$-norm contraction of Bellman operators, and existing ordinary differential equation (ODE) arguments often use the non-differentiable $\infty$-norm directly. This paper develops a smooth polynomial Lyapunov-function-based stability certificate for convergence of Q-learning by transferring $\infty$-norm contraction to a weighted degree-$2p$ polynomial Lyapunov function induced by a finite $2p$-norm. The framework is conceptual and structural: it avoids non-differentiability, handles preconditioned dynamics arising in Q-learning and its variants, and gives a unified stability argument for standard Q-learning and smooth variants based on log-sum-exp (LSE), mellowmax, and Boltzmann softmax operators. For contractive operators, including the max, LSE, and mellowmax cases, the associated ODEs are globally exponentially stable and, under the stated independent and identically distributed (i.i.d.) sampling model, the stochastic approximation iterates converge almost surely. For the Boltzmann operator, which need not be contractive, the same framework yields convergence to an explicit invariant error set around the optimal Q-function. The resulting theory is not intended as a finite-time bound, but as a clean ODE foundation that unifies and simplifies asymptotic analyses of Q-learning and its smooth variants.
\end{abstract}


\section{Introduction}

Reinforcement learning (RL) provides a mathematical framework for sequential decision-making in unknown environments, where an agent learns policies by interacting with the environment and observing rewards~\citep{sutton1998reinforcement}. Its recent empirical successes, including human-level or superhuman performance in challenging domains~\citep{mnih2015human}, have intensified interest in understanding the stability and convergence properties of RL algorithms. Among these algorithms, Q-learning~\citep{watkins1992q} remains one of the most fundamental: it is simple, model-free, and directly targets the optimal action-value function. Its asymptotic convergence has been studied extensively~\citep{jaakkola1994convergence,borkar2000ode,lee2020unified,tsitsiklis1994asynchronous,singh2000convergence}, and more recent work has developed finite-time guarantees that quantify the sample complexity of convergence~\citep{szepesvari1998asymptotic,even2003learning,beck2012error,wainwright2019stochastic,qu2020finite,li2020sample,chen2021lyapunov,chen2020finite,lee2024final}.

The ordinary differential equation (ODE) method~\citep{borkar2000ode} is a particularly useful tool for studying stochastic-approximation algorithms in RL\@. It replaces the noisy discrete-time recursion with a limiting continuous-time dynamical system and derives convergence from stability of the associated ODE\@. Although finite-time analyses provide sharper quantitative rates, the ODE viewpoint remains valuable because it often yields cleaner proofs, exposes the underlying dynamical mechanism, and provides a useful first stability check before pursuing more refined non-asymptotic results. This perspective has been applied to a broad range of RL and stochastic-approximation schemes~\citep{borkar2000ode,lee2020unified,sutton2009convergent,sutton2009fast,ghiassian2020gradient,lee2022new,melo2008analysis,bhatnagar2012stochastic}.

This paper develops an ODE-based convergence framework for standard Q-learning and several smooth Q-learning variants~\citep{haarnoja2017reinforcement,song2019revisiting,pan2020reinforcement,asadi2017alternative,barber2023smoothed}. Smooth variants replace the hard max operator in the Bellman update with a differentiable approximation, such as log-sum-exp (LSE)~\citep{haarnoja2017reinforcement}, mellowmax~\citep{asadi2017alternative}, or Boltzmann softmax~\citep{pan2020reinforcement}. These operators are often used to improve empirical behavior, encourage smoother action selection, and reduce overestimation effects~\citep{haarnoja2017reinforcement,pan2020reinforcement,song2019revisiting,barber2023smoothed}. However, a unified ODE-based convergence theory that covers both standard Q-learning and such smooth variants is still limited.

Our main idea is to replace the classical non-smooth $\infty$-norm Lyapunov argument with a smooth weighted degree-$2p$ polynomial Lyapunov function. The framework transfers $\infty$-norm contraction of Bellman-type operators to a differentiable Lyapunov certificate, while also handling preconditioned dynamics of the form $\dot x = D(F(x)-x)$. Here, $x\in\mathbb R^n$ is the state of the limiting ODE, $F:\mathbb R^n\to\mathbb R^n$ is the operator whose fixed point is the target solution, and $D$ is a positive diagonal matrix that scales the coordinatewise dynamics; in asynchronous Q-learning, its diagonal entries represent coordinate-update frequencies. At the ODE level, this allows a single proof template to cover the separately defined synchronous full-update mean field with $D=I$ and asynchronous mean fields with a positive diagonal $D$ induced by coordinate-update frequencies. For the max, LSE, and mellowmax operators, the induced Bellman operators are contractive and the associated ODEs are globally exponentially stable. For the Boltzmann operator, which is generally not contractive, the same framework yields convergence to an explicit invariant error set around the optimal Q-function.

The main contributions are summarized as follows:
\begin{enumerate}
\item \textbf{A smooth Lyapunov principle for preconditioned operator dynamics.}
We establish stability results for ODEs of the form $\dot x_t = D F(x_t)-D x_t$ using a weighted degree-$2p$ polynomial Lyapunov function. When $F$ is an $\infty$-norm contraction, the framework transfers this contraction to a differentiable Lyapunov certificate and avoids the non-differentiability of the $\infty$-norm.

\item \textbf{A unified ODE stability analysis for standard and smooth Q-learning.}
Applying the general result to Bellman operators, we prove global exponential stability of the ODE models associated with standard Q-learning and smooth variants based on the LSE and mellowmax operators. The analysis covers both the separately defined synchronous full-update mean field with $D=I$ and asynchronous mean fields with a positive diagonal $D$ induced by coordinate-update frequencies.

\item \textbf{A non-contractive stability result for Boltzmann smooth Q-learning.}
For the Boltzmann softmax operator, which need not be non-expansive, we prove that the corresponding ODE converges to an explicit error set around $Q^*_{\max}$. The radius of this set scales as $\gamma\lambda\ln(|{\cal A}|)w_{\max}/((1-\gamma)w_{\min})$, showing how the approximation error shrinks as the temperature parameter $\lambda$ decreases to zero.

\item \textbf{A unified asymptotic convergence theorem for the stochastic algorithms.}
Using the ODE stability results, we derive almost-sure convergence guarantees for the corresponding discrete-time Q-learning variants under independent and identically distributed (i.i.d.) sampling and standard diminishing step-sizes. The resulting analysis gives a single ODE-based proof architecture for standard Q-learning, contractive smooth Q-learning variants, and the non-contractive Boltzmann case.
\end{enumerate}

Overall, the paper provides a smooth Lyapunov-based ODE foundation for Q-learning and its smooth variants. The framework is intentionally asymptotic rather than finite-time: its purpose is to clarify the stability mechanism and to offer a compact route for proving convergence of both classical and smoothed Bellman updates.

\subsection{Related work}

Despite a growing body of work on smooth variants of Q-learning, an ODE-based convergence analysis has not yet been fully established. In what follows, we briefly review prior studies on smooth Q-learning and clarify how our contributions differ from and extend existing results.

{\bf Smooth Q-learning algorithms:}
Smooth Bellman updates have been studied in several settings. Haarnoja et al.~\citep{haarnoja2017reinforcement} developed deep energy-based RL with an LSE structure, while Song et al.~\citep{song2019revisiting} analyzed theoretical properties of the Boltzmann softmax Bellman operator and evaluated it with deep Q-learning. These works did not establish almost-sure convergence of the tabular stochastic-approximation recursions considered here. Asadi and Littman~\citep{asadi2017alternative} introduced the mellowmax operator, proved its non-expansiveness, and established convergence for generalized value iteration and a corresponding state--action--reward--state--action (SARSA) scheme. Pan et al.~\citep{pan2020reinforcement} later analyzed tabular learning with dynamic Boltzmann softmax updates, whereas Barber~\citep{barber2023smoothed} proposed a different convergent off-policy rule that replaces maximization with averaging. These studies do not establish stability of the associated ODE models. The direct finite-time analysis in~\cite{jeong2025unified} covers the LSE, Boltzmann, and mellowmax variants, but likewise does not provide a unified ODE stability theory. In contrast, our analysis develops one ODE framework for the standard max operator and all three smooth operators considered here.

{\bf ODE-based analysis of Q-learning:}
Recent work~\citep{lee2020unified} developed a switching-system model~\citep{liberzon2003switching} for asynchronous Q-learning and employed the ODE method~\citep{borkar2000ode}, leveraging tools from switching-system theory to establish global asymptotic stability without explicit Lyapunov-function constructions. A key limitation, however, is that its global stability result requires restrictive conditions (e.g., quasi-monotonicity) on the induced switching systems, which makes the approach difficult to extend to other RL algorithms such as smooth Q-learning variants. In contrast, the proposed method applies more broadly and yields a unified treatment of both standard Q-learning and a variety of smooth Q-learning algorithms. Consequently, it provides an alternative ODE-based route for analyzing asynchronous Q-learning beyond~\cite{lee2020unified}.

Prior ODE analyses~\citep{borkar2000ode,borkar1997analog} established stability of Q-learning mean fields using the $\infty$-norm as a Lyapunov function. The analysis in~\cite{borkar1997analog}, however, addressed only synchronous Q-learning, whereas the ODE models studied here include both synchronous and asynchronous Q-learning and their smooth variants. Moreover, $\infty$-norm--based arguments require careful treatment because the $\infty$-norm is non-differentiable. By using a smooth polynomial Lyapunov function induced by a weighted $2p$-norm, our approach avoids this difficulty while retaining a direct stability argument. Thus, relative to~\cite{borkar2000ode,borkar1997analog}, we cover positive-diagonally scaled mean fields arising from asynchronous updates and obtain concise stability proofs through a smooth polynomial certificate.

{\bf Finite-time analysis of smooth Q-learning:}
In~\cite{jeong2025unified}, a finite-time analysis was developed for smooth Q-learning algorithms based on the LSE, Boltzmann softmax, and mellowmax operators. That work analyzed the discrete-time algorithms directly without invoking an ODE framework. In contrast, this paper establishes asymptotic convergence through ODE analysis. Beyond the discrete-time iterates, we study a positive-diagonally scaled ODE family that includes the mean fields induced by asynchronous updates.

\section{Preliminaries}

\subsection{Markov decision process}
We consider a finite discounted Markov decision process (MDP) with state space ${\cal S}=\{1,2,\ldots,|{\cal S}|\}$, action space ${\cal A}=\{1,2,\ldots,|{\cal A}|\}$, transition probabilities $P(s'\mid s,a)$, deterministic rewards $r(s,a,s')\in{\mathbb R}$, and discount factor $\gamma\in(0,1)$. At time $k$, the agent observes $s_k$, selects $a_k$, receives $r_{k+1}:=r(s_k,a_k,s_{k+1})$, and moves to $s_{k+1}$ according to $P(\cdot\mid s_k,a_k)$.

A deterministic policy $\pi:{\cal S}\to{\cal A}$ induces the action-value function
\[
Q^{\pi}(s,a)={\mathbb E}\left[\left.\sum_{k=0}^{\infty}\gamma^k r_{k+1}\right|s_0=s,a_0=a,\pi\right].
\]
The optimal action-value function is $Q^*(s,a)=Q^{\pi^*}(s,a)$, where $\pi^*$ maximizes the expected discounted return over deterministic policies. Once $Q^*$ is known, an optimal policy is obtained by the greedy rule $\pi^*(s)\in\argmax_{a\in{\cal A}}Q^*(s,a)$. Whenever a state-sampling distribution $p$ is used below, it is treated as a fixed distribution. It may be chosen as the stationary distribution of an ergodic behavior chain, but ergodicity alone does not make successive observations independent; the convergence theorems separately assume independent resampling from $p$.

\subsection{Definitions}
This subsection introduces the definitions and auxiliary results used below.
Throughout the paper, we use the following notation for compact matrix--vector representations:
\begin{align}
P:=& \begin{bmatrix}
   P_1\\
   \vdots\\
   P_{|{\cal A}|}\\
\end{bmatrix}\in{\mathbb R}^{|{\cal S}\times {\cal A}| \times |{\cal S}|  },\; R:= \begin{bmatrix}
   R_1 \\
   \vdots \\
   R_{|{\cal A}|} \\
\end{bmatrix}\in {\mathbb R}^{|{\cal S}\times {\cal A}|},
\; Q:= \begin{bmatrix}
   Q(\cdot,1)\\
  \vdots\\
   Q(\cdot,|{\cal A}|)\\
\end{bmatrix}\in {\mathbb R}^{|{\cal S}\times {\cal A}|},\label{eq:definitions1}
\end{align}
where $P_a\in {\mathbb R}^{|{\cal S}| \times |{\cal S}|}$ is the transition probability matrix under the action $a\in {\cal A}$, $Q(\cdot,a)\in {\mathbb R}^{|{\cal S}|}$ for $a\in {\cal A}$, and $R_a(s):={\mathbb E}[r(s,a,s')\mid s,a]$.
In this notation, the Q-function is encoded as a single vector $Q \in {\mathbb R}^{|{\cal S}\times {\cal A}|}$ that enumerates $Q(s,a)$ for all $s \in {\cal S}$ and $a \in {\cal A}$. In particular, $Q(s,a) = (e_a \otimes e_s)^\top Q$, where $e_s \in {\mathbb R}^{|{\cal S}|}$ and $e_a \in {\mathbb R}^{|{\cal A}|}$ are the $s$-th and $a$-th standard basis vectors, respectively.
Similarly, $R(s,a) = (e_a \otimes e_s)^\top R$ and $P(s'\mid s,a) = (e_a \otimes e_s)^\top P e_{s'}$.

We focus on the weighted $2p$-norm and its associated polynomial Lyapunov function. For $p\in\mathbb N:=\{1,2,\ldots\}$, define
\begin{align}
\|x\|_{2p,w}&:=\left(\sum_{i=1}^n w_i x_i^{2p}\right)^{1/(2p)},\qquad
W_p(x):=\frac{1}{2p}\|x\|_{2p,w}^{2p}.\label{eq:Wp-definition}
\end{align}
The real numbers $w_i>0$, $i\in\{1,2,\ldots,n\}$, are fixed weights. The weighted $2p$-norm reduces to the standard $2p$-norm when $w_i=1$ for every $i$. Because $2p$ is even, $W_p$ is a polynomial and is therefore infinitely differentiable on all of $\mathbb R^n$, including the origin. Moreover, $\lim_{p\to\infty}\|x\|_{2p,w}=\|x\|_\infty$. This limiting relation is crucial in our analysis.
The following norm comparisons will be used in both the polynomial and norm-based estimates.
\begin{lemma}[Norm comparisons]\label{lemma:2}
For any $x\in {\mathbb R}^n$ and any $p\in\mathbb N$, we have
\begin{enumerate}
\item $w_{\min }^{1/(2p)}{\left\| x \right\|_\infty } \le {\left\| x \right\|_{2p,w}} \le {n^{1/(2p)}}w_{\max }^{1/(2p)}{\left\| x \right\|_\infty }$

\item $w_{\min }^{1/(2p)}{\left\| x \right\|_{2p}} \le {\left\| x \right\|_{2p,w}} \le w_{\max }^{1/(2p)}{\left\| x \right\|_{2p}}$

\item ${\left\| x \right\|_{\infty}} \le {\left\| x \right\|_{2p}} \le {n^{1/(2p)}}{\left\| x \right\|_{\infty}}$
\end{enumerate}
where ${w_{\min }}: = {\min _{i \in \{ 1,2, \ldots ,n\} }}{w_i}$ and ${w_{\max }}: = {\max _{i \in \{ 1,2, \ldots ,n\} }}{w_i}$.
\end{lemma}
\par\noindent
The proof of \cref{lemma:2} is provided in \cref{sec:app:lemma1-proof}.
\begin{definition}[Monotone operator]\label{def:monotone-operator}
A mapping $h:\mathbb R^n\to\mathbb R$ is monotone with respect to the componentwise order if $u_i\le v_i$ for every $i$ implies $h(u)\le h(v)$.
\end{definition}
Define the mappings ${h_{\max }}:{\mathbb R}^n \to {\mathbb R}$, ${h_{\rm mm }}:{\mathbb R}^n \to {\mathbb R}$, ${h_{\rm lse}}: {\mathbb R}^n \to {\mathbb R}$, and ${h_{\rm bz}}: {\mathbb R}^n \to {\mathbb R}$ by
\begin{align*}
h_{\max}(x)
&:=\max_{i\in\{1,2,\ldots,n\}}x_i,
&h_{\rm lse}^{\lambda}(x)
&:=\lambda\ln\!\left(\sum_{i=1}^{n}e^{x_i/\lambda}\right),\\
h_{\rm mm}^{\lambda}(x)
&:=\lambda\ln\!\left(\frac{1}{n}\sum_{i=1}^{n}e^{x_i/\lambda}\right),
&h_{\rm bz}^{\lambda}(x)
&:=\frac{\sum_{i=1}^{n}x_i e^{x_i/\lambda}}
{\sum_{i=1}^{n}e^{x_i/\lambda}}.
\end{align*}
Here, $\lambda>0$ is the temperature parameter for the log-sum-exp (LSE), mellowmax, and Boltzmann operators, and each operator uses the factor $e^{x_i/\lambda}$. The mapping $h_{\max}$ is the standard max operator. The mapping $h_{\rm lse}^{\lambda}$ is the LSE, or smooth-max, operator used in machine learning and reinforcement learning (RL)~\citep{haarnoja2017reinforcement,dai2018sbeed}. The mellowmax operator $h_{\rm mm}^{\lambda}$ was introduced in~\cite{asadi2017alternative} to address limitations of the Boltzmann softmax operator $h_{\rm bz}^{\lambda}$. The latter forms a smooth approximation of the max operator through softmax weights that are widely used in RL~\citep{pan2020reinforcement,gao2017properties}.
For each choice of $h\in \{h_{\max},h_{\rm lse}^{\lambda},h_{\rm mm}^{\lambda},h_{\rm bz}^{\lambda} \}$, define $H: {\mathbb R}^{|{\mathcal S}\times {\mathcal A}|}\to {\mathbb R}^{|{\mathcal S}|}$ by
\[H(Q): = \left[ {\begin{array}{*{20}{c}}
 h(Q(1,\cdot ))\\
 h(Q(2,\cdot ))\\
 \vdots \\
h (Q(|{\mathcal S}|, \cdot ))
\end{array}} \right] \in {\mathbb R}^{|{\mathcal S}|}.\]
Accordingly, $H_{\max}$, $H_{\rm lse}^{\lambda}$, $H_{\rm mm}^{\lambda}$, and $H_{\rm bz}^{\lambda}$ denote the mappings obtained from $h_{\max}$, $h_{\rm lse}^{\lambda}$, $h_{\rm mm}^{\lambda}$, and $h_{\rm bz}^{\lambda}$, respectively.
In this paper, we consider smooth variants of Q-learning based on these approximations of the max operator and analyze their convergence in a unified manner using ordinary differential equation (ODE) analysis.
The proofs of the two main polynomial Lyapunov-function-based stability theorems, together with the norm-based and stochastic-approximation results, are provided in the Appendix.

\section{Stability of nonlinear ODE models under contraction}

We consider the following ODE:
\begin{align}
\frac{d}{{dt}}{x_t} = DF({x_t}) - D{x_t},\quad \forall t\ge 0,\quad  x_0 \in {\mathbb R}^n\label{eq:ODE1}
\end{align}
where $t\ge 0$ is the continuous time, $x_t \in {\mathbb R}^n$ is the state at time $t$, $F: {\mathbb R}^n \to {\mathbb R}^n$ is a mapping that will be specified later, and $D \in {\mathbb R}^{n\times n}$ is a positive definite diagonal matrix with strictly positive diagonal elements $d_i>0$, $i\in\{1,2,\ldots,n\}$. This ODE will be used to model Q-learning and its variants in the remainder of the paper.
A similar ODE was originally considered in~\cite{borkar1997analog}. The distinction here is the presence of $D$: when $D=I_n$, where $I_n$ is the $n$-dimensional identity matrix,~\eqref{eq:ODE1} reduces to the ODE in~\cite{borkar1997analog}. When $D$ is later induced by the update probabilities of the one-coordinate recursion in~\cref{algo:Q-learning}, its diagonal entries sum to one, so $D=I_n$ does not arise from that sampler unless $n=1$. Instead, $D=I_n$ represents the mean field of a separately defined synchronous full-update recursion in which every coordinate is updated at each iteration.
To account for the diagonal scaling, we choose $w_i=1/d_i$ and use the polynomial Lyapunov function $W_p$ in~\eqref{eq:Wp-definition}. Because $2p$ is even, this function is smooth on all of ${\mathbb R}^n$, including at the origin. The following result is the first main stability result for~\eqref{eq:ODE1}.

\begin{theorem}[Polynomial contractive certificate]\label{thm:stability2}
Let $x_t$ be the unique solution of~\eqref{eq:ODE1}. Suppose that $F:{\mathbb R}^n\to{\mathbb R}^n$ is an $\infty$-norm contraction: for some $\alpha\in(0,1)$,
\[
\|F(x)-F(y)\|_\infty\le \alpha\|x-y\|_\infty,
\qquad x,y\in{\mathbb R}^n.
\]
Let $x^*$ be its unique fixed point, and choose $w_i=1/d_i$. For any $p\in\mathbb N$ satisfying
\[
\left\lceil \frac{\ln n}{\ln(\alpha^{-1})}\right\rceil<2p,
\]
we have
\begin{align}
\frac{d}{dt}W_p(x_t-x^*)
&\le -\left(1-\alpha n^{1/(2p)}\right)\|x_t-x^*\|_{2p}^{2p}\label{eq:Wp-contract-derivative}\\
&\le -\frac{2p\left(1-\alpha n^{1/(2p)}\right)}{w_{\max}}W_p(x_t-x^*),\nonumber
\end{align}
and hence
\begin{align}
W_p(x_t-x^*)
\le \exp\!\left(-\frac{2p\left(1-\alpha n^{1/(2p)}\right)}{w_{\max}}t\right)W_p(x_0-x^*),
\qquad t\ge0,\label{eq:Wp-contract-bound}
\end{align}
where $w_{\max}:=\max_{i\in\{1,2,\ldots,n\}}w_i$. Therefore, $W_p$ is a globally smooth polynomial Lyapunov function and $x^*$ is the unique globally exponentially stable equilibrium.
\end{theorem}
\par\noindent
The proof of \cref{thm:stability2} is provided in \cref{sec:app:thm1-proof}.
\Cref{thm:stability2} uses $W_p$ itself as the Lyapunov function. The condition on $p$ guarantees that $1-\alpha n^{1/(2p)}>0$. For any fixed $n$ and $\alpha\in(0,1)$, a sufficiently large integer $p$ satisfies this condition, while $W_p$ remains a finite-degree polynomial. As $p\to\infty$, the contraction margin $1-\alpha n^{1/(2p)}$ approaches $1-\alpha$.

Weighted $p$-norm Lyapunov functions have previously been used to analyze the contractive dynamics
\[
\dot x_t=F(x_t)-x_t
\]
in~\cite{borkar1997analog,borkar2009stochastic}. In those works, the Lyapunov function is matched to a mapping that is contractive in the same weighted $p$-norm. In contrast, in~\Cref{thm:stability2}, $F$ is assumed to be contractive only in the $\infty$-norm, while the polynomial $W_p$ is induced by a weighted finite $2p$-norm. The comparison between the finite $2p$-norm and the $\infty$-norm transfers the contraction property to the smooth polynomial certificate. This approach avoids the directional-derivative or subgradient arguments required when the non-differentiable $\infty$-norm is used directly, accommodates the preconditioned dynamics in~\eqref{eq:ODE1}, and keeps the stability proof concise.

Next, we consider a mapping that is uniformly close to an $\infty$-norm contraction.
\begin{theorem}[Polynomial perturbation certificate]\label{thm:stability4}
Suppose that $F:{\mathbb R}^n\to{\mathbb R}^n$ is globally Lipschitz continuous, and let $x_t$ be the unique global solution of~\eqref{eq:ODE1}. Suppose that $F':{\mathbb R}^n\to{\mathbb R}^n$ is an $\infty$-norm contraction with factor $\alpha\in(0,1)$ and fixed point $x^*$:
\[
\|F'(x)-F'(y)\|_\infty\le \alpha\|x-y\|_\infty,
\qquad x,y\in{\mathbb R}^n.
\]
Suppose also that $F$ satisfies
\[
\|F(x)-F'(x)\|_\infty\le\eta,
\qquad x\in{\mathbb R}^n,
\]
for some $\eta>0$. Choose $w_i=1/d_i$, and let
$w_{\min}:=\min_i w_i$ and $w_{\max}:=\max_i w_i$. For any $p\in\mathbb N$ satisfying
\[
\left\lceil \frac{\ln n}{\ln(\alpha^{-1})}\right\rceil<2p,
\]
we have
\begin{align}
\frac{d}{dt}W_p(x_t-x^*)
\le{}&-\left(1-\alpha n^{1/(2p)}\right)\|x_t-x^*\|_{2p}^{2p}
+n^{1/(2p)}\eta\|x_t-x^*\|_{2p}^{2p-1}.\label{eq:Wp-perturb-derivative}
\end{align}
Moreover,
\begin{align}
\|x_t-x^*\|_{2p,w}
\le{}&\|x_0-x^*\|_{2p,w}
\exp\!\left(\frac{\alpha n^{1/(2p)}-1}{w_{\max}}t\right)\nonumber\\
&+\frac{n^{1/(2p)}\eta w_{\max}}
{\left(1-\alpha n^{1/(2p)}\right)w_{\min}^{(2p-1)/(2p)}},
\qquad t\ge0.\label{eq:Wp-perturb-bound}
\end{align}
In particular, $W_p(x_t-x^*)$ decreases whenever
$\|x_t-x^*\|_{2p}>n^{1/(2p)}\eta/(1-\alpha n^{1/(2p)})$.
\end{theorem}
\par\noindent
The proof of \cref{thm:stability4} is provided in \cref{sec:app:thm2-proof}.
\Cref{thm:stability4} shows that the polynomial Lyapunov function is strictly decreasing outside an explicit neighborhood of $x^*$. Thus, even when $F$ is not itself contractive, the solution of~\eqref{eq:ODE1} is ultimately bounded under the stated uniform approximation condition. This result will be used for the Boltzmann Q-learning ODE.

\section{ODE model of Q-learning and its smooth variants}

In this section, we consider the following specific ODE model associated with Q-learning and analyze its convergence properties:
\begin{align}
\frac{d}{{dt}}{Q_t} =& DR + \gamma DPH({Q_t}) - D{Q_t},\quad \forall t\ge 0,\quad Q_0 \in {\mathbb R}^{|{\mathcal S}\times {\mathcal A}|}.\label{eq:ODE-Q-learning1}
\end{align}
Here, $H \in \{H_{\max}, H_{\rm lse}^{\lambda}, H_{\rm mm}^{\lambda}, H_{\rm bz}^{\lambda} \}$, and $D \in {\mathbb R}^{n\times n}$ is a positive definite diagonal matrix with strictly positive diagonal elements $d_i>0$, $i\in\{1,2,\ldots,n\}$.
The ODE is a special case of~\eqref{eq:ODE1} and is the continuous-time model commonly associated with Q-learning and its smooth variants. Define the Bellman operator
\[
F(Q):=R+\gamma PH(Q).
\]
Then~\eqref{eq:ODE-Q-learning1} can be written as
\begin{align}
\frac{d}{{dt}}{Q_t}=DF(Q_t)-DQ_t,\quad \forall t\ge0,\quad Q_0\in{\mathbb R}^{|{\mathcal S}\times{\mathcal A}|},\label{eq:ODE-Q-learning2}
\end{align}
which has the form of~\eqref{eq:ODE1}. The four Bellman operators considered here are
\begin{align*}
F_{\max}(Q)&:=R+\gamma PH_{\max}(Q),
&F_{\rm bz}^{\lambda}(Q)&:=R+\gamma PH_{\rm bz}^{\lambda}(Q),\\
F_{\rm lse}^{\lambda}(Q)&:=R+\gamma PH_{\rm lse}^{\lambda}(Q),
&F_{\rm mm}^{\lambda}(Q)&:=R+\gamma PH_{\rm mm}^{\lambda}(Q).
\end{align*}
The max, LSE, and mellowmax operators are non-expansive~\citep{asadi2017alternative,dai2018sbeed}; consequently, their Bellman operators are $\gamma$-contractions by~\cref{lemma:contraction}. Their unique fixed points are defined by
\begin{align}
Q_{\max}^*&=F_{\max}(Q_{\max}^*),
&Q_{\rm lse}^{\lambda}&=F_{\rm lse}^{\lambda}(Q_{\rm lse}^{\lambda}),
&Q_{\rm mm}^{\lambda}&=F_{\rm mm}^{\lambda}(Q_{\rm mm}^{\lambda}).\label{eq:4}
\end{align}
The Boltzmann softmax operator is not non-expansive in general~\citep{asadi2017alternative}. We therefore first treat the three contractive cases and then use the perturbation result for the Boltzmann case.

\begin{theorem}\label{thm:stability5}
Consider~\eqref{eq:ODE-Q-learning1} with
$H\in\{H_{\max},H_{\rm lse}^{\lambda},H_{\rm mm}^{\lambda}\}$, and let $Q_t$ be its solution. Let $Q_e$ denote the corresponding fixed point in~\eqref{eq:4}, set $n=|{\cal S}\times{\cal A}|$, choose $w_i=1/d_i$, and let $w_{\max}:=\max_i w_i$. For any $p\in\mathbb N$ satisfying
\[
\left\lceil \frac{\ln n}{\ln(\gamma^{-1})}\right\rceil<2p,
\]
we have
\begin{align}
W_p(Q_t-Q_e)
\le \exp\!\left(-\frac{2p\left(1-\gamma n^{1/(2p)}\right)}{w_{\max}}t\right)
W_p(Q_0-Q_e),\qquad t\ge0.\label{eq:Wp-Q-contract}
\end{align}
Therefore, $Q_e$ is the unique globally exponentially stable equilibrium of the corresponding ODE.
\end{theorem}
\par\noindent
The proof of \cref{thm:stability5} is provided in \cref{sec:app:thm3-proof}.

\Cref{thm:stability5} states that the ODE associated with the max, LSE, or mellowmax operator converges exponentially to its respective fixed point, with the smooth polynomial $W_p$ serving as a Lyapunov function.
Because the Boltzmann softmax operator is not contractive in general, we instead combine its uniform approximation to the max operator with the polynomial perturbation certificate in~\cref{thm:stability4} to obtain the following result.
\begin{theorem}\label{thm:stability6}
Consider~\eqref{eq:ODE-Q-learning1} with $H=H_{\rm bz}^{\lambda}$, and let $Q_t$ be its solution. Set $n=|{\cal S}\times{\cal A}|$, choose $w_i=1/d_i$, and let $w_{\min}:=\min_i w_i$ and $w_{\max}:=\max_i w_i$. For any $p\in\mathbb N$ satisfying
\[
\left\lceil \frac{\ln n}{\ln(\gamma^{-1})}\right\rceil<2p,
\]
we have
\begin{align}
\frac{d}{dt}W_p(Q_t-Q_{\max}^*)
\le{}&-\left(1-\gamma n^{1/(2p)}\right)
\|Q_t-Q_{\max}^*\|_{2p}^{2p}\nonumber\\
&+n^{1/(2p)}\gamma\lambda\ln(|\mathcal A|)
\|Q_t-Q_{\max}^*\|_{2p}^{2p-1},\label{eq:Wp-Q-boltzmann-derivative}
\end{align}
and
\begin{align}
\|Q_t-Q_{\max}^*\|_{2p,w}
\le{}&\|Q_0-Q_{\max}^*\|_{2p,w}
\exp\!\left(\frac{\gamma n^{1/(2p)}-1}{w_{\max}}t\right)\nonumber\\
&+\frac{n^{1/(2p)}w_{\max}}
{\left(1-\gamma n^{1/(2p)}\right)w_{\min}^{(2p-1)/(2p)}}
\gamma\lambda\ln(|\mathcal A|),
\qquad t\ge0.\label{eq:Wp-Q-boltzmann-bound}
\end{align}
\end{theorem}
\par\noindent
The proof of \cref{thm:stability6} is provided in \cref{sec:app:thm4-proof}.
\Cref{thm:stability6} shows that $W_p$ decreases outside an explicit neighborhood of $Q_{\max}^*$ and gives a quantitative ultimate bound in the weighted $2p$-norm. The exact $\infty$-norm invariant-set counterpart and its numerical illustration are collected in the appendix.

\section{Q-learning and its smooth variants}

We consider the recursion in~\cref{algo:Q-learning}, where
\[{Q_k}({s_k}', \cdot ): = \left[ {\begin{array}{*{20}{c}}
{{Q_k}({s_k}',1)}\\
{{Q_k}({s_k}',2)}\\
 \vdots \\
{{Q_k}({s_k}',|{\mathcal A}|)}
\end{array}} \right] \in {\mathbb R}^{|{\mathcal A}|},\]
and $h: {\mathbb R}^{|{\mathcal A}|} \to {\mathbb R}$ is chosen from $\{h_{\max},h_{\rm lse}^{\lambda},h_{\rm mm}^{\lambda},h_{\rm bz}^{\lambda} \}$. The choice $h=h_{\max}$ gives standard Q-learning~\citep{watkins1992q}; $h=h_{\rm lse}^{\lambda}$ gives the LSE variant~\citep{haarnoja2017reinforcement}; $h=h_{\rm mm}^{\lambda}$ gives the mellowmax variant~\citep{asadi2017alternative}; and $h=h_{\rm bz}^{\lambda}$ gives the Boltzmann softmax variant~\citep{pan2020reinforcement,song2019revisiting}.
\begin{algorithm}[h!]
\caption{Asynchronous Q-learning variants under independent sampling}
\label{algo:Q-learning}
\begin{algorithmic}[1]
\State Initialize $Q_0 \in {\mathbb R}^{|{\mathcal S}\times {\mathcal A}|}$
\For{$k=0,1,\ldots$}
  \State Sample $s_k \sim p(\cdot)$ and $a_k \sim \beta(\cdot\mid s_k)$
  \State Sample $s_k' \sim P(\cdot\mid s_k,a_k)$ and $r(s_k,a_k,s_k')$
  \State Update $Q_{k+1}(s_k,a_k) = Q_k(s_k,a_k) + \alpha_k\bigl\{ r(s_k,a_k,s_k') + \gamma\, h(Q_k(s_k',\cdot)) - Q_k(s_k,a_k)\bigr\}$
\EndFor
\end{algorithmic}
\end{algorithm}

\Cref{algo:Q-learning} is the asynchronous single-coordinate recursion. Its mean field contains the diagonal matrix with entries $d(s,a)=p(s)\beta(a\mid s)$. A synchronous full-update recursion is a separate algorithm that updates every state--action coordinate at each iteration using one Bellman sample for each coordinate; its mean field is obtained by setting $D=I$ in~\eqref{eq:ODE-Q-learning2}. Thus, $D=I$ is not obtained by assigning sampling probabilities in~\cref{algo:Q-learning}.

In this paper, $\{(s_k,a_k,s_k')\}_{k=0}^{\infty}$ are independent and identically distributed (i.i.d.) samples generated as follows: $s_k\sim p$, $a_k\sim\beta(\cdot\mid s_k)$, and $s_k'\sim P(\cdot\mid s_k,a_k)$. This is a generative-model or independent replay-sampling assumption, not the distribution of successive observations along a single Markov trajectory. In particular, ergodicity of a behavior chain may justify the choice of a stationary distribution $p$, but it does not imply the independence required here.
Note that $s_k'$ denotes the next-state sample associated with the independently sampled pair $(s_k,a_k)$ at iteration $k$. The current state used at iteration $k+1$ is resampled from $p$ and is therefore independent of $s_k$. Under this assumption, the joint state--action distribution at every iteration is
\begin{align}
d(s,a) = p(s)\beta(a\mid s),\quad (s,a) \in {\mathcal S} \times {\mathcal A}.\label{eq:d}
\end{align}
Throughout the paper, we adopt the following assumption for convenience.
\begin{assumption}\label{assumption:positive-distribution}
$d(s,a)> 0$ holds for all $(s,a)\in {\mathcal S} \times {\mathcal A}$.
\end{assumption}
Under the i.i.d.\ sampling model,~\cref{assumption:positive-distribution} guarantees that every state--action pair is sampled infinitely often with probability one. Similar assumptions have been adopted in~\cite{li2020sample} and~\cite{chen2021lyapunov}. In contrast,~\cite{beck2012error} imposes an alternative exploration requirement, known as the \emph{cover time} condition, which posits that within a prescribed time horizon every state--action pair is expected to be visited at least once. Related variants of cover-time conditions have also been used in~\cite{even2003learning} and~\cite{li2020sample} to derive convergence-rate guarantees.

In practice, replay sampling can reduce temporal dependence, but samples drawn from a finite replay buffer are not generally exactly i.i.d.\ The convergence results below require the i.i.d.\ model stated above. The analysis can in principle be extended to Markovian observation settings using recently developed tools, albeit at the cost of substantially more involved arguments~\citep{lim2024finite,qu2020finite}. Such extensions, however, are not the main focus of this paper. Our primary goal is to develop a useful framework that interfaces naturally with the ODE method and the Borkar--Meyn theorem, two widely used tools for establishing convergence in RL. Many subsequent refinements build on these tools. The corresponding numerical illustration of smooth Q-learning with the Boltzmann operator is provided in~\cref{fig:4}.

\section{Convergence analysis}\label{sec:convergence-summary}

This section summarizes the convergence consequences of the ODE stability results. The detailed stochastic-approximation verification and the proofs of the two convergence claims are provided in~\cref{sec:app:convergence-analysis}.
Under the i.i.d.\ sampling model in~\cref{algo:Q-learning} and the step-size conditions $\alpha_k>0$, $\sum_{k=0}^{\infty}\alpha_k=\infty$, and $\sum_{k=0}^{\infty}\alpha_k^2<\infty$, the update can be written as a stochastic approximation recursion
\[
Q_{k+1}=Q_k+\alpha_k\{f(Q_k)+\varepsilon_{k+1}\},
\]
where $f(Q)=DR+\gamma DPH(Q)-DQ$ and $\varepsilon_{k+1}$ is a martingale-difference noise term. Thus convergence follows by combining the ODE stability results in~\cref{thm:stability5,thm:stability6} with standard ODE-based stochastic approximation arguments.
For $h\in\{h_{\max},h_{\rm lse}^{\lambda},h_{\rm mm}^{\lambda}\}$, the corresponding Bellman operator is an $\infty$-norm contraction. Hence the associated ODE is globally exponentially stable, and the stochastic iterates converge almost surely to the corresponding fixed point $Q_{\max}^*$, $Q_{\rm lse}^{\lambda}$, or $Q_{\rm mm}^{\lambda}$; see~\cref{sec:app:convergence-analysis} and~\cref{thm:Q-learning-convergence}.
For the Boltzmann softmax operator, the induced Bellman operator need not be contractive. In this case, the ODE and the stochastic iterates are guaranteed to approach the invariant error set
\[
{\cal H}:=\left\{Q\in{\mathbb R}^{|{\cal S}\times{\cal A}|}:\|Q-Q_{\max}^*\|_{\infty}\le \frac{\gamma\lambda\ln(|{\cal A}|)w_{\max}}{(1-\gamma)w_{\min}}\right\}.
\]
Consequently, the limiting error shrinks as the temperature parameter $\lambda$ decreases to zero; see~\cref{sec:app:convergence-analysis} and~\cref{thm:Q-learning-convergence2}.

\section*{Conclusion}
This paper presents a unified ODE framework for analyzing convergence of Q-learning and its smooth variants. A smooth polynomial Lyapunov function treats positive-diagonal mean fields arising from asynchronous updates, while $D=I$ represents the separately defined synchronous full-update mean field. The framework also complements switching-system analyses of asynchronous Q-learning by avoiding restrictive structural conditions on the ODE model.



\bibliographystyle{iclr2027_conference}
\bibliography{paper}

\clearpage
\appendix
\begin{center}
{\LARGE\bfseries Appendix}
\end{center}
\vspace{0.5em}

\section{Proof of Lemma 1}\label{sec:app:lemma1-proof}

\begin{framed}
\noindent\textbf{Restatement of \Cref{lemma:2} (Norm comparisons).}\par\medskip
For any $x\in\mathbb R^n$ and any $p\in\mathbb N$, we have
\begin{enumerate}
\item $w_{\min}^{1/(2p)}\|x\|_\infty\le\|x\|_{2p,w}
\le n^{1/(2p)}w_{\max}^{1/(2p)}\|x\|_\infty$;
\item $w_{\min}^{1/(2p)}\|x\|_{2p}\le\|x\|_{2p,w}
\le w_{\max}^{1/(2p)}\|x\|_{2p}$;
\item $\|x\|_\infty\le\|x\|_{2p}\le n^{1/(2p)}\|x\|_\infty$,
\end{enumerate}
where $w_{\min}:=\min_i w_i$ and $w_{\max}:=\max_i w_i$.
\end{framed}

\begin{proof}[Proof of \Cref{lemma:2}]
For the first statement, the lower bound follows from
\[{\left\| x \right\|_{2p,w}} \ge {({w_j}{x_j}^{2p})^{1/(2p)}} = w_j^{1/(2p)}|{x_j}| \ge w_{\min }^{1/(2p)}|{x_j}|,\]
for any $j\in \{1,2,\ldots,n\}$. By taking the maximum over $j$, we have ${\left\| x \right\|_{2p,w}} \ge w_{\min }^{1/(2p)}{\left\| x \right\|_\infty }$.
For the upper bound, we have
\begin{align*}
{\left\| x \right\|_{2p,w}} =& {\left( {\sum\limits_{i = 1}^n {{w_i}{x_i}^{2p}} } \right)^{1/(2p)}}\le  {\left( {\sum\limits_{i = 1}^n {{w_i}\left\| x \right\|_\infty ^{2p}} } \right)^{1/(2p)}}= {\left\| x \right\|_\infty }{\left( {\sum\limits_{i = 1}^n {{w_i}} } \right)^{1/(2p)}}\\
\le& {\left\| x \right\|_\infty }{\left( {n{w_{\max }}} \right)^{1/(2p)}} = {n^{1/(2p)}}w_{\max }^{1/(2p)}{\left\| x \right\|_\infty }.
\end{align*}
The second statement follows from the inequalities
\begin{align*}
\|x\|_{2p,w}
&=\left(\sum_{i=1}^n w_ix_i^{2p}\right)^{1/(2p)}\\
&\le \left(w_{\max}\sum_{i=1}^nx_i^{2p}\right)^{1/(2p)}\\
&=w_{\max}^{1/(2p)}\left(\sum_{i=1}^nx_i^{2p}\right)^{1/(2p)}
=w_{\max}^{1/(2p)}\|x\|_{2p}
\end{align*}
and
\begin{align*}
\|x\|_{2p,w}
&=\left(\sum_{i=1}^n w_ix_i^{2p}\right)^{1/(2p)}\\
&\ge \left(w_{\min}\sum_{i=1}^nx_i^{2p}\right)^{1/(2p)}\\
&=w_{\min}^{1/(2p)}\left(\sum_{i=1}^nx_i^{2p}\right)^{1/(2p)}
=w_{\min}^{1/(2p)}\|x\|_{2p}.
\end{align*}
For the third statement, the largest coordinate satisfies
\begin{align}
\|x\|_\infty^{2p}
=\max_{1\le i\le n}x_i^{2p}
\le\sum_{i=1}^nx_i^{2p}
\le n\max_{1\le i\le n}x_i^{2p}
=n\|x\|_\infty^{2p}.
\label{eq:app:unweighted-norm-comparison}
\end{align}
Taking the $2p$-th root of every term in~\eqref{eq:app:unweighted-norm-comparison} gives
$\|x\|_\infty\le\|x\|_{2p}\le n^{1/(2p)}\|x\|_\infty$.
This completes the proof.
\end{proof}

\section{Proof of Theorem 1}\label{sec:app:thm1-proof}

\begin{framed}
\noindent\textbf{Restatement of \Cref{thm:stability2} (Polynomial contractive certificate).}\par\medskip
Let $x_t$ be the unique solution of~\eqref{eq:ODE1}. For convenience, the equation referred to as~\eqref{eq:ODE1} is repeated here:
\begin{align*}
\frac{d}{dt}x_t=D\bigl(F(x_t)-x_t\bigr),
\qquad t\ge0,
\qquad x_0\in{\mathbb R}^n.
\end{align*}
Suppose that $F:{\mathbb R}^n\to{\mathbb R}^n$ is an $\infty$-norm contraction: for some $\alpha\in(0,1)$,
\[
\|F(x)-F(y)\|_\infty\le \alpha\|x-y\|_\infty,
\qquad x,y\in{\mathbb R}^n.
\]
Let $x^*$ be its unique fixed point, and choose $w_i=1/d_i$. For any $p\in\mathbb N$ satisfying
\[
\left\lceil \frac{\ln n}{\ln(\alpha^{-1})}\right\rceil<2p,
\]
we have
\begin{align*}
\frac{d}{dt}W_p(x_t-x^*)
&\le -\left(1-\alpha n^{1/(2p)}\right)\|x_t-x^*\|_{2p}^{2p}\\
&\le -\frac{2p\left(1-\alpha n^{1/(2p)}\right)}{w_{\max}}W_p(x_t-x^*),
\end{align*}
and hence
\begin{align*}
W_p(x_t-x^*)
\le \exp\!\left(-\frac{2p\left(1-\alpha n^{1/(2p)}\right)}{w_{\max}}t\right)W_p(x_0-x^*),
\qquad t\ge0,
\end{align*}
where $w_{\max}:=\max_{i\in\{1,2,\ldots,n\}}w_i$. Therefore, $W_p$ is a globally smooth polynomial Lyapunov function and $x^*$ is the unique globally exponentially stable equilibrium.
\end{framed}

Before estimating the Lyapunov derivative, we make the well-posedness and equilibrium arguments explicit. Since $({\mathbb R}^n,\|\cdot\|_\infty)$ is complete and $F$ is a contraction, the Banach fixed-point theorem guarantees a unique fixed point $x^*$. Moreover, every contraction is globally Lipschitz, so the vector field $x\mapsto D(F(x)-x)$ is globally Lipschitz. Hence, \cref{lemma:existence} guarantees that~\eqref{eq:ODE1} has a unique global solution for every initial state. Since $F(x^*)=x^*$, subtracting the constant equilibrium from~\eqref{eq:ODE1} gives the exact error dynamics
\begin{align}
\frac{d}{dt}(x_t-x^*)
=D\bigl(F(x_t)-F(x^*)-(x_t-x^*)\bigr).
\label{eq:app:error-dynamics-contract}
\end{align}
For the componentwise calculations below, $F_i(x)$ denotes the $i$th component of $F(x)$, so
$F(x)=(F_1(x),\ldots,F_n(x))^\top$.
The following three lemmas establish the polynomial comparison, derivative, and integration steps separately.

\begin{framed}
\begin{lemma}[Two-sided comparison for $W_p$]\label{lemma:app:Wp-two-sided}
Let $w_{\min}:=\min_{1\le i\le n}w_i$ and $w_{\max}:=\max_{1\le i\le n}w_i$. For every $z\in\mathbb R^n$,
\begin{align}
\frac{w_{\min}}{2p}\|z\|_{2p}^{2p}
\le W_p(z)
\le \frac{w_{\max}}{2p}\|z\|_{2p}^{2p}.
\label{eq:app:Wp-two-sided-contract}
\end{align}
\end{lemma}
\end{framed}
\begin{proof}
The main-text definition in~\eqref{eq:Wp-definition} is repeated here for convenience:
\begin{align*}
\|z\|_{2p,w}
&:=\left(\sum_{i=1}^n w_i z_i^{2p}\right)^{1/(2p)},
&
W_p(z)
&:=\frac{1}{2p}\|z\|_{2p,w}^{2p}.
\end{align*}
Because $2p$ is even, every $z_i^{2p}$ is nonnegative, and the coordinate form is
\begin{align}
W_p(z)=\frac{1}{2p}\sum_{i=1}^n w_i z_i^{2p}.
\label{eq:app:Wp-coordinate-form}
\end{align}
Using $w_{\min}\le w_i\le w_{\max}$ in~\eqref{eq:app:Wp-coordinate-form} gives
\begin{align*}
\frac{w_{\min}}{2p}\|z\|_{2p}^{2p}
&=\frac{w_{\min}}{2p}\sum_{i=1}^n z_i^{2p}
\le \frac{1}{2p}\sum_{i=1}^n w_i z_i^{2p}\\
&=W_p(z)
\le \frac{w_{\max}}{2p}\sum_{i=1}^n z_i^{2p}
=\frac{w_{\max}}{2p}\|z\|_{2p}^{2p},
\end{align*}
which proves~\eqref{eq:app:Wp-two-sided-contract}.
\end{proof}

\begin{framed}
\begin{lemma}[Derivative estimate for \Cref{thm:stability2}]\label{lemma:app:thm1-derivative}
Under the assumptions of \cref{thm:stability2},
\begin{align*}
\frac{d}{dt}W_p(x_t-x^*)
&\le -\left(1-\alpha n^{1/(2p)}\right)\|x_t-x^*\|_{2p}^{2p}\\
&\le -\frac{2p\left(1-\alpha n^{1/(2p)}\right)}{w_{\max}}W_p(x_t-x^*).
\end{align*}
\end{lemma}
\end{framed}
\begin{proof}
Differentiating the polynomial coordinate form in~\eqref{eq:app:Wp-coordinate-form} with respect to $z_i$ gives
\begin{align}
\frac{\partial W_p(z)}{\partial z_i}
&=\frac{w_i}{2p}\frac{d}{dz_i}z_i^{2p}
=w_i z_i^{2p-1},
\qquad i\in\{1,2,\ldots,n\},
\label{eq:app:Wp-gradient}
\end{align}
and hence
\[
\nabla_z W_p(z)
=\begin{bmatrix}
w_1z_1^{2p-1}\\
w_2z_2^{2p-1}\\
\vdots\\
w_nz_n^{2p-1}
\end{bmatrix}
=D^{-1}\begin{bmatrix}
z_1^{2p-1}\\
z_2^{2p-1}\\
\vdots\\
z_n^{2p-1}
\end{bmatrix}.
\]
Applying the chain rule using~\eqref{eq:app:Wp-gradient} and then applying the error dynamics in~\eqref{eq:app:error-dynamics-contract} coordinatewise yields
\begin{align}
\frac{d}{dt}W_p(x_t-x^*)
&=\sum_{i=1}^n w_i(x_{t,i}-x_i^*)^{2p-1}\dot x_{t,i}\nonumber\\
&=\sum_{i=1}^n w_i d_i(x_{t,i}-x_i^*)^{2p-1}\nonumber\\
&\quad\times\bigl(F_i(x_t)-F_i(x^*)-(x_{t,i}-x_i^*)\bigr).
\label{eq:app:Wp-chain-contract}
\end{align}
Using $w_i d_i=1$ in~\eqref{eq:app:Wp-chain-contract} gives
\begin{align}
\frac{d}{dt}W_p(x_t-x^*)
={}&\sum_{i=1}^n (x_{t,i}-x_i^*)^{2p-1}
\bigl(F_i(x_t)-F_i(x^*)\bigr)\nonumber\\
&-\sum_{i=1}^n(x_{t,i}-x_i^*)^{2p}.
\label{eq:app:Wp-split-contract}
\end{align}
The exponents $2p/(2p-1)$ and $2p$ are conjugate because
\begin{align*}
\frac{1}{2p/(2p-1)}+\frac{1}{2p}
=\frac{2p-1}{2p}+\frac{1}{2p}=1.
\end{align*}
Moreover, $(2p-1)\,2p/(2p-1)=2p$. Therefore, H{\"o}lder's inequality and the fact that $2p$ is even give
\begin{align}
&\left|\sum_{i=1}^n (x_{t,i}-x_i^*)^{2p-1}
\bigl(F_i(x_t)-F_i(x^*)\bigr)\right|\nonumber\\
&\quad\le \sum_{i=1}^n |x_{t,i}-x_i^*|^{2p-1}
|F_i(x_t)-F_i(x^*)|\nonumber\\
&\quad\le
\left(\sum_{i=1}^n
\left(|x_{t,i}-x_i^*|^{2p-1}\right)^{\frac{2p}{2p-1}}
\right)^{\frac{2p-1}{2p}}\nonumber\\
&\qquad\times
\left(\sum_{i=1}^n
\bigl(F_i(x_t)-F_i(x^*)\bigr)^{2p}
\right)^{\frac{1}{2p}}\nonumber\\
&\quad=
\left(\sum_{i=1}^n(x_{t,i}-x_i^*)^{2p}\right)^{\frac{2p-1}{2p}}
\left(\sum_{i=1}^n
\bigl(F_i(x_t)-F_i(x^*)\bigr)^{2p}
\right)^{\frac{1}{2p}}\nonumber\\
&\quad=\|x_t-x^*\|_{2p}^{2p-1}
\|F(x_t)-F(x^*)\|_{2p}.
\label{eq:app:holder-contract}
\end{align}
The finite-dimensional norm comparison in~\cref{lemma:2} and the contraction assumption further give
\begin{align}
\|F(x_t)-F(x^*)\|_{2p}
&\le n^{1/(2p)}\|F(x_t)-F(x^*)\|_\infty\nonumber\\
&\le \alpha n^{1/(2p)}\|x_t-x^*\|_\infty\nonumber\\
&\le \alpha n^{1/(2p)}\|x_t-x^*\|_{2p}.
\label{eq:app:F-contract-2p}
\end{align}
Using~\eqref{eq:app:holder-contract} in~\eqref{eq:app:Wp-split-contract}, and then using~\eqref{eq:app:F-contract-2p}, yields the complete chain
\begin{align}
\frac{d}{dt}W_p(x_t-x^*)
&\le
\left|\sum_{i=1}^n (x_{t,i}-x_i^*)^{2p-1}
\bigl(F_i(x_t)-F_i(x^*)\bigr)\right|
-\sum_{i=1}^n(x_{t,i}-x_i^*)^{2p}\nonumber\\
&\le \|x_t-x^*\|_{2p}^{2p-1}
\|F(x_t)-F(x^*)\|_{2p}
-\|x_t-x^*\|_{2p}^{2p}\nonumber\\
&\le \|x_t-x^*\|_{2p}^{2p-1}
\left(\alpha n^{1/(2p)}\|x_t-x^*\|_{2p}\right)
-\|x_t-x^*\|_{2p}^{2p}\nonumber\\
&=\alpha n^{1/(2p)}\|x_t-x^*\|_{2p}^{2p}
-\|x_t-x^*\|_{2p}^{2p}\nonumber\\
&=-\left(1-\alpha n^{1/(2p)}\right)
\|x_t-x^*\|_{2p}^{2p}.
\label{eq:app:Wp-contract-core}
\end{align}
This proves the first inequality in~\eqref{eq:Wp-contract-derivative}, which is repeated here:
\begin{align*}
\frac{d}{dt}W_p(x_t-x^*)
\le-\left(1-\alpha n^{1/(2p)}\right)\|x_t-x^*\|_{2p}^{2p}.
\end{align*}

The upper estimate in~\cref{lemma:app:Wp-two-sided}, applied with $z=x_t-x^*$, gives
\begin{align*}
W_p(x_t-x^*)
\le \frac{w_{\max}}{2p}\|x_t-x^*\|_{2p}^{2p}.
\end{align*}
Equivalently,
\begin{align}
\|x_t-x^*\|_{2p}^{2p}
\ge \frac{2p}{w_{\max}}W_p(x_t-x^*).
\label{eq:app:Wp-unweighted-lower}
\end{align}
Moreover, the stated condition on $p$ implies
\begin{align}
2p&>\frac{\ln n}{\ln(\alpha^{-1})},
&\frac{\ln n}{2p}&<\ln(\alpha^{-1}),
&\alpha n^{1/(2p)}
&=\exp\!\left(\ln\alpha+\frac{\ln n}{2p}\right)<1.
\label{eq:app:contract-margin}
\end{align}
Thus, $1-\alpha n^{1/(2p)}>0$. Because this coefficient is positive, using~\eqref{eq:app:Wp-unweighted-lower} in the negative term of~\eqref{eq:app:Wp-contract-core} gives
\begin{align}
\frac{d}{dt}W_p(x_t-x^*)
&\le-\left(1-\alpha n^{1/(2p)}\right)
\|x_t-x^*\|_{2p}^{2p}\nonumber\\
&\le-\left(1-\alpha n^{1/(2p)}\right)
\frac{2p}{w_{\max}}W_p(x_t-x^*)\nonumber\\
&=-\frac{2p\left(1-\alpha n^{1/(2p)}\right)}{w_{\max}}
W_p(x_t-x^*).
\label{eq:app:Wp-linear-decay}
\end{align}
This proves the second inequality in~\eqref{eq:Wp-contract-derivative}, which is repeated here:
\begin{align*}
\frac{d}{dt}W_p(x_t-x^*)
\le-\frac{2p\left(1-\alpha n^{1/(2p)}\right)}{w_{\max}}
W_p(x_t-x^*).
\end{align*}
\end{proof}

\begin{framed}
\begin{lemma}\label{lemma:app:thm1-integration}
Under the assumptions of \cref{thm:stability2}, let $w_{\min}:=\min_i w_i$. Then
\begin{align*}
W_p(x_t-x^*)
&\le \exp\!\left(-\frac{2p(1-\alpha n^{1/(2p)})}{w_{\max}}t\right)
W_p(x_0-x^*),
\qquad t\ge0,
\end{align*}
and
\begin{align*}
\|x_t-x^*\|_{2p}
&\le\left(\frac{w_{\max}}{w_{\min}}\right)^{1/(2p)}
\exp\!\left(-\frac{1-\alpha n^{1/(2p)}}{w_{\max}}t\right)
\|x_0-x^*\|_{2p},
\qquad t\ge0.
\end{align*}
\end{lemma}
\end{framed}
\begin{proof}
By~\eqref{eq:app:contract-margin},
\begin{align}
\frac{2p\left(1-\alpha n^{1/(2p)}\right)}{w_{\max}}>0.
\label{eq:app:Wp-gronwall-rate}
\end{align}
Since $W_p(x_t-x^*)$ is nonnegative and continuously differentiable,
\eqref{eq:app:Wp-linear-decay} can be written as
\begin{align}
\frac{d}{dt}W_p(x_t-x^*)
\le -\frac{2p\left(1-\alpha n^{1/(2p)}\right)}{w_{\max}}W_p(x_t-x^*).
\label{eq:app:Wp-gronwall-inequality}
\end{align}
Applying Gr\"{o}nwall's inequality to~\eqref{eq:app:Wp-gronwall-inequality} with zero additive term gives
\begin{align}
W_p(x_t-x^*)
\le \exp\!\left(-\frac{2p\left(1-\alpha n^{1/(2p)}\right)}{w_{\max}}t\right)W_p(x_0-x^*),
\qquad t\ge0.
\label{eq:app:Wp-gronwall-bound}
\end{align}
This is the main-text estimate in~\eqref{eq:Wp-contract-bound}.
Applying the lower estimate in~\cref{lemma:app:Wp-two-sided} at time $t$, the bound in~\eqref{eq:app:Wp-gronwall-bound}, and the upper estimate in~\cref{lemma:app:Wp-two-sided} at time $0$ gives
\begin{align}
\frac{w_{\min}}{2p}\|x_t-x^*\|_{2p}^{2p}
&\le W_p(x_t-x^*)\nonumber\\
&\le \exp\!\left(-\frac{2p(1-\alpha n^{1/(2p)})}{w_{\max}}t\right)
W_p(x_0-x^*)\nonumber\\
&\le \exp\!\left(-\frac{2p(1-\alpha n^{1/(2p)})}{w_{\max}}t\right)
\frac{w_{\max}}{2p}\|x_0-x^*\|_{2p}^{2p}.
\label{eq:app:Wp-state-chain}
\end{align}
Multiplying~\eqref{eq:app:Wp-state-chain} by $2p/w_{\min}$ and taking the positive $2p$th root yields
\begin{align}
\|x_t-x^*\|_{2p}
\le\left(\frac{w_{\max}}{w_{\min}}\right)^{1/(2p)}
\exp\!\left(-\frac{1-\alpha n^{1/(2p)}}{w_{\max}}t\right)
\|x_0-x^*\|_{2p}.
\label{eq:app:state-contract-bound}
\end{align}
Because $D$ is nonsingular, every equilibrium of~\eqref{eq:ODE1} satisfies $F(x)=x$. The contraction property gives the unique fixed point $x^*$, and~\eqref{eq:app:state-contract-bound} proves global exponential stability.
\end{proof}

\begin{proof}[Proof of \Cref{thm:stability2}]
The two-sided polynomial comparison is proved in~\cref{lemma:app:Wp-two-sided}. The derivative inequalities stated in the theorem follow from~\cref{lemma:app:thm1-derivative}, and the exponential estimates follow from~\cref{lemma:app:thm1-integration}. Since $D$ is nonsingular, every equilibrium of the ODE is a fixed point of $F$, and the contraction property makes $x^*$ the unique equilibrium. This proves the theorem.
\end{proof}

\section{Proof of Theorem 2}\label{sec:app:thm2-proof}

\begin{framed}
\noindent\textbf{Restatement of \Cref{thm:stability4} (Polynomial perturbation certificate).}\par\medskip
Suppose that $F:{\mathbb R}^n\to{\mathbb R}^n$ is globally Lipschitz continuous, and let $x_t$ be the unique global solution of~\eqref{eq:ODE1}. For convenience, the equation referred to as~\eqref{eq:ODE1} is repeated here:
\begin{align*}
\frac{d}{dt}x_t=D\bigl(F(x_t)-x_t\bigr),
\qquad t\ge0,
\qquad x_0\in{\mathbb R}^n.
\end{align*}
Suppose that $F':{\mathbb R}^n\to{\mathbb R}^n$ is an $\infty$-norm contraction with factor $\alpha\in(0,1)$ and fixed point $x^*$:
\[
\|F'(x)-F'(y)\|_\infty\le \alpha\|x-y\|_\infty,
\qquad x,y\in{\mathbb R}^n.
\]
Suppose also that $F$ satisfies
\[
\|F(x)-F'(x)\|_\infty\le\eta,
\qquad x\in{\mathbb R}^n,
\]
for some $\eta>0$. Choose $w_i=1/d_i$, and let
$w_{\min}:=\min_i w_i$ and $w_{\max}:=\max_i w_i$. For any $p\in\mathbb N$ satisfying
\[
\left\lceil \frac{\ln n}{\ln(\alpha^{-1})}\right\rceil<2p,
\]
we have
\begin{align*}
\frac{d}{dt}W_p(x_t-x^*)
\le{}&-\left(1-\alpha n^{1/(2p)}\right)\|x_t-x^*\|_{2p}^{2p}
+n^{1/(2p)}\eta\|x_t-x^*\|_{2p}^{2p-1}.
\end{align*}
Moreover,
\begin{align*}
\|x_t-x^*\|_{2p,w}
\le{}&\|x_0-x^*\|_{2p,w}
\exp\!\left(\frac{\alpha n^{1/(2p)}-1}{w_{\max}}t\right)\\
&+\frac{n^{1/(2p)}\eta w_{\max}}
{\left(1-\alpha n^{1/(2p)}\right)w_{\min}^{(2p-1)/(2p)}},
\qquad t\ge0.
\end{align*}
In particular, $W_p(x_t-x^*)$ decreases whenever
$\|x_t-x^*\|_{2p}>n^{1/(2p)}\eta/(1-\alpha n^{1/(2p)})$.
\end{framed}

\begin{framed}
\begin{lemma}[Derivative estimate for \Cref{thm:stability4}]\label{lemma:app:thm2-derivative}
Under the assumptions of \cref{thm:stability4},
\begin{align*}
\frac{d}{dt}W_p(x_t-x^*)
\le{}&-\left(1-\alpha n^{1/(2p)}\right)\|x_t-x^*\|_{2p}^{2p}
+n^{1/(2p)}\eta\|x_t-x^*\|_{2p}^{2p-1}.
\end{align*}
\end{lemma}
\end{framed}
\begin{proof}
For the componentwise calculations, $F_i(x)$ and $F_i'(x)$ denote the $i$th components of $F(x)$ and $F'(x)$, respectively. The coordinate derivative in~\eqref{eq:app:Wp-gradient}, the chain rule, and the ODE repeated in the theorem restatement give
\begin{align}
\frac{d}{dt}W_p(x_t-x^*)
&=\sum_{i=1}^n w_i(x_{t,i}-x_i^*)^{2p-1}\dot x_{t,i}\nonumber\\
&=\sum_{i=1}^n (x_{t,i}-x_i^*)^{2p-1}
\bigl(F_i(x_t)-x_{t,i}\bigr),
\label{eq:app:Wp-chain-perturb}
\end{align}
where $w_i d_i=1$ was used in the second equality. Since $F'(x^*)=x^*$,
\begin{align}
F_i(x_t)-x_{t,i}
={}&F_i'(x_t)-F_i'(x^*)
+F_i(x_t)-F_i'(x_t)
-(x_{t,i}-x_i^*).
\label{eq:app:perturb-decomposition}
\end{align}
Substituting~\eqref{eq:app:perturb-decomposition} into~\eqref{eq:app:Wp-chain-perturb} yields
\begin{align}
\frac{d}{dt}W_p(x_t-x^*)
={}&\sum_{i=1}^n (x_{t,i}-x_i^*)^{2p-1}
\bigl(F_i'(x_t)-F_i'(x^*)\bigr)\nonumber\\
&+\sum_{i=1}^n (x_{t,i}-x_i^*)^{2p-1}
\bigl(F_i(x_t)-F_i'(x_t)\bigr)\nonumber\\
&-\sum_{i=1}^n(x_{t,i}-x_i^*)^{2p}.
\label{eq:app:Wp-split-perturb}
\end{align}
Applying H{\"o}lder's inequality to the first sum in~\eqref{eq:app:Wp-split-perturb} gives
\begin{align}
&\left|\sum_{i=1}^n (x_{t,i}-x_i^*)^{2p-1}
\bigl(F_i'(x_t)-F_i'(x^*)\bigr)\right|\nonumber\\
&\quad\le\|x_t-x^*\|_{2p}^{2p-1}
\|F'(x_t)-F'(x^*)\|_{2p}.
\label{eq:app:holder-perturb-contract}
\end{align}
The same conjugate exponents applied to the second sum in~\eqref{eq:app:Wp-split-perturb} give
\begin{align}
&\left|\sum_{i=1}^n (x_{t,i}-x_i^*)^{2p-1}
\bigl(F_i(x_t)-F_i'(x_t)\bigr)\right|\nonumber\\
&\quad\le\|x_t-x^*\|_{2p}^{2p-1}
\|F(x_t)-F'(x_t)\|_{2p}.
\label{eq:app:holder-perturb-error}
\end{align}
Using~\eqref{eq:app:holder-perturb-contract} and~\eqref{eq:app:holder-perturb-error} in~\eqref{eq:app:Wp-split-perturb} yields
\begin{align}
\frac{d}{dt}W_p(x_t-x^*)
\le{}&\|x_t-x^*\|_{2p}^{2p-1}
\|F'(x_t)-F'(x^*)\|_{2p}\nonumber\\
&+\|x_t-x^*\|_{2p}^{2p-1}
\|F(x_t)-F'(x_t)\|_{2p}
-\|x_t-x^*\|_{2p}^{2p}.
\label{eq:app:Wp-perturb-raw}
\end{align}
The contraction assumption and~\cref{lemma:2} imply
\begin{align}
\|F'(x_t)-F'(x^*)\|_{2p}
&\le n^{1/(2p)}\|F'(x_t)-F'(x^*)\|_\infty\nonumber\\
&\le \alpha n^{1/(2p)}\|x_t-x^*\|_\infty\nonumber\\
&\le \alpha n^{1/(2p)}\|x_t-x^*\|_{2p},
\label{eq:app:perturb-contraction-bound}
\end{align}
whereas the uniform perturbation bound gives
\begin{align}
\|F(x_t)-F'(x_t)\|_{2p}
&\le n^{1/(2p)}\|F(x_t)-F'(x_t)\|_\infty\nonumber\\
&\le n^{1/(2p)}\eta.
\label{eq:app:perturb-error-bound}
\end{align}
Substituting~\eqref{eq:app:perturb-contraction-bound} and~\eqref{eq:app:perturb-error-bound} into~\eqref{eq:app:Wp-perturb-raw} yields
\begin{align}
\frac{d}{dt}W_p(x_t-x^*)
\le{}&-\left(1-\alpha n^{1/(2p)}\right)\|x_t-x^*\|_{2p}^{2p}
+n^{1/(2p)}\eta\|x_t-x^*\|_{2p}^{2p-1}.
\label{eq:app:Wp-perturb-final}
\end{align}
Equation~\eqref{eq:app:Wp-perturb-final} proves the main-text estimate in~\eqref{eq:Wp-perturb-derivative}, which is repeated here:
\begin{align*}
\frac{d}{dt}W_p(x_t-x^*)
\le{}&-\left(1-\alpha n^{1/(2p)}\right)\|x_t-x^*\|_{2p}^{2p}
+n^{1/(2p)}\eta\|x_t-x^*\|_{2p}^{2p-1}.
\end{align*}
\end{proof}
{\color{blue}******}
\begin{framed}
\begin{lemma}[Scalar comparison for \Cref{thm:stability4}]\label{lemma:app:thm2-comparison}
Under the assumptions of \cref{thm:stability4},
\begin{align*}
\|x_t-x^*\|_{2p,w}
\le{}&\|x_0-x^*\|_{2p,w}
\exp\!\left(-\frac{1-\alpha n^{1/(2p)}}{w_{\max}}t\right)\\
&+\frac{n^{1/(2p)}\eta w_{\max}}
{\left(1-\alpha n^{1/(2p)}\right)w_{\min}^{(2p-1)/(2p)}},
\qquad t\ge0.
\end{align*}
Moreover, $W_p(x_t-x^*)$ is strictly decreasing whenever
$\|x_t-x^*\|_{2p}>n^{1/(2p)}\eta/(1-\alpha n^{1/(2p)})$.
\end{lemma}
\end{framed}
\begin{proof}
As shown in~\eqref{eq:app:contract-margin}, the condition on $p$ ensures that
$1-\alpha n^{1/(2p)}>0$. For $x_t\ne x^*$, the identity
\begin{align}
W_p(x_t-x^*)=\frac{1}{2p}\|x_t-x^*\|_{2p,w}^{2p}
\label{eq:app:Wp-norm-identity}
\end{align}
gives, by differentiation,
\begin{align}
\frac{d}{dt}W_p(x_t-x^*)
=\|x_t-x^*\|_{2p,w}^{2p-1}
\frac{d}{dt}\|x_t-x^*\|_{2p,w}.
\label{eq:app:norm-derivative-identity}
\end{align}
The weighted and unweighted norms satisfy
\begin{align}
\|x_t-x^*\|_{2p}^{2p}
&\ge \frac{1}{w_{\max}}\|x_t-x^*\|_{2p,w}^{2p},\nonumber\\
\|x_t-x^*\|_{2p}^{2p-1}
&\le \frac{1}{w_{\min}^{(2p-1)/(2p)}}
\|x_t-x^*\|_{2p,w}^{2p-1}.
\label{eq:app:weighted-power-comparison}
\end{align}
Using~\eqref{eq:app:weighted-power-comparison} in~\eqref{eq:app:Wp-perturb-final} and dividing according to~\eqref{eq:app:norm-derivative-identity} gives
\begin{align}
\frac{d}{dt}\|x_t-x^*\|_{2p,w}
\le{}&-\frac{1-\alpha n^{1/(2p)}}{w_{\max}}
\|x_t-x^*\|_{2p,w}
+\frac{n^{1/(2p)}\eta}{w_{\min}^{(2p-1)/(2p)}}.
\label{eq:app:perturb-scalar}
\end{align}
The norm trajectory is nonnegative and absolutely continuous on every finite time interval. At almost every time at which it is zero, its derivative is also zero; therefore,~\eqref{eq:app:perturb-scalar} holds almost everywhere. These properties justify applying Gr\"{o}nwall's inequality, including when the trajectory passes through $x^*$:
\begin{align}
\|x_t-x^*\|_{2p,w}
\le{}&\exp\!\left(-\frac{1-\alpha n^{1/(2p)}}{w_{\max}}t\right)
\|x_0-x^*\|_{2p,w}\nonumber\\
&+\frac{n^{1/(2p)}\eta}{w_{\min}^{(2p-1)/(2p)}}
\int_0^t\exp\!\left(-\frac{1-\alpha n^{1/(2p)}}{w_{\max}}(t-\tau)\right)\,d\tau\nonumber\\
={}&\exp\!\left(-\frac{1-\alpha n^{1/(2p)}}{w_{\max}}t\right)
\|x_0-x^*\|_{2p,w}\nonumber\\
&+\frac{n^{1/(2p)}\eta w_{\max}}
{\left(1-\alpha n^{1/(2p)}\right)w_{\min}^{(2p-1)/(2p)}}
\left[1-\exp\!\left(-\frac{1-\alpha n^{1/(2p)}}{w_{\max}}t\right)\right].
\label{eq:app:perturb-gronwall-bound}
\end{align}
Since the expression in square brackets in~\eqref{eq:app:perturb-gronwall-bound} lies in $[0,1]$, it follows that
\begin{align}
\|x_t-x^*\|_{2p,w}
\le{}&\exp\!\left(-\frac{1-\alpha n^{1/(2p)}}{w_{\max}}t\right)
\|x_0-x^*\|_{2p,w}\nonumber\\
&+\frac{n^{1/(2p)}\eta w_{\max}}
{\left(1-\alpha n^{1/(2p)}\right)w_{\min}^{(2p-1)/(2p)}},
\qquad t\ge0.
\label{eq:app:perturb-bound-repeated}
\end{align}
This is the main-text estimate in~\eqref{eq:Wp-perturb-bound}. Finally, factoring~\eqref{eq:app:Wp-perturb-final} gives
\begin{align}
\frac{d}{dt}W_p(x_t-x^*)
\le \|x_t-x^*\|_{2p}^{2p-1}
\left[-\left(1-\alpha n^{1/(2p)}\right)\|x_t-x^*\|_{2p}
+n^{1/(2p)}\eta\right].
\label{eq:app:perturb-factor}
\end{align}
The sign of the bracket in~\eqref{eq:app:perturb-factor} proves the stated strict-decrease condition.
\end{proof}

\begin{proof}[Proof of \Cref{thm:stability4}]
The derivative statement follows from~\cref{lemma:app:thm2-derivative}. The weighted-norm estimate and the strict-decrease condition follow from~\cref{lemma:app:thm2-comparison}. This proves the theorem.
\end{proof}

\section{Proof of Theorem 3}\label{sec:app:thm3-proof}

\begin{framed}
\noindent\textbf{Restatement of \Cref{thm:stability5}.}\par\medskip
Consider~\eqref{eq:ODE-Q-learning1} with
$H\in\{H_{\max},H_{\rm lse}^{\lambda},H_{\rm mm}^{\lambda}\}$, and let $Q_t$ be its solution. The referenced ODE is repeated here:
\begin{align*}
\frac{d}{dt}Q_t=DR+\gamma DPH(Q_t)-DQ_t,
\qquad t\ge0,
\qquad Q_0\in{\mathbb R}^{|{\cal S}\times{\cal A}|}.
\end{align*}
Let $Q_e$ denote the corresponding fixed point in~\eqref{eq:4}. The fixed-point relations in~\eqref{eq:4} are repeated here:
\begin{align*}
Q_{\max}^*&=F_{\max}(Q_{\max}^*),
&Q_{\rm lse}^{\lambda}&=F_{\rm lse}^{\lambda}(Q_{\rm lse}^{\lambda}),
&Q_{\rm mm}^{\lambda}&=F_{\rm mm}^{\lambda}(Q_{\rm mm}^{\lambda}).
\end{align*}
Set $n=|{\cal S}\times{\cal A}|$, choose $w_i=1/d_i$, and let $w_{\max}:=\max_i w_i$. For any $p\in\mathbb N$ satisfying
\[
\left\lceil\frac{\ln n}{\ln(\gamma^{-1})}\right\rceil<2p,
\]
we have
\begin{align*}
W_p(Q_t-Q_e)
\le \exp\!\left(-\frac{2p(1-\gamma n^{1/(2p)})}{w_{\max}}t\right)
W_p(Q_0-Q_e),\qquad t\ge0.
\end{align*}
Therefore, $Q_e$ is the unique globally exponentially stable equilibrium of the corresponding ODE.
\end{framed}

\begin{proof}[Proof of \Cref{thm:stability5}]
Fix $H\in\{H_{\max},H_{\rm lse}^{\lambda},H_{\rm mm}^{\lambda}\}$ and let $F$ be the corresponding Bellman operator in~\eqref{eq:ODE-Q-learning2}.
By \cref{lemma:contraction}, $F$ is a $\gamma$-contraction in the $\infty$-norm. Hence, the Banach fixed-point theorem guarantees the unique fixed point $Q_e$ specified in the theorem.

Moreover, the condition on $p$ implies
\[
\gamma n^{1/(2p)}<1.
\]
Therefore, applying \cref{thm:stability2} with
$x_t=Q_t$, $x^*=Q_e$, and $\alpha=\gamma$ gives
\begin{align}
W_p(Q_t-Q_e)
\le
\exp\!\left(
-\frac{2p\left(1-\gamma n^{1/(2p)}\right)}{w_{\max}}t
\right)
W_p(Q_0-Q_e),
\qquad t\ge0.
\end{align}
Since $D$ is nonsingular, the equilibria of~\eqref{eq:ODE-Q-learning2} coincide with the fixed points of $F$. Thus, $Q_e$ is the unique globally exponentially stable equilibrium.
\end{proof}

\section{Proof of Theorem 4}\label{sec:app:thm4-proof}

\begin{framed}
\noindent\textbf{Restatement of \Cref{thm:stability6}.}\par\medskip
Consider~\eqref{eq:ODE-Q-learning1} with $H=H_{\rm bz}^{\lambda}$, and let $Q_t$ be its solution. The referenced ODE is repeated here:
\begin{align*}
\frac{d}{dt}Q_t=DR+\gamma DPH_{\rm bz}^{\lambda}(Q_t)-DQ_t,
\qquad t\ge0,
\qquad Q_0\in{\mathbb R}^{|{\cal S}\times{\cal A}|}.
\end{align*}
Set $n=|{\cal S}\times{\cal A}|$, choose $w_i=1/d_i$, and let $w_{\min}:=\min_iw_i$ and $w_{\max}:=\max_iw_i$. For any $p\in\mathbb N$ satisfying
\[
\left\lceil\frac{\ln n}{\ln(\gamma^{-1})}\right\rceil<2p,
\]
we have
\begin{align*}
\frac{d}{dt}W_p(Q_t-Q_{\max}^*)
\le{}&-\left(1-\gamma n^{1/(2p)}\right)
\|Q_t-Q_{\max}^*\|_{2p}^{2p}\\
&+n^{1/(2p)}\gamma\lambda\ln(|\mathcal A|)
\|Q_t-Q_{\max}^*\|_{2p}^{2p-1},
\end{align*}
and
\begin{align*}
\|Q_t-Q_{\max}^*\|_{2p,w}
\le{}&\|Q_0-Q_{\max}^*\|_{2p,w}
\exp\!\left(\frac{\gamma n^{1/(2p)}-1}{w_{\max}}t\right)\\
&+\frac{n^{1/(2p)}w_{\max}\gamma\lambda\ln(|\mathcal A|)}
{\left(1-\gamma n^{1/(2p)}\right)w_{\min}^{(2p-1)/(2p)}},
\qquad t\ge0.
\end{align*}
\end{framed}
\begin{proof}[Proof of \Cref{thm:stability6}]
If $|{\cal A}|=1$, then $H_{\rm bz}^{\lambda}=H_{\max}$ and $F_{\rm bz}^{\lambda}=F_{\max}$. The derivative estimate follows from~\cref{thm:stability2} with $\alpha=\gamma$. Taking the positive $2p$th root of~\eqref{eq:Wp-contract-bound} and using~\eqref{eq:app:Wp-norm-identity} gives the stated weighted-norm bound with zero additive term. Thus, it remains to consider $|{\cal A}|\ge2$, for which $\gamma\lambda\ln(|{\cal A}|)>0$.

By \cref{lemma:Lipschitz1}, $F_{\rm bz}^{\lambda}$ is globally Lipschitz continuous. By \cref{lemma:contraction2}, the Boltzmann and max Bellman operators satisfy
\begin{align}
\|F_{\rm bz}^{\lambda}(Q)-F_{\max}(Q)\|_\infty
&\le\gamma\lambda\ln(|{\cal A}|),
\qquad Q\in{\mathbb R}^{n}.
\label{eq:app:thm4-uniform-gap}
\end{align}
Moreover, \cref{lemma:contraction} gives
\begin{align}
\|F_{\max}(Q)-F_{\max}(Q')\|_\infty
&\le\gamma\|Q-Q'\|_\infty,
\qquad Q,Q'\in{\mathbb R}^{n},
\label{eq:app:thm4-max-contraction}
\end{align}
and $F_{\max}(Q_{\max}^*)=Q_{\max}^*$. Therefore,
\cref{thm:stability4} applies with
\begin{align}
F'&=F_{\max}, &F&=F_{\rm bz}^{\lambda},
&x_t&=Q_t, &x^*&=Q_{\max}^*,\nonumber\\
\alpha&=\gamma, &\eta&=\gamma\lambda\ln(|{\cal A}|).
\label{eq:app:thm4-substitution}
\end{align}
The general derivative estimate in~\eqref{eq:Wp-perturb-derivative} is repeated here:
\begin{align*}
\frac{d}{dt}W_p(x_t-x^*)
\le{}&-\left(1-\alpha n^{1/(2p)}\right)\|x_t-x^*\|_{2p}^{2p}
+n^{1/(2p)}\eta\|x_t-x^*\|_{2p}^{2p-1}.
\end{align*}
Substituting~\eqref{eq:app:thm4-substitution} into this repeated estimate gives
\begin{align}
\frac{d}{dt}W_p(Q_t-Q_{\max}^*)
\le{}&-\left(1-\gamma n^{1/(2p)}\right)\|Q_t-Q_{\max}^*\|_{2p}^{2p}\nonumber\\
&+n^{1/(2p)}\gamma\lambda\ln(|{\cal A}|)
\|Q_t-Q_{\max}^*\|_{2p}^{2p-1}.
\label{eq:app:thm4-polynomial-derivative}
\end{align}
This proves the main-text estimate in~\eqref{eq:Wp-Q-boltzmann-derivative}. To obtain the scalar differential inequality explicitly, first use~\eqref{eq:Wp-definition} at any time such that $Q_t\ne Q_{\max}^*$:
\begin{align}
\frac{d}{dt}W_p(Q_t-Q_{\max}^*)
&=\frac{1}{2p}\frac{d}{dt}\|Q_t-Q_{\max}^*\|_{2p,w}^{2p}\nonumber\\
&=\|Q_t-Q_{\max}^*\|_{2p,w}^{2p-1}
\frac{d}{dt}\|Q_t-Q_{\max}^*\|_{2p,w}.
\label{eq:app:thm4-norm-identity}
\end{align}
The norm comparisons in~\cref{lemma:2} give
\begin{align}
\|Q_t-Q_{\max}^*\|_{2p}^{2p}
&\ge\frac{1}{w_{\max}}\|Q_t-Q_{\max}^*\|_{2p,w}^{2p},\nonumber\\
\|Q_t-Q_{\max}^*\|_{2p}^{2p-1}
&\le\frac{1}{w_{\min}^{(2p-1)/(2p)}}
\|Q_t-Q_{\max}^*\|_{2p,w}^{2p-1}.
\label{eq:app:thm4-weighted-comparison}
\end{align}
Because $1-\gamma n^{1/(2p)}>0$, the lower bound in~\eqref{eq:app:thm4-weighted-comparison} applies to the negative term of~\eqref{eq:app:thm4-polynomial-derivative}, whereas the upper bound applies to its nonnegative term. Dividing by the positive factor in~\eqref{eq:app:thm4-norm-identity} therefore yields
\begin{align}
\frac{d}{dt}\|Q_t-Q_{\max}^*\|_{2p,w}
={}&\frac{\frac{d}{dt}W_p(Q_t-Q_{\max}^*)}
{\|Q_t-Q_{\max}^*\|_{2p,w}^{2p-1}}\nonumber\\
\le{}&-\left(1-\gamma n^{1/(2p)}\right)
\frac{\|Q_t-Q_{\max}^*\|_{2p}^{2p}}
{\|Q_t-Q_{\max}^*\|_{2p,w}^{2p-1}}\nonumber\\
&+n^{1/(2p)}\gamma\lambda\ln(|{\cal A}|)
\frac{\|Q_t-Q_{\max}^*\|_{2p}^{2p-1}}
{\|Q_t-Q_{\max}^*\|_{2p,w}^{2p-1}}\nonumber\\
\le{}&-\frac{1-\gamma n^{1/(2p)}}{w_{\max}}
\frac{\|Q_t-Q_{\max}^*\|_{2p,w}^{2p}}
{\|Q_t-Q_{\max}^*\|_{2p,w}^{2p-1}}\nonumber\\
&+\frac{n^{1/(2p)}\gamma\lambda\ln(|{\cal A}|)}
{w_{\min}^{(2p-1)/(2p)}}
\frac{\|Q_t-Q_{\max}^*\|_{2p,w}^{2p-1}}
{\|Q_t-Q_{\max}^*\|_{2p,w}^{2p-1}}\nonumber\\
={}&-\frac{1-\gamma n^{1/(2p)}}{w_{\max}}
\|Q_t-Q_{\max}^*\|_{2p,w}
+\frac{n^{1/(2p)}\gamma\lambda\ln(|{\cal A}|)}
{w_{\min}^{(2p-1)/(2p)}}.
\label{eq:app:thm4-scalar-comparison}
\end{align}
The norm trajectory is nonnegative and absolutely continuous on every finite time interval. Its derivative is zero at almost every time at which the norm is zero, so~\eqref{eq:app:thm4-scalar-comparison} holds almost everywhere. This justifies the application of Gr\"{o}nwall's inequality below even when $Q_t=Q_{\max}^*$ at some time. Set
\begin{align}
a_p:=\frac{1-\gamma n^{1/(2p)}}{w_{\max}}>0.
\label{eq:app:thm4-gronwall-rate}
\end{align}
Applying Gr\"{o}nwall's inequality to~\eqref{eq:app:thm4-scalar-comparison} gives
\begin{align}
\|Q_t-Q_{\max}^*\|_{2p,w}
\le{}&e^{-a_pt}\|Q_0-Q_{\max}^*\|_{2p,w}\nonumber\\
&+\frac{n^{1/(2p)}\gamma\lambda\ln(|{\cal A}|)}
{a_p w_{\min}^{(2p-1)/(2p)}},
\qquad t\ge0.
\label{eq:app:thm4-gronwall-bound}
\end{align}
Substituting~\eqref{eq:app:thm4-gronwall-rate} into~\eqref{eq:app:thm4-gronwall-bound} yields
\begin{align}
\|Q_t-Q_{\max}^*\|_{2p,w}
\le{}&\exp\!\left(-\frac{1-\gamma n^{1/(2p)}}{w_{\max}}t\right)
\|Q_0-Q_{\max}^*\|_{2p,w}\nonumber\\
&+\frac{n^{1/(2p)}w_{\max}\gamma\lambda\ln(|{\cal A}|)}
{\left(1-\gamma n^{1/(2p)}\right)w_{\min}^{(2p-1)/(2p)}},
\qquad t\ge0.
\label{eq:app:thm4-bound-repeated}
\end{align}
This is the main-text estimate in~\eqref{eq:Wp-Q-boltzmann-bound}. In addition,
\eqref{eq:app:thm4-polynomial-derivative} shows that $W_p(Q_t-Q_{\max}^*)$ is strictly
decreasing whenever
\begin{align*}
\|Q_t-Q_{\max}^*\|_{2p}>
\frac{n^{1/(2p)}\gamma\lambda\ln(|{\cal A}|)}
{1-\gamma n^{1/(2p)}}.
\end{align*}
\end{proof}

\section{Basics of nonlinear system theory}
We use several standard notions from nonlinear-system theory in the ODE analysis. Consider the system
\begin{align}
\frac{d}{dt}x_t=f(x_t),\quad t\ge 0,\quad x_0 \in {\mathbb R}^n,\label{eq:nonlinear-system}
\end{align}
where $x_t\in {\mathbb R}^n$ is the state at time $t\ge 0$ and $f:{\mathbb R}^n \to {\mathbb R}^n$ is a nonlinear mapping.
For simplicity, we assume that the solution to~\eqref{eq:nonlinear-system} exists and is unique. This holds if $f$ is globally Lipschitz continuous.
\begin{framed}
\begin{lemma}[{\citep[Theorem~3.2]{khalil2002nonlinear}}]\label{lemma:existence}
Consider the nonlinear system~\eqref{eq:nonlinear-system} and assume that $f$ is globally Lipschitz continuous, i.e.,
\begin{align}
\|f(x)-f(y)\|\le L \|x-y\|,\;\forall x,y \in {\mathbb R}^n,
\end{align}
for some real number $L>0$ and norm $\|\cdot\|$. Then, the system has a unique solution $x_t$ for every $t\geq 0$ and every $x_0 \in {\mathbb R}^n$.
\end{lemma}
\end{framed}
A central concept in nonlinear-system analysis is an equilibrium point. A point $x=x^e$ in the state space is said to be an equilibrium point of~\eqref{eq:nonlinear-system} if it has the property that whenever the state of the system starts at $x^e$, it will remain at $x^e$~\citep{khalil2002nonlinear}. For~\eqref{eq:nonlinear-system}, the equilibrium points are the real roots of the equation $f(x)=0$. The equilibrium point $x^e$ is said to be globally asymptotically stable if it is Lyapunov stable and $x_t \to x^e$ as $t \to \infty$ for every initial state $x_0 \in {\mathbb R}^n$. Here, Lyapunov stability means that, for every $\varepsilon>0$, there exists $\delta>0$ such that $\|x_0-x^e\|<\delta$ implies $\|x_t-x^e\|<\varepsilon$ for all $t\ge0$. The equilibrium point is said to be globally exponentially stable if there exist a norm $\|\cdot\|$ and constants $C\ge1$ and $r>0$, independent of the initial state $x_0$, such that
\[
\|x_t-x^e\|\le C e^{-rt}\|x_0-x^e\|,
\qquad t\ge0,\qquad x_0\in{\mathbb R}^n.
\]

\section{Supporting lemmas}\label{sec:app:lemmas1}
This section collects the supporting lemmas used in the subsequent analysis.
First, the classical differential form of Gr\"{o}nwall's inequality is stated as follows.
\begin{framed}
\begin{lemma}[{\citep{gronwall1919note}}]\label{lemma:Gronwall}
Let $\psi _t: [0,\infty) \to [0,\infty)$ be an absolutely continuous function satisfying the differential inequality
\[\frac{d}{{dt}}{\psi _t} \le  - a{\psi _t} + b,\quad \text{for almost every }t\ge 0\]
where $a>0, b\ge 0$. Then,
\[{\psi _t} \le {\psi _0}{e^{ - at}} + \frac{b}{a},\quad \forall t\ge 0.\]
\end{lemma}
\end{framed}
The inequality in~\cref{lemma:Gronwall} is used repeatedly to establish stability of the ODEs considered below.
The following relations are essential for analyzing the smooth Q-learning algorithms.
Although proofs of these relations appear in~\cite{pan2020reinforcement} and related studies, we include them here for completeness and convenience.
In particular, the proof of the bound for the Boltzmann softmax operator is simpler than the proof in~\cite{pan2020reinforcement}.
\begin{framed}
\begin{lemma}\label{lemma:5}
For any $x=(x_1,\ldots,x_n)\in\mathbb R^n$ and any $\lambda>0$, we have
\begin{enumerate}
\item $h_{\max}(x)\le h_{\rm lse}^{\lambda}(x)
\le h_{\max}(x)+\lambda\ln n$;
\item $h_{\max}(x)-\lambda\ln n\le h_{\rm mm}^{\lambda}(x)
\le h_{\max}(x)$;
\item $h_{\max}(x)-\lambda\ln n\le h_{\rm bz}^{\lambda}(x)
\le h_{\max}(x)$;
\item $h_{\rm bz}^{\lambda}(x)\le h_{\rm lse}^{\lambda}(x)
\le h_{\rm bz}^{\lambda}(x)+\lambda\ln n$.
\end{enumerate}
\end{lemma}
\end{framed}
\begin{proof}
Since $x_i\le h_{\max}(x)$ for every $i$ and equality holds for at least one coordinate,
\begin{align}
e^{h_{\max}(x)/\lambda}
\le \sum_{i=1}^n e^{x_i/\lambda}
\le n e^{h_{\max}(x)/\lambda}.
\label{eq:app:lse-sum-bounds}
\end{align}
Taking logarithms in~\eqref{eq:app:lse-sum-bounds} and multiplying by $\lambda$ gives
\begin{align}
h_{\max}(x)
\le \lambda\ln\!\left(\sum_{i=1}^n e^{x_i/\lambda}\right)
\le h_{\max}(x)+\lambda\ln n,
\label{eq:app:lse-max-bounds}
\end{align}
which proves the first statement. Dividing~\eqref{eq:app:lse-sum-bounds} by $n$ before taking logarithms gives
\begin{align}
h_{\max}(x)-\lambda\ln n
\le \lambda\ln\!\left(\frac{1}{n}\sum_{i=1}^n e^{x_i/\lambda}\right)
\le h_{\max}(x),
\label{eq:app:mm-max-bounds}
\end{align}
which proves the second statement.

For the Boltzmann operator, define
\begin{align}
Z&:=\sum_{j=1}^n e^{x_j/\lambda},
&\pi_i&:=\frac{e^{x_i/\lambda}}{Z},
\qquad i\in\{1,\ldots,n\}.
\label{eq:app:boltzmann-weights}
\end{align}
Then $\pi_i\ge0$, $\sum_i\pi_i=1$, and
\begin{align}
h_{\rm bz}^{\lambda}(x)
=\sum_{i=1}^n\pi_i x_i
\le h_{\max}(x).
\label{eq:app:bz-upper-max}
\end{align}
For the lower bound, we use the convexity of $h_{\rm lse}^{\lambda}$ and its gradient formula~\citep{gao2017properties}. In terms of the weights in~\eqref{eq:app:boltzmann-weights},
\begin{align}
\nabla h_{\rm lse}^{\lambda}(x)^\top x
=\sum_{i=1}^n\pi_i x_i
=h_{\rm bz}^{\lambda}(x).
\label{eq:app:lse-gradient-bz}
\end{align}
The first-order inequality for a differentiable convex function gives
\begin{align}
h_{\rm lse}^{\lambda}(y)
\ge h_{\rm lse}^{\lambda}(x)
+\nabla h_{\rm lse}^{\lambda}(x)^\top(y-x),
\qquad x,y\in\mathbb R^n.
\label{eq:app:lse-convexity}
\end{align}
Setting $y=0$ and using~\eqref{eq:app:lse-gradient-bz} yields
\begin{align}
\lambda\ln n
=h_{\rm lse}^{\lambda}(0)
\ge h_{\rm lse}^{\lambda}(x)-h_{\rm bz}^{\lambda}(x).
\label{eq:app:lse-bz-convex-bound}
\end{align}
Together with the lower bound in~\eqref{eq:app:lse-max-bounds}, this implies
\begin{align}
h_{\max}(x)
\le h_{\rm lse}^{\lambda}(x)
\le h_{\rm bz}^{\lambda}(x)+\lambda\ln n,
\end{align}
so $h_{\rm bz}^{\lambda}(x)\ge h_{\max}(x)-\lambda\ln n$. Together with~\eqref{eq:app:bz-upper-max}, this proves the third statement.
Finally, the fourth statement follows from~\eqref{eq:app:lse-bz-convex-bound} and the inequalities $h_{\rm bz}^{\lambda}(x)\le h_{\max}(x)\le h_{\rm lse}^{\lambda}(x)$ already proved above.
\end{proof}
Furthermore, the following lemma will be used in the analysis.
\begin{framed}
\begin{lemma}\label{lemma:4}
For every $\lambda>0$,
\[
\frac{h_{\max}(cx)}{c},\qquad
\frac{h_{\rm lse}^{\lambda}(cx)}{c},\qquad
\frac{h_{\rm mm}^{\lambda}(cx)}{c},\qquad
\frac{h_{\rm bz}^{\lambda}(cx)}{c}
\]
all converge to $h_{\max}(x)$ as $c\to\infty$, uniformly on every compact subset of ${\mathbb R}^n$.
\end{lemma}
\end{framed}
\begin{proof}
The max operator is positively homogeneous, so
\begin{align}
\frac{h_{\max}(cx)}{c}=h_{\max}(x),
\qquad c>0.
\label{eq:app:max-scaling-exact}
\end{align}
Applying~\cref{lemma:5} to $cx$ and dividing each inequality by $c$ gives
\begin{align}
0
&\le \frac{h_{\rm lse}^{\lambda}(cx)}{c}-h_{\max}(x)
\le \frac{\lambda\ln n}{c},
\label{eq:app:lse-scaling-gap}\\
0
&\le h_{\max}(x)-\frac{h_{\rm mm}^{\lambda}(cx)}{c}
\le \frac{\lambda\ln n}{c},
\label{eq:app:mm-scaling-gap}\\
0
&\le h_{\max}(x)-\frac{h_{\rm bz}^{\lambda}(cx)}{c}
\le \frac{\lambda\ln n}{c}.
\label{eq:app:bz-scaling-gap}
\end{align}
Consequently,
\begin{align}
\sup_{x\in\mathbb R^n}
\left|\frac{h_{\rm lse}^{\lambda}(cx)}{c}-h_{\max}(x)\right|
&\le\frac{\lambda\ln n}{c},\\
\sup_{x\in\mathbb R^n}
\left|\frac{h_{\rm mm}^{\lambda}(cx)}{c}-h_{\max}(x)\right|
&\le\frac{\lambda\ln n}{c},\\
\sup_{x\in\mathbb R^n}
\left|\frac{h_{\rm bz}^{\lambda}(cx)}{c}-h_{\max}(x)\right|
&\le\frac{\lambda\ln n}{c}.
\label{eq:app:uniform-scaling-gaps}
\end{align}
The right-hand sides tend to zero as $c\to\infty$ and do not depend on $x$. Together with~\eqref{eq:app:max-scaling-exact}, this proves uniform convergence on $\mathbb R^n$, and therefore on every compact subset.
\end{proof}
The following lemma collects the order and translation properties used in the contraction argument.
\begin{framed}
\begin{unnumberedlemma}[Monotonicity and translation equivariance]
Each mapping in $\{h_{\max},h_{\rm lse}^{\lambda},h_{\rm mm}^{\lambda}\}$ is monotone in the sense of~\cref{def:monotone-operator}. Moreover, for every $z\in\mathbb R^{|{\cal A}|}$ and $c\in\mathbb R$,
\begin{align}
h_{\max}(z+c{\bf 1})&=h_{\max}(z)+c,\nonumber\\
h_{\rm lse}^{\lambda}(z+c{\bf 1})
&=\lambda\ln\!\left(\sum_i e^{(z_i+c)/\lambda}\right)
=c+h_{\rm lse}^{\lambda}(z),\nonumber\\
h_{\rm mm}^{\lambda}(z+c{\bf 1})
&=\lambda\ln\!\left(\frac{1}{|{\cal A}|}\sum_i e^{(z_i+c)/\lambda}\right)
=c+h_{\rm mm}^{\lambda}(z).
\label{eq:app:translation-equivariance}
\end{align}
\end{unnumberedlemma}
\end{framed}
\begin{proof}
If $u\le v$ componentwise, then $h_{\max}(u)\le h_{\max}(v)$. The same implication holds for $h_{\rm lse}^{\lambda}$ and $h_{\rm mm}^{\lambda}$ because the exponential, summation, and logarithm operations preserve the componentwise order. The three translation identities follow by factoring $e^{c/\lambda}$ from the corresponding exponential sums; the max identity is immediate.
\end{proof}

The following result states that the Bellman operators induced by the max, LSE, and mellowmax operators are contractions.
\begin{framed}
\begin{lemma}[Bellman contraction for the max, LSE, and mellowmax operators]\label{lemma:contraction}
For every $F\in\{F_{\max},F_{\rm lse}^{\lambda},F_{\rm mm}^{\lambda}\}$,
\[
\|F(x)-F(y)\|_\infty\le\gamma\|x-y\|_\infty,
\qquad x,y\in\mathbb R^{|\mathcal S\times\mathcal A|}.
\]
\end{lemma}
\end{framed}
\begin{proof}
Let $h\in\{h_{\max},h_{\rm lse}^{\lambda},h_{\rm mm}^{\lambda}\}$ and let $u,v\in\mathbb R^{|{\cal A}|}$. Then
\begin{align}
v-\|u-v\|_\infty{\bf 1}\le u\le v+\|u-v\|_\infty{\bf 1}
\label{eq:app:coordinate-sandwich}
\end{align}
componentwise. Applying the preceding monotonicity and translation-equivariance lemma to~\eqref{eq:app:coordinate-sandwich} gives
\begin{align}
h(v)-\|u-v\|_\infty
&=h(v-\|u-v\|_\infty{\bf 1})
\le h(u)\nonumber\\
&\le h(v+\|u-v\|_\infty{\bf 1})
=h(v)+\|u-v\|_\infty.
\end{align}
Therefore,
\begin{align}
|h(u)-h(v)|\le\|u-v\|_\infty.
\label{eq:app:scalar-nonexpansive}
\end{align}
Applying~\eqref{eq:app:scalar-nonexpansive} state by state yields
\begin{align}
\|H(x)-H(y)\|_\infty\le\|x-y\|_\infty.
\label{eq:app:H-nonexpansive}
\end{align}
Since every row of $P$ is a probability vector, $\|Pz\|_\infty\le\|z\|_\infty$. Hence
\begin{align}
\|F(x)-F(y)\|_\infty
&=\gamma\|P(H(x)-H(y))\|_\infty\nonumber\\
&\le\gamma\|H(x)-H(y)\|_\infty
\le\gamma\|x-y\|_\infty.
\end{align}
This proves the claim.
\end{proof}

\section{Auxiliary stability results for preconditioned operator dynamics}\label{sec:app:preconditioned-auxiliary}
Consider the system in~\eqref{eq:ODE1}:
\begin{align*}
\frac{d}{{dt}}{x_t} = DF({x_t}) - D{x_t},\quad \forall t\ge 0,\quad  x_0 \in {\mathbb R}^n,
\end{align*}
where $t\ge 0$ is the continuous time, $x_t \in {\mathbb R}^n$ is the state at time $t$, $F: {\mathbb R}^n \to {\mathbb R}^n$ is a mapping that will be specified later, and $D \in {\mathbb R}^{n\times n}$ is a positive definite diagonal matrix with strictly positive diagonal elements $d_i>0, i\in \{1,2,\ldots,n \}$.
The following result is an important intermediate step for our stability analysis of this ODE.
\begin{framed}
\begin{lemma}\label{lemma:3}
Consider the system in~\eqref{eq:ODE1}. Let $x_t \in {\mathbb R}^n$, $t\ge 0$, be its unique solution, and suppose that $x^*$ is the unique fixed point of $F$, i.e., $x^*= F(x^*)$. Moreover, suppose that $p\in\mathbb N$ and set $w_i=1/d_i$ for all $i\in\{1,2,\ldots,n\}$. Then, at every $t\ge0$ such that $x_t\ne x^*$, we have
\begin{align*}
\frac{d}{{dt}}{\left\| x_t - x^* \right\|_{2p,w}}\le& - \frac{{\left\| x_t - x^* \right\|_{2p}^{2p}}}{{\left\| x_t - x^* \right\|_{2p,w}^{2p - 1}}} + \frac{{\left\| x_t - x^* \right\|_{2p}^{2p - 1}}}{{\left\| x_t - x^* \right\|_{2p,w}^{2p - 1}}}{\left\| F(x_t) - F(x^*) \right\|_{2p}}.
\end{align*}
\end{lemma}
\end{framed}
\begin{proof}
Fix any $t\ge0$ such that $x_t\ne x^*$. Using
\begin{align*}
\frac{\partial }{{\partial {x_i}}}{\left( {\sum\limits_{j = 1}^n {{w_j}{x_j}^{2p}} } \right)^{1/(2p)}} = \frac{{{w_i}{x_i}^{2p-1}}}{{\left\| x \right\|_{2p,w}^{2p - 1}}},
\end{align*}
the chain rule gives
\begin{align*}
&\frac{d}{dt}{\left\| {{x_t} - {x^*}} \right\|_{2p,w}}\\
 =& \frac{d}{{dt}}{\left( {\sum\limits_{j = 1}^n {{w_j}({x_{t,j}} - x_j^*)^{2p}} } \right)^{1/(2p)}}\\
=& \sum\limits_{i = 1}^n {{{\left. {\frac{\partial }{{\partial {x_i}}}{{\left( {\sum\limits_{j = 1}^n {{w_j}({x_j} - x_j^*)^{2p}} } \right)}^{1/(2p)}}} \right|}_{{x_i} = {x_{t,i}}}}\frac{{d{x_{t,i}}}}{dt}} \\
=& \frac{1}{{\left\| {{x_t} - {x^*}} \right\|_{2p,w}^{2p - 1}}}{\left[ {\begin{array}{*{20}{c}}
{{w_1}({x_{t,1}} - x_1^*)^{2p-1}}\\
{{w_2}({x_{t,2}} - x_2^*)^{2p-1}}\\
 \vdots \\
{{w_n}({x_{t,n}} - x_n^*)^{2p-1}}
\end{array}} \right]^\top} (DF({x_t}) - D{x_t} + D{x^*} - DF({x^*}))\\
=& {T_1} + {T_2},
\end{align*}
where $T_1$ and $T_2$ are defined below and $x^*= F(x^*)$ is used in the last line.
For $T_1$, we obtain
\begin{align*}
{T_1} =& \frac{1}{{\left\| {{x_t} - {x^*}} \right\|_{2p,w}^{2p - 1}}}{\left[ {\begin{array}{*{20}{c}}
{{w_1}({x_{t,1}} - x_1^*)^{2p-1}}\\
{{w_2}({x_{t,2}} - x_2^*)^{2p-1}}\\
 \vdots \\
{{w_n}({x_{t,n}} - x_n^*)^{2p-1}}
\end{array}} \right]^\top} (D{x^*} - D{x_t})\\
=&  - \frac{{\left\| x_t - x^* \right\|_{2p}^{2p}}}{{\left\| x_t - x^* \right\|_{2p,w}^{2p - 1}}},
\end{align*}
where we use $w_i = \frac{1}{d_i}$ for all $i\in \{1,2,\ldots, n \}$ in the second line.
Moreover, $T_2$ can be bounded as
\begin{align*}
{T_2} =& \frac{1}{{\left\| x_t - x^* \right\|_{2p,w}^{2p - 1}}}{\left[ {\begin{array}{*{20}{c}}
{(x_{t,1} - x_1^*)^{2p-1}}\\
{(x_{t,2} - x_2^*)^{2p-1}}\\
 \vdots \\
{({x_{t,n}} - x_n^*)^{2p-1}}
\end{array}} \right]^\top} (F(x_t) - F(x^*))\\
\le& \frac{1}{\left\| x_t - x^* \right\|_{2p,w}^{2p - 1}}{\left\| {\left[ {\begin{array}{*{20}{c}}
{(x_{t,1} - x_1^*)^{2p-1}}\\
{(x_{t,2} - x_2^*)^{2p-1}}\\
 \vdots \\
{(x_{t,n} - x_n^*)^{2p-1}}
\end{array}} \right]} \right\|_q} {\left\| F(x_t) - F(x^*) \right\|_{2p}}\\
=& \frac{{\left\| x_t - x^* \right\|_{2p}^{2p - 1}}}{{\left\| x_t - x^* \right\|_{2p,w}^{2p - 1}}}{\left\| {F(x_t) - F({x^*})} \right\|_{2p}},
\end{align*}
where H\"{o}lder's inequality is used in the first inequality, and $1/(2p) + 1/q = 1$ is used in the remaining parts.
\end{proof}
The preceding derivative estimate yields the following stability result.
\begin{framed}
\begin{theorem}\label{thm:stability1}
Consider the system in~\eqref{eq:ODE1}, and let $x_t \in {\mathbb R}^n$, $t\ge 0$, be its unique solution.
Suppose that $p\in\mathbb N$ and the mapping $F: {\mathbb R}^n \to {\mathbb R}^n$ is a contraction with respect to $\| \cdot \|_{2p}$, i.e., ${\left\| {F(x) - F(y)} \right\|_{2p}} \le \alpha {\left\| {x - y} \right\|_{2p}},\forall x,y\in {\mathbb R}^n$ for some $\alpha  \in (0,1)$ so that it admits the unique fixed point $F({x^*}) = {x^*}$. Then, we have
\begin{align}
&\frac{d}{{dt}}{\left\| x_t - x^* \right\|_{2p,w}} \le \frac{{(\alpha  - 1)}}{{w_{\max }}}{\left\| x_t - x^* \right\|_{2p,w}},\quad \forall t\ge 0,\quad x_0 \in {\mathbb R}^n\label{eq:3}
\end{align}
and
\begin{align*}
{\left\| x_t - x^* \right\|_{2p,w}} \le& {\left\| x_0 - x^* \right\|_{2p,w}}\exp \left( {\frac{{(\alpha  - 1)t}}{{w_{\max }}}} \right),\quad \forall t\ge 0,\quad x_0 \in {\mathbb R}^n,
\end{align*}
where ${w_{\min }}: = {\min _{i \in \{ 1,2, \ldots ,n\} }}{w_i}$, ${w_{\max }}: = {\max _{i \in \{ 1,2, \ldots ,n\} }}{w_i}$, and $w_i = \frac{1}{d_i}, \forall i \in \{1,2,\ldots,n \}$.
Therefore, $x^*$ is the unique globally exponentially stable equilibrium point.
\end{theorem}
\end{framed}
\begin{proof}
Using~\cref{lemma:3} and the hypothesis
\[{\left\| {F(x) - F(y)} \right\|_{2p}} \le \alpha {\left\| {x - y} \right\|_{2p}},\forall x,y\in {\mathbb R}^n\]
we have
\begin{align*}
\frac{d}{dt}{\left\| x_t - x^* \right\|_{2p,w}} \le& (\alpha  - 1)\frac{{\left\| x_t - x^* \right\|_{2p}^{2p}}}{{\left\| x_t - x^* \right\|_{2p,w}^{2p - 1}}}\le (\alpha  - 1)\frac{1}{{w_{\max }}}\frac{{\left\| x_t - x^* \right\|_{2p,w}^{2p}}}{{\left\| x_t - x^* \right\|_{2p,w}^{2p - 1}}}\\
=& \frac{{(\alpha  - 1)}}{{w_{\max }}}{\left\| x_t - x^* \right\|_{2p,w}},
\end{align*}
where~\cref{lemma:2} is used in the second line. Next, using Gr\"{o}nwall's inequality in~\cref{lemma:Gronwall} yields the desired conclusion.
\end{proof}
The result in~\cref{thm:stability1} holds for every $p\in\mathbb N$.
Moreover,~\eqref{eq:3} shows that $V(x):=\|x-x^*\|_{2p,w}$ is a Lyapunov function~\citep{khalil2002nonlinear}.
Note that when $p=\infty$, $\left\| x - x^* \right\|_{2p}$ becomes non-differentiable in $x$. In order to bypass this issue, we can use the approximation property in~\cref{lemma:2} instead of directly dealing with $\infty$-norm. This approach provides simpler stability analysis compared to the approach in~\cite{borkar1997analog}.

\begin{framed}
\begin{lemma}\label{lemma:1}
Suppose that the mapping $F':{\mathbb R}^n \to {\mathbb R}^n$ is a contraction mapping, i.e.
\begin{align*}
{\left\| {F'(x) - F'(y)} \right\|_{2p}} \le \alpha {\left\| {x - y} \right\|_{2p}},\quad \forall x,y \in {\mathbb R}^n
\end{align*}
for some $\alpha  \in (0,1)$, where $p\in\mathbb N$.
Moreover, suppose that the mapping $F:{\mathbb R}^n \to {\mathbb R}^n$ satisfies ${\left\| {F(x) - F'(x)} \right\|_{2p}} \le \eta,\forall x\in {\mathbb R}^n$, where $\eta >0$ is some real number.
Then, we have ${\left\| {F(x) - F'(y)} \right\|_{2p}} \le \alpha {\left\| {x - y} \right\|_{2p}} + \eta,\forall x,y\in {\mathbb R}^n$.
\end{lemma}
\end{framed}
\begin{proof}
The triangle inequality gives
\begin{align*}
{\left\| F(x) - F'(y) \right\|_{2p} }
\le& {\left\| F(x) - F'(x) \right\|_{2p} } + {\left\| F'(x) - F'(y) \right\|_{2p} }\\
\le& \eta + \alpha {\left\| {x - y} \right\|_{2p} },
\end{align*}
where the last line follows from the hypotheses on $F$ and $F'$. This completes the proof.
\end{proof}

Next, we study properties related to the convergence of the ODE~\eqref{eq:ODE1} when $F$ is not a contraction. To this end, we first need the following result, which serves as an intermediate step toward establishing the stability of the ODE~\eqref{eq:ODE1}.
\begin{framed}
\begin{theorem}\label{thm:stability3}
Suppose that $F:{\mathbb R}^n\to{\mathbb R}^n$ is globally Lipschitz continuous, and let $x_t \in {\mathbb R}^n$, $t\ge 0$, be the unique global solution of~\eqref{eq:ODE1}.
Suppose that $p\in\mathbb N$, the mapping $F':{\mathbb R}^n \to {\mathbb R}^n$ is a contraction mapping with respect to the $2p$-norm, i.e.,
\begin{align*}
{\left\| {F'(x) - F'(y)} \right\|_{2p}} \le \alpha {\left\| {x - y} \right\|_{2p}},\quad \forall x,y \in {\mathbb R}^n
\end{align*}
for some $\alpha  \in (0,1)$, and the corresponding fixed point is $x^*$. Moreover, suppose that $F$ satisfies ${\left\| {F(x) - F'(x)} \right\|_{2p}} \le \eta,\forall x\in {\mathbb R}^n$, where $\eta >0$ is some real number. Then, for every initial state $x_0\in{\mathbb R}^n$, the map $t\mapsto\|x_t-x^*\|_{2p,w}$ is absolutely continuous on every finite time interval and satisfies
\begin{align*}
\frac{d}{dt}\|x_t-x^*\|_{2p,w}
\le\frac{\alpha-1}{w_{\max}}\|x_t-x^*\|_{2p,w}
+\frac{\eta}{w_{\min}^{(2p-1)/(2p)}}
\end{align*}
for almost every $t\ge0$. Moreover,
\begin{align*}
\|x_t-x^*\|_{2p,w}
\le\|x_0-x^*\|_{2p,w}\exp\!\left(\frac{\alpha-1}{w_{\max}}t\right)
+\frac{\eta w_{\max}}{(1-\alpha)w_{\min}^{(2p-1)/(2p)}},
\qquad t\ge0,
\end{align*}
where ${w_{\min }}: = {\min _{i \in \{ 1,2, \ldots ,n\} }}{w_i}$, ${w_{\max }}: = {\max _{i \in \{ 1,2, \ldots ,n\} }}{w_i}$, $w_i=1/d_i$ for all $i\in\{1,2,\ldots,n\}$, and $d_i>0$ are the diagonal elements of $D$ in~\eqref{eq:ODE1}.
Therefore,
\begin{align*}
{x_t} \to {\cal H}: = \left\{ {x \in {\mathbb R}^n:{{\left\| {x - {x^*}} \right\|}_{2p,w}} \le \frac{{\eta w_{\max }}}{{(1 - \alpha)w_{\min }^{(2p - 1)/(2p)}}}} \right\}
\end{align*}
as $t\to \infty$.
\end{theorem}
\end{framed}
\begin{proof}
Fix a time $t\ge0$ such that $x_t\ne x^*$. Proceeding as in~\cref{lemma:3}, we obtain
\begin{align*}
&\frac{d}{{dt}}{\left\| {{x_t} - {x^*}} \right\|_{2p,w}}\\
 =& \frac{1}{{\left\| {{x_t} - {x^*}} \right\|_{2p,w}^{2p - 1}}}{\left[ {\begin{array}{*{20}{c}}
{{w_1}({x_{t,1}} - x_1^*)^{2p-1}}\\
{{w_2}({x_{t,2}} - x_2^*)^{2p-1}}\\
 \vdots \\
{{w_n}({x_{t,n}} - x_n^*)^{2p-1}}
\end{array}} \right]^\top}(DF({x_t}) - D{x_t} + D{x^*} - DF'({x^*}))\\
=& T_1 + T_2
\end{align*}
where $T_1$ and $T_2$ are defined below.
For $T_1$, we obtain
\begin{align*}
{T_1} =& \frac{1}{{\left\| {{x_t} - {x^*}} \right\|_{2p,w}^{2p - 1}}}{\left[ {\begin{array}{*{20}{c}}
{{w_1}({x_{t,1}} - x_1^*)^{2p-1}}\\
{{w_2}({x_{t,2}} - x_2^*)^{2p-1}}\\
 \vdots \\
{{w_n}({x_{t,n}} - x_n^*)^{2p-1}}
\end{array}} \right]^\top} (D{x^*} - D{x_t})\\
=&  - \frac{{\left\| x_t - x^* \right\|_{2p}^{2p}}}{{\left\| x_t - x^* \right\|_{2p,w}^{2p - 1}}},
\end{align*}
where we use $w_i = \frac{1}{d_i}$ for all $i\in \{1,2,\ldots, n \}$ in the second line.
Moreover, $T_2$ can be bounded as
\begin{align*}
{T_2} =& \frac{1}{{\left\| x_t - x^* \right\|_{2p,w}^{2p - 1}}}{\left[ {\begin{array}{*{20}{c}}
{(x_{t,1} - x_1^*)^{2p-1}}\\
{(x_{t,2} - x_2^*)^{2p-1}}\\
 \vdots \\
{({x_{t,n}} - x_n^*)^{2p-1}}
\end{array}} \right]^\top} (F(x_t) - F'(x^*))\\
\le& \frac{1}{\left\| x_t - x^* \right\|_{2p,w}^{2p - 1}}{\left\| {\left[ {\begin{array}{*{20}{c}}
{(x_{t,1} - x_1^*)^{2p-1}}\\
{(x_{t,2} - x_2^*)^{2p-1}}\\
 \vdots \\
{(x_{t,n} - x_n^*)^{2p-1}}
\end{array}} \right]} \right\|_q} {\left\| F(x_t) - F'(x^*) \right\|_{2p}}\\
=& \frac{{\left\| x_t - x^* \right\|_{2p}^{2p - 1}}}{{\left\| x_t - x^* \right\|_{2p,w}^{2p - 1}}}{\left\| {F(x_t) - F'({x^*})} \right\|_{2p}},
\end{align*}
where H\"{o}lder's inequality is used in the first inequality, and $1/(2p) + 1/q = 1$ is used in the remaining parts.
Next, using~\cref{lemma:1}, one has
\begin{align*}
T_2
&\le \frac{\|x_t-x^*\|_{2p}^{2p-1}}
{\|x_t-x^*\|_{2p,w}^{2p-1}}
\|F(x_t)-F'(x^*)\|_{2p}\\
&\le \alpha\frac{\|x_t-x^*\|_{2p}^{2p}}
{\|x_t-x^*\|_{2p,w}^{2p-1}}
+\eta\frac{\|x_t-x^*\|_{2p}^{2p-1}}
{\|x_t-x^*\|_{2p,w}^{2p-1}}\\
&\le \alpha\frac{\|x_t-x^*\|_{2p}^{2p}}
{\|x_t-x^*\|_{2p,w}^{2p-1}}
+\frac{\eta}{w_{\min}^{(2p-1)/(2p)}}.
\end{align*}
Here, the second inequality uses~\cref{lemma:1}:
\begin{align*}
{\left\| {F({x_t}) - F'({x^*})} \right\|_{2p}} =& {\left\| {F({x_t}) - F'({x_t}) + F'({x_t}) - F'({x^*})} \right\|_{2p}}\\
\le& {\left\| {F({x_t}) - F'({x_t})} \right\|_{2p}} + {\left\| {F'({x_t}) - F'({x^*})} \right\|_{2p}}\\
\le& \eta  + \alpha {\left\| {{x_t} - {x^*}} \right\|_{2p}}.
\end{align*}

Combining the bounds on $T_1$ and $T_2$ gives
\begin{align*}
\frac{d}{dt}\|x_t-x^*\|_{2p,w}
&\le (\alpha-1)\frac{\|x_t-x^*\|_{2p}^{2p}}
{\|x_t-x^*\|_{2p,w}^{2p-1}}
+\frac{\eta}{w_{\min}^{(2p-1)/(2p)}}\\
&\le (\alpha-1)\frac{\|x_t-x^*\|_{2p,w}^{2p}}
{w_{\max}\|x_t-x^*\|_{2p,w}^{2p-1}}
+\frac{\eta}{w_{\min}^{(2p-1)/(2p)}}\\
&=\frac{\alpha-1}{w_{\max}}\|x_t-x^*\|_{2p,w}
+\frac{\eta}{w_{\min}^{(2p-1)/(2p)}}.
\end{align*}
The map $t\mapsto\|x_t-x^*\|_{2p,w}$ is nonnegative and absolutely continuous on every finite time interval, since $x_t$ is continuously differentiable and the norm is Lipschitz continuous. At any interior time at which this map is differentiable and equals zero, it has a local minimum, so its derivative is zero. The additive term in the preceding bound is nonnegative; hence the same inequality holds at these times. Since an absolutely continuous function is differentiable almost everywhere, we obtain
\begin{align*}
\frac{d}{dt}\|x_t-x^*\|_{2p,w}
\le\frac{\alpha-1}{w_{\max}}\|x_t-x^*\|_{2p,w}
+\frac{\eta}{w_{\min}^{(2p-1)/(2p)}}
\end{align*}
for almost every $t\ge0$. Applying Gr\"{o}nwall's inequality in~\cref{lemma:Gronwall} on each finite time interval gives
\begin{align*}
\|x_t-x^*\|_{2p,w}
\le\|x_0-x^*\|_{2p,w}\exp\!\left(\frac{\alpha-1}{w_{\max}}t\right)
+\frac{\eta w_{\max}}{(1-\alpha)w_{\min}^{(2p-1)/(2p)}},
\qquad t\ge0.
\end{align*}
The final claim follows from this bound. This completes the proof.
\end{proof}
\cref{thm:stability3} uses the $2p$-norm as a Lyapunov function for a finite $p\in\mathbb N$. This argument can also be extended to the case $p\to\infty$.
\Cref{thm:norm-perturb} shows that without the contraction mapping property, the ODE in~\eqref{eq:ODE1} may still approach an error set around the fixed point $x^*$ of $F'$.

The following result is required to prove the stability of the ODE~\eqref{eq:ODE1} with the Boltzmann softmax operator and~\cref{thm:stability3}.
\begin{framed}
\begin{lemma}\label{lemma:contraction2}
For the Boltzmann softmax operator, the mapping $F_{\rm bz}^{\lambda}$ satisfies
\begin{align*}
{\left\| {F_{\rm{bz}}^\lambda (x) - {F_{\max }}(x)} \right\|_\infty } \le \gamma\lambda\ln(|{\cal A}|),\quad \forall x \in {\mathbb R}^{|{\cal S} \times {\cal A}|}.
\end{align*}
\end{lemma}
\end{framed}
\begin{proof}
To begin with,~\cref{lemma:5} gives $H_{\max}(x)-\lambda\ln(|{\cal A}|){\bf 1}\le H_{\rm bz}^{\lambda}(x)\le H_{\max}(x)$, where ${\bf 1}\in{\mathbb R}^{|{\cal S}|}$ has all entries equal to one. Equivalently, $0\le H_{\max}(x)-H_{\rm bz}^{\lambda}(x)\le\lambda\ln(|{\cal A}|){\bf 1}$. Therefore,
\begin{align}
{\left\| {H_{\rm bz}^{\lambda}(x) - H_{\max}(x)} \right\|_\infty } \le \lambda\ln(|{\cal A}|).\label{eq:11}
\end{align}
Then, we have
\begin{align*}
{\left\| {F_{\max }(x) - F_{\rm{bz}}^\lambda (x)} \right\|_\infty } \le& {\left\| {\gamma P{H_{\max}}(x) - \gamma PH_{\rm{bz}}^\lambda (x)} \right\|_\infty }\\
\le& \gamma {\left\| {{H_{\max}}(x) - H_{{\rm{bz}}}^\lambda (x)} \right\|_\infty } \le \gamma\lambda\ln(|{\cal A}|),
\end{align*}
where the last inequality is due to~\eqref{eq:11}.
This concludes the proof.
\end{proof}
The next lemma shows that the Boltzmann operator is globally Lipschitz. The precise constant is not important for the stability argument; we only need global Lipschitz continuity.
\begin{framed}
\begin{lemma}\label{lemma:Lipschitz1}
For the Boltzmann softmax operator, the mapping $H_{\rm bz}^{\lambda}$ satisfies
\begin{align*}
{\left\| {H_{\rm bz}^{\lambda}(x) - H_{\rm bz}^{\lambda}(y)} \right\|_\infty} \le L_{\rm bz}{\left\| {x - y} \right\|_\infty},\quad\forall x,y\in {\mathbb R}^{|{\cal S}\times {\cal A}|},
\end{align*}
where one may take $L_{\rm bz}:=1+2|{\cal A}|/e$.
\end{lemma}
\end{framed}
\begin{proof}
It suffices to prove the claim for the scalar operator $h_{\rm bz}^{\lambda}:{\mathbb R}^{|{\cal A}|}\to{\mathbb R}$. Define
\[
Z(x):=\sum_j e^{x_j/\lambda},
\qquad
\pi_j(x):=\frac{e^{x_j/\lambda}}{Z(x)},
\qquad
h_{\rm bz}^{\lambda}(x)=\sum_j\pi_j(x)x_j.
\]
These functions are smooth on all of ${\mathbb R}^{|{\cal A}|}$. Differentiating the quotient directly gives
\[
\frac{\partial\pi_j(x)}{\partial x_i}
=\frac{1}{\lambda}\pi_j(x)\bigl({\bf 1}_{\{i=j\}}-\pi_i(x)\bigr),
\]
and hence
\begin{align*}
\frac{\partial h_{\rm bz}^{\lambda}(x)}{\partial x_i}
&=\pi_i(x)+\sum_jx_j\frac{\partial\pi_j(x)}{\partial x_i}\\
&=\pi_i(x)+\frac{1}{\lambda}\pi_i(x)
\left(x_i-\sum_j\pi_j(x)x_j\right)\\
&=\pi_i(x)\left(1+\frac{x_i-h_{\rm bz}^{\lambda}(x)}{\lambda}\right).
\end{align*}
This identity holds at every $x$, including points with multiple maximizing coordinates, because no derivative of the max function has been used.

To bound the gradient, use $x_i\le h_{\max}(x)$ and $h_{\rm bz}^{\lambda}(x)\le h_{\max}(x)$ to obtain
\begin{align}
\sum_{i=1}^{|{\cal A}|}
\left|\frac{\partial h_{\rm bz}^{\lambda}(x)}{\partial x_i}\right|
&=\sum_{i=1}^{|{\cal A}|}\pi_i(x)
\left|1+\frac{x_i-h_{\rm bz}^{\lambda}(x)}{\lambda}\right|\nonumber\\
&\le 1+\frac{1}{\lambda}\sum_{i=1}^{|{\cal A}|}\pi_i(x)
\left|x_i-h_{\rm bz}^{\lambda}(x)\right|\nonumber\\
&\le 1+\frac{1}{\lambda}\sum_{i=1}^{|{\cal A}|}\pi_i(x)
\left[h_{\max}(x)-x_i+h_{\max}(x)-h_{\rm bz}^{\lambda}(x)\right]\nonumber\\
&=1+\frac{2}{\lambda}\sum_{i=1}^{|{\cal A}|}\pi_i(x)
\left(h_{\max}(x)-x_i\right)\nonumber\\
&=1+2\frac{\sum_{i=1}^{|{\cal A}|}
\frac{h_{\max}(x)-x_i}{\lambda}
\exp\!\left(-\frac{h_{\max}(x)-x_i}{\lambda}\right)}
{\sum_{j=1}^{|{\cal A}|}\exp\!\left(-\frac{h_{\max}(x)-x_j}{\lambda}\right)}\nonumber\\
&\le 1+2\sum_{i=1}^{|{\cal A}|}
\frac{h_{\max}(x)-x_i}{\lambda}
\exp\!\left(-\frac{h_{\max}(x)-x_i}{\lambda}\right)\nonumber\\
&\le 1+\frac{2|{\cal A}|}{e}=L_{\rm bz}.
\label{eq:app:bz-gradient-bound}
\end{align}
The first equality after the inequalities uses $\sum_i\pi_i(x)=1$ and
$h_{\max}(x)-h_{\rm bz}^{\lambda}(x)
=\sum_i\pi_i(x)(h_{\max}(x)-x_i)$.
The denominator in~\eqref{eq:app:bz-gradient-bound} is at least one because a maximizing coordinate makes one exponential term equal to one. The last inequality uses $u e^{-u}\le e^{-1}$ for $u\ge0$, applied to each nonnegative ratio $(h_{\max}(x)-x_i)/\lambda$.

For any $u,v\in\mathbb R^{|{\cal A}|}$, the integral form of the mean-value theorem along the line segment from $v$ to $u$ gives
\begin{align}
h_{\rm bz}^{\lambda}(u)-h_{\rm bz}^{\lambda}(v)
&=\int_0^1\frac{d}{dt}
 h_{\rm bz}^{\lambda}\bigl(v+t(u-v)\bigr)\,dt\nonumber\\
&=\int_0^1\nabla h_{\rm bz}^{\lambda}\bigl(v+t(u-v)\bigr)^\top
 (u-v)\,dt.
\label{eq:app:bz-segment-identity}
\end{align}
Using~\eqref{eq:app:bz-gradient-bound} in~\eqref{eq:app:bz-segment-identity} yields
\begin{align}
\left|h_{\rm bz}^{\lambda}(u)-h_{\rm bz}^{\lambda}(v)\right|
&\le\int_0^1\sum_{i=1}^{|{\cal A}|}
 \left|\frac{\partial h_{\rm bz}^{\lambda}}{\partial x_i}
 \bigl(v+t(u-v)\bigr)\right|\,|u_i-v_i|\,dt\nonumber\\
&\le\|u-v\|_\infty\int_0^1
 \left\|\nabla h_{\rm bz}^{\lambda}\bigl(v+t(u-v)\bigr)\right\|_1\,dt\nonumber\\
&\le L_{\rm bz}\|u-v\|_\infty.
\label{eq:app:bz-scalar-lipschitz}
\end{align}
Finally, for $x,y\in\mathbb R^{|{\cal S}\times{\cal A}|}$, apply~\eqref{eq:app:bz-scalar-lipschitz} to the action-value vectors at each state:
\begin{align*}
\left\|H_{\rm bz}^{\lambda}(x)-H_{\rm bz}^{\lambda}(y)\right\|_\infty
&=\max_{s\in{\cal S}}
\left|h_{\rm bz}^{\lambda}\bigl(x(s,\cdot)\bigr)
-h_{\rm bz}^{\lambda}\bigl(y(s,\cdot)\bigr)\right|\\
&\le L_{\rm bz}\max_{s\in{\cal S}}
\|x(s,\cdot)-y(s,\cdot)\|_\infty\\
&=L_{\rm bz}\max_{(s,a)\in{\cal S}\times{\cal A}}
|x(s,a)-y(s,a)|\\
&=L_{\rm bz}\|x-y\|_\infty.
\end{align*}
This proves the claim.
\end{proof}

\begin{framed}
\begin{lemma}\label{lemma:boundedness-bz}
Consider~\cref{algo:Q-learning} with the Boltzmann softmax operator and step-sizes satisfying $\alpha_k>0$ and $\sum_{k=0}^{\infty}\alpha_k^2<\infty$. Then,
\[
\sup_{k\ge0}\|Q_k\|_\infty<\infty
\]
for every sample path.
\end{lemma}
\end{framed}
\begin{proof}
The step-size condition implies $\alpha_k\to0$. Hence, there exists an integer $K\ge0$ such that $0<\alpha_k\le1$ for every $k\ge K$. Set
\[
B:=\max\left\{\max_{0\le j\le K}\|Q_j\|_\infty,\frac{\max_{(s,a,s')\in{\cal S}\times{\cal A}\times{\cal S}}|r(s,a,s')|}{1-\gamma}\right\}.
\]
The number $B$ is finite because only finitely many iterates precede $K$ and the state and action spaces are finite. Moreover, $h_{\rm bz}^{\lambda}(q)$ is a convex combination of the coordinates of $q$, and therefore $|h_{\rm bz}^{\lambda}(q)|\le\|q\|_\infty$. Suppose that $\|Q_k\|_\infty\le B$ for some $k\ge K$. For the updated coordinate, the update rule and $0<\alpha_k\le1$ give
\begin{align*}
|Q_{k+1}(s_k,a_k)|
&=\left|(1-\alpha_k)Q_k(s_k,a_k)
+\alpha_k\left[r(s_k,a_k,s_k')
+\gamma h_{\rm bz}^{\lambda}\bigl(Q_k(s_k',\cdot)\bigr)\right]\right|\\
&\le(1-\alpha_k)|Q_k(s_k,a_k)|
+\alpha_k\left|r(s_k,a_k,s_k')
+\gamma h_{\rm bz}^{\lambda}\bigl(Q_k(s_k',\cdot)\bigr)\right|\\
&\le(1-\alpha_k)|Q_k(s_k,a_k)|
+\alpha_k|r(s_k,a_k,s_k')|
+\alpha_k\gamma\left|h_{\rm bz}^{\lambda}\bigl(Q_k(s_k',\cdot)\bigr)\right|\\
&\le(1-\alpha_k)\|Q_k\|_\infty
+\alpha_k|r(s_k,a_k,s_k')|
+\alpha_k\gamma\|Q_k(s_k',\cdot)\|_\infty\\
&\le(1-\alpha_k)\|Q_k\|_\infty
+\alpha_k\max_{(s,a,s')\in{\cal S}\times{\cal A}\times{\cal S}}|r(s,a,s')|
+\alpha_k\gamma\|Q_k\|_\infty\\
&\le(1-\alpha_k)B
+\alpha_k\left[\max_{(s,a,s')\in{\cal S}\times{\cal A}\times{\cal S}}|r(s,a,s')|
+\gamma B\right]\\
&\le(1-\alpha_k)B+\alpha_k\left[(1-\gamma)B+\gamma B\right]\\
&=(1-\alpha_k)B+\alpha_k B=B.
\end{align*}
The last inequality follows from the definition of $B$, which gives
$\max_{(s,a,s')\in{\cal S}\times{\cal A}\times{\cal S}}|r(s,a,s')|\le(1-\gamma)B$.
For every unselected pair $(s,a)\ne(s_k,a_k)$, the coordinate remains unchanged:
\[
|Q_{k+1}(s,a)|=|Q_k(s,a)|\le B.
\]
Thus $\|Q_{k+1}\|_\infty\le B$. Since $\|Q_K\|_\infty\le B$, induction proves the bound for every $k\ge K$; the finitely many earlier iterates are already included in the definition of $B$.
\end{proof}

\section{Weighted-norm counterparts of the polynomial results}\label{sec:app:norm-counterparts}
The main text formulates the central stability results through the smooth polynomial Lyapunov function $W_p$. This section collects the corresponding estimates stated directly in the weighted $2p$-norm. It also gives the limiting $\infty$-norm estimates and the positively invariant sets used in the stochastic Boltzmann analysis.

\begin{framed}
\begin{theorem}[Weighted-norm contractive estimate]\label{thm:norm-contract}
Let $x_t$ be the unique solution of~\eqref{eq:ODE1}. Suppose that $F:{\mathbb R}^n\to{\mathbb R}^n$ is an $\infty$-norm contraction with factor $\alpha\in(0,1)$ and unique fixed point $x^*$. Choose $w_i=1/d_i$ and let $w_{\max}:=\max_i w_i$. For any $p\in\mathbb N$ satisfying
\[
\left\lceil \frac{\ln n}{\ln(\alpha^{-1})}\right\rceil<2p,
\]
we have
\begin{align}
\frac{d}{dt}\|x_t-x^*\|_{2p,w}
&\le \frac{\alpha n^{1/(2p)}-1}{w_{\max}}\|x_t-x^*\|_{2p,w},\label{eq:norm-contract-derivative}
\end{align}
and
\begin{align}
\|x_t-x^*\|_{2p,w}
&\le \|x_0-x^*\|_{2p,w}
\exp\!\left(\frac{\alpha n^{1/(2p)}-1}{w_{\max}}t\right),
\qquad t\ge0.\label{eq:norm-contract-bound}
\end{align}
Therefore, $x^*$ is the unique globally exponentially stable equilibrium.
\end{theorem}
\end{framed}
\begin{proof}
At every $t\ge0$ such that $x_t\ne x^*$, differentiating the weighted $2p$-norm directly and using $w_i d_i=1$ gives
\begin{align*}
\frac{d}{dt}\|x_t-x^*\|_{2p,w}
&=\frac{\displaystyle\sum_{i=1}^n (x_{t,i}-x_i^*)^{2p-1}
\bigl(F_i(x_t)-F_i(x^*)-(x_{t,i}-x_i^*)\bigr)}
{\|x_t-x^*\|_{2p,w}^{2p-1}}\\
&\le\frac{\|x_t-x^*\|_{2p}^{2p-1}\|F(x_t)-F(x^*)\|_{2p}
-\|x_t-x^*\|_{2p}^{2p}}
{\|x_t-x^*\|_{2p,w}^{2p-1}}\\
&\le-\left(1-\alpha n^{1/(2p)}\right)
\frac{\|x_t-x^*\|_{2p}^{2p}}
{\|x_t-x^*\|_{2p,w}^{2p-1}}\\
&\le\frac{\alpha n^{1/(2p)}-1}{w_{\max}}
\|x_t-x^*\|_{2p,w}.
\end{align*}
Here, the first inequality follows from H{\"o}lder's inequality, the second from the $\infty$-norm contraction and~\cref{lemma:2}, and the third from $\|x_t-x^*\|_{2p,w}^{2p}\le w_{\max}\|x_t-x^*\|_{2p}^{2p}$. If $x_t=x^*$ at some time, the trajectory remains at $x^*$ thereafter, so the same estimate holds almost everywhere. This proves~\eqref{eq:norm-contract-derivative}. Applying~\cref{lemma:Gronwall} with zero additive term gives~\eqref{eq:norm-contract-bound}.
\end{proof}

\Cref{thm:norm-contract} is the direct weighted-norm counterpart of~\cref{thm:stability2}. The threshold on $p$ ensures that $\alpha n^{1/(2p)}<1$. Weighted $2p$-norm Lyapunov functions have been used for mappings that are contractive in the same norm~\citep{borkar1997analog,borkar2009stochastic}. Here, the finite-norm estimate is obtained from an $\infty$-norm contraction through the norm comparisons in~\cref{lemma:2}.

\begin{framed}
\begin{theorem}[Weighted-norm perturbation estimate]\label{thm:norm-perturb}
Suppose that $F:{\mathbb R}^n\to{\mathbb R}^n$ is globally Lipschitz continuous, and let $x_t$ be the unique global solution of~\eqref{eq:ODE1}. Suppose that $F':{\mathbb R}^n\to{\mathbb R}^n$ is an $\infty$-norm contraction with factor $\alpha\in(0,1)$ and fixed point $x^*$, and suppose that
\[
\|F(x)-F'(x)\|_\infty\le\eta,
\qquad x\in{\mathbb R}^n,
\]
for some $\eta>0$. Choose $w_i=1/d_i$, let $w_{\min}:=\min_iw_i$ and $w_{\max}:=\max_iw_i$, and take any $p\in\mathbb N$ satisfying
\[
\left\lceil \frac{\ln n}{\ln(\alpha^{-1})}\right\rceil<2p.
\]
Then
\begin{align}
\frac{d}{dt}\|x_t-x^*\|_{2p,w}
\le{}&\frac{\alpha n^{1/(2p)}-1}{w_{\max}}\|x_t-x^*\|_{2p,w}
+\frac{n^{1/(2p)}\eta}{w_{\min}^{(2p-1)/(2p)}},\label{eq:norm-perturb-derivative}
\end{align}
for almost every $t\ge0$, and
\begin{align}
\|x_t-x^*\|_{2p,w}
\le{}&\|x_0-x^*\|_{2p,w}
\exp\!\left(\frac{\alpha n^{1/(2p)}-1}{w_{\max}}t\right)\\
&+\frac{n^{1/(2p)}\eta w_{\max}}
{\left(1-\alpha n^{1/(2p)}\right)w_{\min}^{(2p-1)/(2p)}},
\qquad t\ge0.\label{eq:norm-perturb-bound}
\end{align}
Moreover,
\begin{align}
\|x_t-x^*\|_\infty
\le{}&\|x_0-x^*\|_\infty
\exp\!\left(\frac{\alpha-1}{w_{\max}}t\right)
+\frac{\eta w_{\max}}{(1-\alpha)w_{\min}},
\qquad t\ge0.\label{eq:norm-perturb-infinity}
\end{align}
Consequently,
\begin{align}
{\cal H}:=\left\{x\in{\mathbb R}^n:
\|x-x^*\|_\infty\le\frac{\eta w_{\max}}{(1-\alpha)w_{\min}}\right\}\label{eq:norm-perturb-set}
\end{align}
is positively invariant and satisfies ${\rm dist}(x_t,{\cal H})\to0$ as $t\to\infty$.
\end{theorem}
\end{framed}
\begin{proof}
The calculation leading to~\eqref{eq:app:perturb-scalar} gives~\eqref{eq:norm-perturb-derivative} whenever $x_t\ne x^*$. The norm trajectory is nonnegative and absolutely continuous on every finite time interval. At almost every time at which it equals zero, its derivative is zero; hence~\eqref{eq:norm-perturb-derivative} holds almost everywhere. Applying~\cref{lemma:Gronwall} on each finite time interval gives~\eqref{eq:norm-perturb-bound} for every $t\ge0$. Applying~\cref{lemma:2} to~\eqref{eq:norm-perturb-bound} gives
\begin{align*}
\|x_t-x^*\|_\infty
\le{}&\left(\frac{n w_{\max}}{w_{\min}}\right)^{1/(2p)}
\|x_0-x^*\|_\infty
\exp\!\left(\frac{\alpha n^{1/(2p)}-1}{w_{\max}}t\right)\\
&+\frac{n^{1/(2p)}\eta w_{\max}}
{\left(1-\alpha n^{1/(2p)}\right)w_{\min}}.
\end{align*}
Taking $p\to\infty$ proves~\eqref{eq:norm-perturb-infinity}.

It remains to verify positive invariance. At any boundary point of ${\cal H}$, each coordinate satisfying
$|x_{t,i}-x_i^*|=\|x_t-x^*\|_\infty$ obeys
\begin{align*}
\operatorname{sgn}(x_{t,i}-x_i^*)\frac{d}{dt}(x_{t,i}-x_i^*)
\le d_i\left(\eta-(1-\alpha)\|x_t-x^*\|_\infty\right)\le0,
\end{align*}
because $w_{\max}\ge w_{\min}$. Thus, the vector field does not point outside ${\cal H}$. Finally,~\eqref{eq:norm-perturb-infinity} implies ${\rm dist}(x_t,{\cal H})\to0$.
\end{proof}

\Cref{thm:norm-perturb} states the norm estimate corresponding to the polynomial perturbation result in~\cref{thm:stability4}. It makes the limiting $\infty$-norm neighborhood explicit and provides the invariant-set statement used below.

\begin{framed}
\begin{theorem}[Weighted-norm estimate for contractive Q-learning ODEs]\label{thm:norm-Q-contract}
Consider~\eqref{eq:ODE-Q-learning1} with
$H\in\{H_{\max},H_{\rm lse}^{\lambda},H_{\rm mm}^{\lambda}\}$, and let $Q_t$ be its solution. Let $Q_e$ denote the corresponding fixed point in~\eqref{eq:4}, set $n=|{\cal S}\times{\cal A}|$, choose $w_i=1/d_i$, and let $w_{\max}:=\max_iw_i$. For any $p\in\mathbb N$ satisfying
\[
\left\lceil \frac{\ln n}{\ln(\gamma^{-1})}\right\rceil<2p,
\]
we have
\begin{align}
\|Q_t-Q_e\|_{2p,w}
&\le \|Q_0-Q_e\|_{2p,w}
\exp\!\left(\frac{\gamma n^{1/(2p)}-1}{w_{\max}}t\right),
\qquad t\ge0.\label{eq:norm-Q-contract-bound}
\end{align}
\end{theorem}
\end{framed}
\begin{proof}
The Bellman operators $F_{\max}$, $F_{\rm lse}^{\lambda}$, and $F_{\rm mm}^{\lambda}$ are $\gamma$-contractions by~\cref{lemma:contraction}. Applying~\cref{thm:norm-contract} with $x_t=Q_t$, $x^*=Q_e$, and $\alpha=\gamma$ proves the result.
\end{proof}

\Cref{thm:norm-Q-contract} is the weighted-norm counterpart of~\cref{thm:stability5}. It gives exponential convergence for the max, LSE, and mellowmax ODEs in the weighted finite norm.

\begin{framed}
\begin{theorem}[Weighted-norm estimate for the Boltzmann ODE]\label{thm:norm-Q-boltzmann}
Consider~\eqref{eq:ODE-Q-learning1} with $H=H_{\rm bz}^{\lambda}$, and let $Q_t$ be its solution. Set $n=|{\cal S}\times{\cal A}|$, choose $w_i=1/d_i$, and let $w_{\min}:=\min_iw_i$ and $w_{\max}:=\max_iw_i$. For any $p\in\mathbb N$ satisfying
\[
\left\lceil \frac{\ln n}{\ln(\gamma^{-1})}\right\rceil<2p,
\]
we have
\begin{align}
\|Q_t-Q_{\max}^*\|_{2p,w}
\le{}&\|Q_0-Q_{\max}^*\|_{2p,w}
\exp\!\left(\frac{\gamma n^{1/(2p)}-1}{w_{\max}}t\right)\\
&+\frac{n^{1/(2p)}w_{\max}}
{\left(1-\gamma n^{1/(2p)}\right)w_{\min}^{(2p-1)/(2p)}}
\gamma\lambda\ln(|{\cal A}|),
\qquad t\ge0,\label{eq:norm-Q-boltzmann-bound}
\end{align}
and
\begin{align}
\|Q_t-Q_{\max}^*\|_\infty
\le{}&\|Q_0-Q_{\max}^*\|_\infty
\exp\!\left(\frac{\gamma-1}{w_{\max}}t\right)
+\frac{\gamma\lambda\ln(|{\cal A}|)w_{\max}}
{(1-\gamma)w_{\min}},
\qquad t\ge0.\label{eq:norm-Q-boltzmann-infinity}
\end{align}
Consequently,
\begin{align}
{\cal H}:=\left\{Q\in{\mathbb R}^{|{\cal S}\times{\cal A}|}:
\|Q-Q_{\max}^*\|_\infty
\le\frac{\gamma\lambda\ln(|{\cal A}|)w_{\max}}
{(1-\gamma)w_{\min}}\right\}\label{eq:norm-Q-boltzmann-set}
\end{align}
is positively invariant and satisfies ${\rm dist}(Q_t,{\cal H})\to0$ as $t\to\infty$.
\end{theorem}
\end{framed}
\begin{proof}
If $|{\cal A}|=1$, then $F_{\rm bz}^{\lambda}=F_{\max}$ and ${\cal H}=\{Q_{\max}^*\}$. The weighted-norm estimate follows from~\cref{thm:norm-Q-contract}; applying~\cref{lemma:2} and taking $p\to\infty$ gives the stated $\infty$-norm estimate with zero additive term. Since $Q_{\max}^*$ is an equilibrium, ${\cal H}$ is positively invariant, and the error estimate gives ${\rm dist}(Q_t,{\cal H})\to0$.

For $|{\cal A}|\ge2$, we have $\gamma\lambda\ln(|{\cal A}|)>0$. By~\cref{lemma:Lipschitz1}, $F_{\rm bz}^{\lambda}$ is globally Lipschitz continuous. By~\cref{lemma:contraction2},
\[
\|F_{\rm bz}^{\lambda}(Q)-F_{\max}(Q)\|_\infty
\le\gamma\lambda\ln(|{\cal A}|).
\]
Applying~\cref{thm:norm-perturb} with $F'=F_{\max}$, $F=F_{\rm bz}^{\lambda}$, $\alpha=\gamma$, $x_t=Q_t$, $x^*=Q_{\max}^*$, and $\eta=\gamma\lambda\ln(|{\cal A}|)$ proves the result.
\end{proof}

\Cref{thm:norm-Q-boltzmann} is the weighted-norm counterpart of~\cref{thm:stability6}. It provides the exact $\infty$-norm invariant set used in the stochastic set-convergence result. The following example illustrates convergence toward this guaranteed neighborhood.
\begin{figure}[ht!]
    \centering
    \renewcommand\thesubfigure{(\alph{subfigure})}

    \begin{minipage}[b]{0.48\textwidth}
        \centering
        \includegraphics[width=\textwidth, keepaspectratio=true]{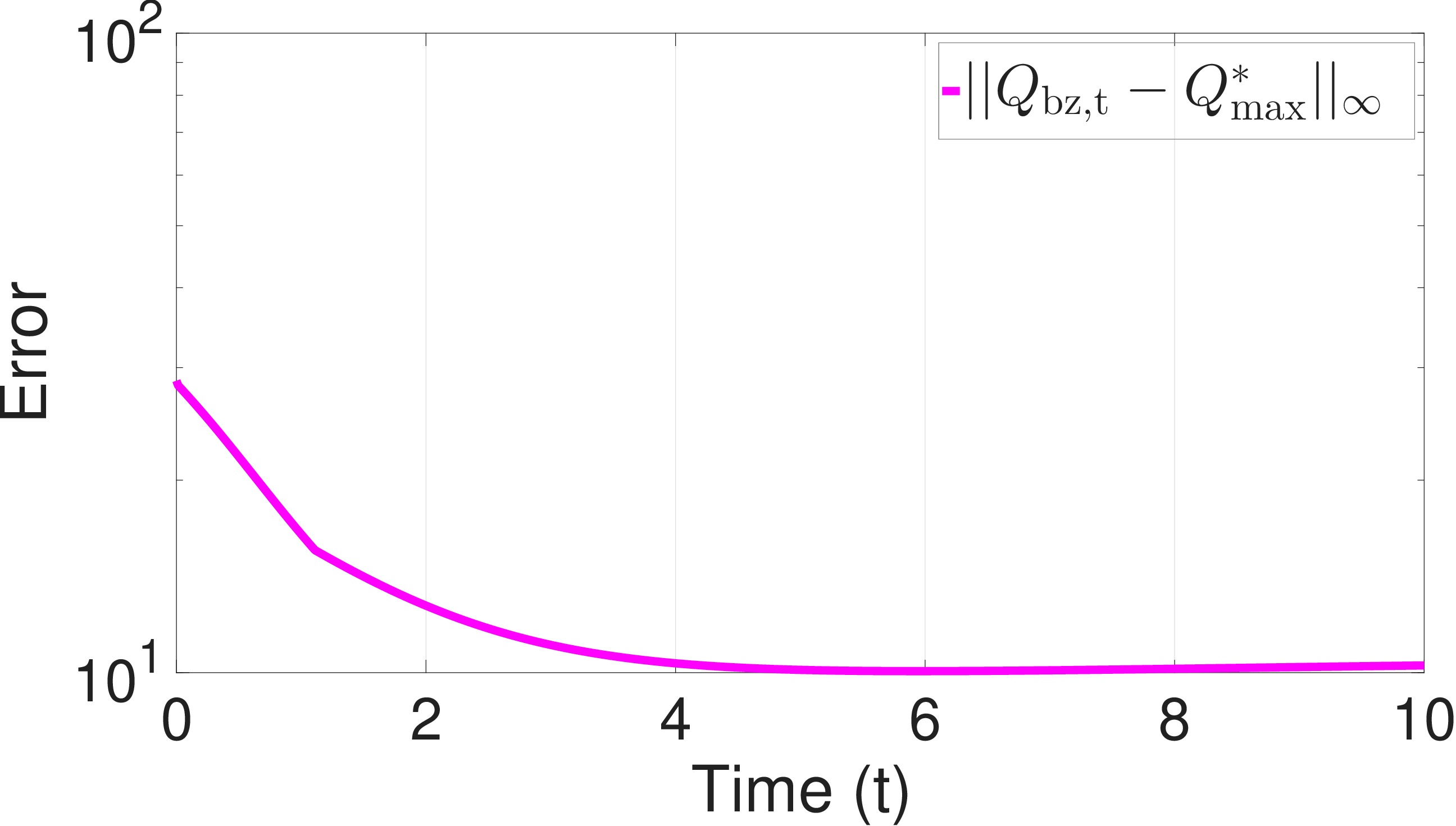}
        \vspace{2mm}
        \centerline{\small (a) Error convergence curve}
        \phantomsubcaption\label{fig:bz_err}
    \end{minipage}
    \hfill
    \begin{minipage}[b]{0.48\textwidth}
        \centering
        \vspace{1.5mm} 
        \includegraphics[width=\textwidth, keepaspectratio=true]{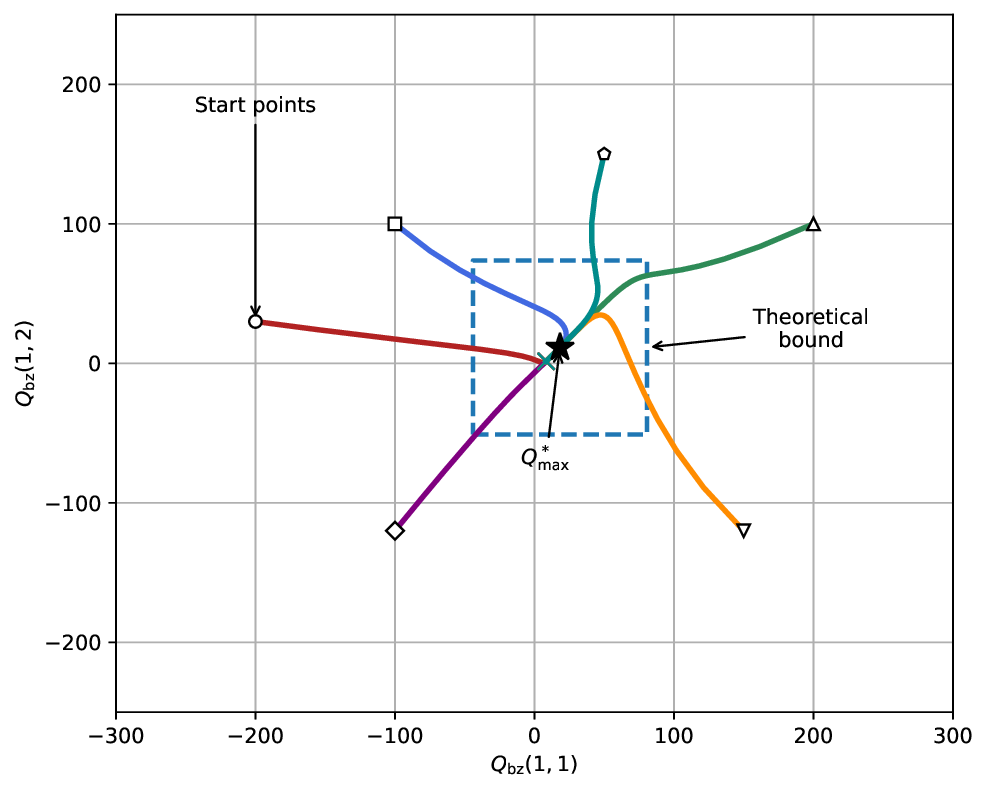}
        \vspace{2mm}
        \centerline{\small (b) Phase plane trajectories}
        \phantomsubcaption\label{fig:bz_init_multi_ph}
    \end{minipage}

    \caption{Empirical stability analysis of the ODE with the Boltzmann operator. (a) Evolution of $\|Q_t-Q_{\max}^*\|_\infty$. (b) Solutions from different initial conditions.}
    \label{fig:1}
\end{figure}

\section{ODE-based stochastic approximation under i.i.d.\ sampling}\label{sec:ODE-stochastic-approximation}

Because of its generality, the ODE method is widely used to analyze convergence of RL algorithms~\citep{bhatnagar2012stochastic,kushner2003stochastic}. It studies a stochastic recursion through the stability of its limiting ODE, using the fact that diminishing-step-size iterates track the ODE asymptotically. A standard version of this approach is based on the Borkar--Meyn theorem~\citep{borkar2000ode}. We summarize the assumptions needed below.
\begin{align}
&\theta_{k+1}=\theta_k+\alpha_k (f(\theta_k)+\varepsilon_{k+1})\label{eq:general-stochastic-recursion}
\end{align}
where $f:{\mathbb R}^n \to {\mathbb R}^n$ is a nonlinear mapping and the integer $k \ge 0$ is the iteration step. Basic technical assumptions are given below.
\begin{framed}
\begin{assumption}\label{assumption:1}
$\,$\begin{enumerate}
\item The mapping $f:{\mathbb R}^n  \to {\mathbb R}^n$ is
globally Lipschitz continuous.

\item There exists a unique globally asymptotically stable equilibrium
$\theta^e\in {\mathbb R}^n$ for the ODE $\dot x_t=f(x_t)$.

\item There exists a function
$f_\infty:{\mathbb R}^n\to {\mathbb R}^n$ such that, for every compact set $K\subset{\mathbb R}^n$,
\[
\lim_{c\to\infty}\sup_{x\in K}\left\|\frac{f(cx)}{c}-f_\infty(x)\right\|_2=0.
\]

\item The origin in ${\mathbb R}^n$ is the unique globally asymptotically stable
equilibrium for the ODE $\dot x_t=f_\infty (x_t)$.

\item The sequence $\{\varepsilon_k,{\mathcal G}_k,k\ge 1\}$ with ${\mathcal G}_k=\sigma(\theta_i,\varepsilon_i,i\le k)$
is a martingale difference sequence. In addition, there exists a constant $C_0<\infty $ such that for any initial $\theta_0\in
{\mathbb R}^n$, we have ${\mathbb E}[\|\varepsilon_{k+1} \|_2^2 |{\mathcal G}_k]\le C_0(1+\|\theta_k\|_2^2),\forall k \ge 0$.

\item The step-sizes satisfy
\begin{align}
&\alpha_k >0,\quad \sum_{k=0}^\infty {\alpha_k}=\infty,\quad \sum_{k=0}^\infty{\alpha_k^2}<\infty.\label{eq:step-size-rule}
\end{align}
\end{enumerate}
\end{assumption}
\end{framed}

\begin{framed}
\begin{lemma}[{\citep[Borkar and Meyn theorem]{borkar2000ode}}]\label{lemma:Borkar}
Under~\cref{assumption:1}, for any initial $\theta_0\in
{\mathbb R}^n$, we have $\sup_{k\ge 0} \|\theta_k\|_2<\infty$
with probability one. In addition, $\theta_k\to\theta^e$ as
$k\to\infty$ with probability one.
\end{lemma}
\end{framed}
The ODE approach in~\cref{lemma:Borkar} has been widely used to prove convergence of RL algorithms, such as the synchronous Q-learning algorithm~\citep{borkar2000ode}, synchronous temporal-difference (TD) learning~\citep{borkar2000ode}, asynchronous Q-learning~\citep{lee2020unified}, gradient TD-learning algorithms~\citep{sutton2009convergent,sutton2009fast,ghiassian2020gradient,lee2022new}, Q-learning with linear function approximation~\citep{melo2008analysis}, and related algorithms~\citep{bhatnagar2012stochastic}.
A related route uses the Robbins--Monro stochastic-approximation framework~\citep{robbins1951stochastic,bhatnagar2012stochastic,borkar2009stochastic}.
The corresponding technical assumptions are given below.
\begin{framed}
\begin{assumption}\label{assumption:2}
$\,$\begin{enumerate}
\setlength{\itemsep}{0pt}
\setlength{\parskip}{0pt}
\item The mapping $f:{\mathbb R}^n  \to {\mathbb R}^n$ is
globally Lipschitz continuous.

\item The ODE $\dot x_t=f(x_t)$ has a closed bounded positively invariant set $\cal H$ that uniformly attracts bounded sets. Specifically, for every bounded set $B\subset {\mathbb R}^n$,
\[
\lim_{t\to\infty}\sup_{x_0\in B}{\rm dist}(x_t,{\cal H})=0,
\]
where $x_t$ is the solution initialized at $x_0$.

\item The iterates defined by~\eqref{eq:general-stochastic-recursion} remain bounded with probability one, i.e.,
\[{\sup _{k \ge 0}}{\left\| {{\theta _k}} \right\|_2} < \infty \]
with probability one.

\item The sequence $\{\varepsilon_k,{\cal G}_k,k\ge 1\}$ with ${\cal G}_k=\sigma(\theta_i,\varepsilon_i,i\le k)$
is a martingale difference sequence. In addition, there exists a constant $C_0<\infty $ such that for any initial $\theta_0\in
{\mathbb R}^n$, we have ${\mathbb E}[\|\varepsilon_{k+1} \|_2^2 |{\cal G}_k]\le C_0(1+\|\theta_k\|_2^2),\forall k \ge 0$ with probability one.

\item The step-sizes satisfy
$\alpha_k>0, \sum_{k=0}^\infty {\alpha_k}=\infty, \sum_{k=0}^\infty{\alpha_k^2}<\infty$.
\end{enumerate}
\end{assumption}
\end{framed}
\begin{framed}
\begin{lemma}[ODE method for set convergence~{\citep[Chapter~2]{borkar2009stochastic}}]\label{lemma:Robbins}
Under~\cref{assumption:2}, for any initial $\theta_0\in
{\mathbb R}^n$, we have
\[
{\rm dist}(\theta_k,{\cal H})\to 0
\]
with probability one.
\end{lemma}
\end{framed}

The set-convergence form of the ODE method follows the same stochastic-approximation principle as the classical Robbins--Monro theorem~\citep{robbins1951stochastic}, but uses attraction of a closed invariant set rather than convergence to a single equilibrium.
The main difference is that~\cref{lemma:Robbins} assumes bounded iterates in~\cref{assumption:2}, whereas~\cref{lemma:Borkar} establishes boundedness without imposing it as a separate assumption. This advantage comes at the cost of stricter conditions on the underlying ODE model in~\cite{borkar2000ode}, including a unique globally stable equilibrium and the existence and stability of the auxiliary ODE $\dot x_t=f_\infty(x_t)$.

\section{Detailed convergence analysis for Q-learning variants}\label{sec:app:convergence-analysis}

This section provides the detailed convergence analysis summarized in~\cref{sec:convergence-summary}. We prove convergence of~\cref{algo:Q-learning}. The update in~\cref{algo:Q-learning} can be expressed as
\begin{align}
&Q_{k+1}=Q_k+\alpha_k (f(Q_k)+\varepsilon_{k+1})\label{algo:Q-learning2}
\end{align}
with the stochastic error
\[{\varepsilon _{k + 1}} = ({e_{{a_k}}} \otimes {e_{{s_k}}})(r({s_k},{a_k},{s_k}') + \gamma  h(Q_k({s_k}', \cdot )) - {Q_k}({s_k},{a_k})) - f({Q_k})\]
and $f({Q_k}): = DR + \gamma DPH({Q_k}) - D{Q_k}$, where $P\in {\mathbb R}^{|{\mathcal S}\times {\mathcal A}| \times |{\mathcal S}|  }$ is the state-action pair to state transition probability matrix defined in~\eqref{eq:definitions1}, $Q_k\in {\mathbb R}^{|{\mathcal S}\times {\mathcal A}|}$, $R \in {\mathbb R}^{|{\mathcal S}\times {\mathcal A}|}$ is an enumeration of the expected reward defined in~\eqref{eq:definitions1}, $D\in {\mathbb R}^{|{\mathcal S}\times {\mathcal A}| \times |{\mathcal S}\times {\mathcal A}|}$ is a diagonal matrix whose diagonal elements are an enumeration of~\eqref{eq:d}, and the order of the enumeration is defined by $d(s,a) = p (s)\beta (a|s)= (e_a  \otimes e_s )^\top D (e_a  \otimes e_s )$. Under~\cref{assumption:positive-distribution}, $D$ is nonsingular and has strictly positive diagonal elements. Equation~\eqref{algo:Q-learning2} is the mean-field decomposition of the asynchronous single-coordinate recursion in~\cref{algo:Q-learning}. Since $\sum_{s,a}d(s,a)=1$, setting $D=I$ does not correspond to a sampling distribution for this recursion when $|{\cal S}\times{\cal A}|>1$. The synchronous full-update recursion is defined separately by updating all coordinates at each iteration; its mean field is $R+\gamma PH(Q)-Q$, which corresponds to $D=I$. The ODE stability results apply to both mean fields, while the calculations below are written for~\cref{algo:Q-learning}.
We first establish convergence of~\cref{algo:Q-learning} for the $\max$, LSE, and mellowmax operators. Since the corresponding Bellman operator is a contraction, the proof follows by applying the Borkar--Meyn theorem~\citep{borkar2000ode}.
\begin{framed}
\begin{theorem}\label{thm:Q-learning-convergence}
Under the i.i.d.\ sampling model specified for~\cref{algo:Q-learning}, assume that the step-sizes satisfy $\alpha_k>0$, $\sum_{k=0}^\infty\alpha_k=\infty$, and $\sum_{k=0}^\infty\alpha_k^2<\infty$. For the LSE, mellowmax, and max operators, the iterates converge with probability one to the corresponding fixed point in~\eqref{eq:4}.
\end{theorem}
\end{framed}
\begin{proof}
It suffices to verify~\cref{assumption:1} and apply the Borkar--Meyn theorem.
Consider the system $\frac{d Q_t}{dt} = f(Q_t)$, where $f(Q) = DF(Q) - DQ$ and $F \in \{F_{\max},F_{\rm lse}^{\lambda}, F_{\rm mm}^{\lambda} \}$.
For the first statement, $f$ is Lipschitz continuous because
\begin{align*}
\left\| f(x) - f(y) \right\|_\infty\le& {\left\| DF(x) - DF(y) \right\|_\infty } + {\left\| D(x - y) \right\|_\infty }\\
\le& {\left\| D \right\|_\infty }{\left\| {F(x) - F(y)} \right\|_\infty } + {\left\| D \right\|_\infty }{\left\| {x - y} \right\|_\infty }\\
\le& \gamma {\left\| D \right\|_\infty }{\left\| {x - y} \right\|_\infty } + {\left\| D \right\|_\infty }{\left\| {x - y} \right\|_\infty}\\
=& (\gamma  + 1){\left\| D \right\|_\infty }{\left\| {x - y} \right\|_\infty }
\end{align*}
where~\cref{lemma:contraction} is used in the third line.
For the second statement, we note that by~\cref{lemma:contraction}, $F$ is a contraction mapping with respect to $\|\cdot \|_\infty$, and by~\cref{thm:stability2}, the fixed point is the unique globally asymptotically stable equilibrium point.
For the third statement, define $f_\infty(Q):=\gamma DPH_{\max}(Q)-DQ$. By~\cref{lemma:4}, the convergence $H(cQ)/c\to H_{\max}(Q)$ is uniform on compact sets; in fact, for the LSE and mellowmax operators,
\[
\sup_{Q\in{\mathbb R}^{|{\cal S}\times{\cal A}|}}
\left\|\frac{H(cQ)}{c}-H_{\max}(Q)\right\|_\infty
\le \frac{\lambda\ln(|{\cal A}|)}{c},
\]
while the difference is zero for the max operator. Therefore, for every compact set $K$,
\begin{align*}
&\sup_{Q\in K}\left\|\frac{f(cQ)}{c}-f_\infty(Q)\right\|_\infty\\
&\quad\le \frac{\|DR\|_\infty}{c}
+\gamma\|D\|_\infty
\sup_{Q\in K}\left\|\frac{H(cQ)}{c}-H_{\max}(Q)\right\|_\infty
\longrightarrow0.
\end{align*}
Since all norms on the finite-dimensional space are equivalent, the same convergence holds in the $2$-norm required by~\cref{assumption:1}.
For the fourth statement, consider
\[\frac{d Q_t}{dt} = f_\infty(Q_t),\]
where $f_\infty(Q) = D \bar F(Q) - DQ$ and $\bar F(Q) = \gamma PH_{\max}(Q)$. The same argument as in~\cref{lemma:contraction} shows that $\bar F$ is a contraction with respect to $\|\cdot \|_\infty$. Moreover, the fixed point is the origin, which is the unique globally asymptotically stable equilibrium point by~\cref{thm:stability2}.

For the fifth statement, define the information $\sigma$-algebra ${\cal G}_k:=\sigma(Q_0,Q_i,\varepsilon_i,1\le i\le k)$ and the process $(M_k)_{k=0}^\infty$ by $M_k:=\sum_{i=1}^k {\varepsilon_i}$. Under the i.i.d.\ sampling model and the definition of $f(Q_k)$ as the conditional mean update, $\mathbb E[\varepsilon_{k+1}\mid{\cal G}_k]=0$. Hence,
\begin{align*}
{\mathbb E}[M_{k+1}|{\cal G}_k]=& {\mathbb E}\left[ \left. \sum_{i=1}^{k+1}{\varepsilon_i} \right|{\cal G}_k\right]={\mathbb E}[\varepsilon_{k+1}|{\cal G}_k]+{\mathbb E}\left[ \left. \sum_{i=1}^k {\varepsilon_i} \right|{\cal G}_k \right]\\
=&{\mathbb E}\left[\left.\sum_{i=1}^k{\varepsilon_i} \right|{\cal G}_k \right]=M_k,
\end{align*}
where the final equality also uses the ${\cal G}_k$-measurability of $M_k$. Therefore, $(M_k)_{k=0}^\infty$ is a martingale, and $\varepsilon_{k+1} = M_{k+1}-M_k$ is a martingale difference. Moreover, for the second-moment bound in item~5 of~\cref{assumption:1}, we have
\begin{align*}
&\mathbb E[\|\varepsilon_{k+1}\|_2^2\mid{\cal G}_k]\\
={}& \mathbb E[\|(e_{a_k}\otimes e_{s_k})(r(s_k,a_k,s_k')
+\gamma h(Q_k(s_k',\cdot))-Q_k(s_k,a_k))-f(Q_k)\|_2^2\mid{\cal G}_k]\\
={}& \mathbb E[\|(e_{a_k}\otimes e_{s_k})(r(s_k,a_k,s_k')
+\gamma h(Q_k(s_k',\cdot))-Q_k(s_k,a_k))\|_2^2\mid{\cal G}_k]\\
&\quad -\mathbb E[\|f(Q_k)\|_2^2\mid{\cal G}_k]\\
\le{}& \mathbb E[\|(e_{a_k}\otimes e_{s_k})(r(s_k,a_k,s_k')
+\gamma h(Q_k(s_k',\cdot))-Q_k(s_k,a_k))\|_2^2\mid{\cal G}_k]\\
={}& \mathbb E[(r(s_k,a_k,s_k')+\gamma h(Q_k(s_k',\cdot))-Q_k(s_k,a_k))^2\mid{\cal G}_k]\\
\le& 3{\mathbb E}[r{({s_k},{a_k},{s_k}')^2}|{\cal G}_k]
+ 3{\gamma ^2} {\mathbb E}[h{({Q_k}({s_k}', \cdot ))^2}|{\cal G}_k]\\
&\quad + 3{\mathbb E}[{Q_k}{({s_k},{a_k})^2}|{\cal G}_k]\\
\le& 3R_{\max }^2
+ 3{\gamma ^2}{\mathbb E}\left[2\left\| {{Q_k}} \right\|_\infty ^2 + C_h\mid {\cal G}_k\right]\\
&\quad + 3{\mathbb E}[\left\| {{Q_k}} \right\|_\infty ^2|{\cal G}_k]\\
\le& 3R_{\max }^2 + 3\gamma^2 C_h\\
&\quad + (6{\gamma ^2} + 3){\mathbb E}[\left\| {{Q_k}} \right\|_2^2|{\cal G}_k]
\end{align*}
where $R_{\max}:=\max _{(s,a,s') \in {\cal S} \times {\cal A} \times {\cal S}} |r (s,a,s')|$. The second inequality uses $\|a+b+c\|_2^2 \leq 3\|a\|_2^2+3\|b\|_2^2+3\|c\|_2^2$ for any $a,b,c \in {\mathbb R}^n$. We also use $|h(q)|\le \|q\|_\infty$ for $h\in\{h_{\max},h_{\rm mm}^{\lambda}\}$, because the mellowmax value lies between the minimum and maximum coordinates of $q$, and $|h_{\rm lse}^{\lambda}(q)|^2\le 2\|q\|_\infty^2+2(\lambda\ln(|{\cal A}|))^2$. Thus, $C_h=0$ for $h\in\{h_{\max},h_{\rm mm}^{\lambda}\}$ and $C_h=2(\lambda\ln(|{\cal A}|))^2$ for $h=h_{\rm lse}^{\lambda}$. The final inequality follows from $\|\cdot \|_\infty \le \| \cdot \|_2$. This completes the proof.
\end{proof}

Next, we consider the Boltzmann softmax operator in~\cref{algo:Q-learning}. Since the Boltzmann softmax operator is, in general, not non-expansive with respect to the $\infty$-norm~\citep{asadi2017alternative}, the induced mapping $F$ may fail to be a contraction in the $\infty$-norm. Consequently, the contraction-based approach used for the $\max$, LSE, and mellowmax operators does not apply directly. Instead, we show that~\cref{algo:Q-learning} with the Boltzmann softmax operator converges to a bounded neighborhood of $Q^*_{\max}$.
\begin{framed}
\begin{theorem}\label{thm:Q-learning-convergence2}
Under the i.i.d.\ sampling model specified for~\cref{algo:Q-learning}, assume that the step-sizes satisfy $\alpha_k>0$, $\sum_{k=0}^\infty\alpha_k=\infty$, and $\sum_{k=0}^\infty\alpha_k^2<\infty$. Set $w_i=1/d_i$, $w_{\min}:=\min_i w_i$, and $w_{\max}:=\max_i w_i$. For the Boltzmann softmax operator, the corresponding iterates satisfy ${\rm dist}(Q_k,{\cal H})\to0$ with probability one, where
\[
{\cal H}:=\left\{Q\in{\mathbb R}^{|{\cal S}\times{\cal A}|}:\|Q-Q_{\max}^*\|_\infty\le
\frac{\gamma\lambda\ln(|{\cal A}|)w_{\max}}{(1-\gamma)w_{\min}}\right\}.
\]
\end{theorem}
\end{framed}
\begin{proof}
It suffices to verify~\cref{assumption:2} and apply~\cref{lemma:Robbins}.
Consider the system $\frac{d Q_t}{dt} = f(Q_t)$, where $f(Q) = DF_{\rm bz}^{\lambda}(Q) - DQ$.
For the first condition, $f$ is Lipschitz continuous because
\begin{align*}
\left\| f(x) - f(y) \right\|_\infty \le& {\left\| DF_{\rm bz}^{\lambda}(x) - DF_{\rm bz}^{\lambda}(y) \right\|_\infty } + {\left\| D(x - y) \right\|_\infty }\\
\le& {\left\| D \right\|_\infty }{\left\| {F_{\rm bz}^{\lambda}(x) - F_{\rm bz}^{\lambda}(y)} \right\|_\infty } + {\left\| D \right\|_\infty }{\left\| {x - y} \right\|_\infty }\\
\le& (\gamma L_{\rm bz}+1){\left\| D \right\|_\infty }{\left\| {x - y} \right\|_\infty },
\end{align*}
where~\cref{lemma:Lipschitz1} is used in the third line. By its definition, ${\cal H}$ is closed and bounded. By~\cref{thm:norm-Q-boltzmann}, it is positively invariant, and the explicit exponential estimate is uniform over every bounded set of initial conditions. Therefore, the second condition in~\cref{assumption:2} is satisfied. By~\cref{lemma:boundedness-bz}, the iterates are bounded for the step-sizes considered here, so the third condition is also satisfied. Finally, the martingale-difference condition and the second-moment bound are proved as in the proof of~\cref{thm:Q-learning-convergence}, using $|h_{\rm bz}^{\lambda}(q)|\le \|q\|_\infty$. Applying~\cref{lemma:Robbins} gives ${\rm dist}(Q_k,{\cal H})\to 0$ with probability one. This completes the proof.
\end{proof}
From~\cref{thm:Q-learning-convergence2}, we see that as $\lambda$ decreases to zero, the limiting error set around the true solution $Q^*_{\max}$ becomes smaller. This is consistent with $h_{\rm bz}^{\lambda}(x)\to h_{\max}(x)$ as $\lambda\downarrow0$.

\section{Examples}
\begin{figure}[!htbp]
    \centering

    \begin{minipage}[b]{0.48\textwidth}
        \centering
        \includegraphics[width=\textwidth, height=4.5cm, keepaspectratio=false]{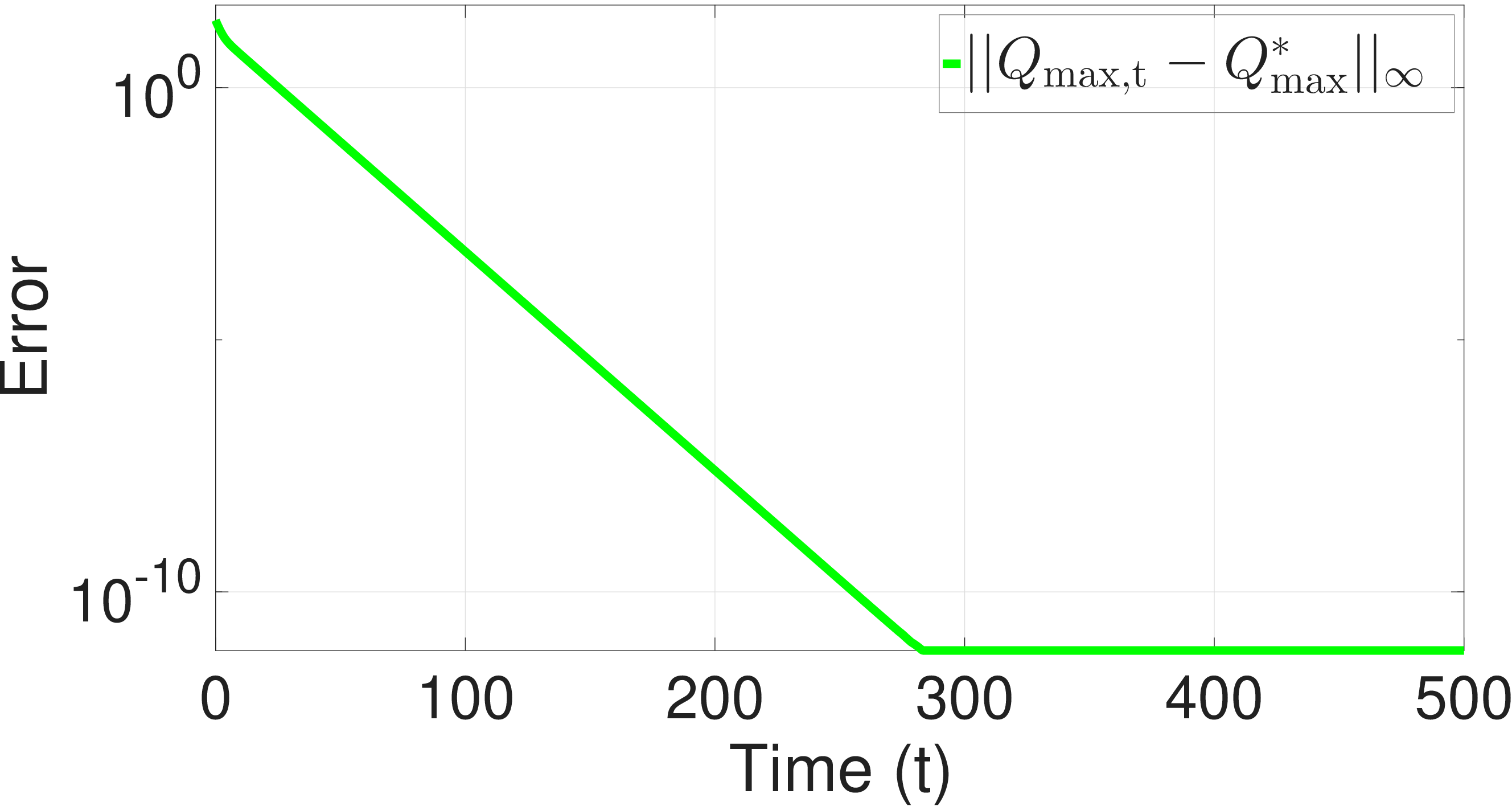}
    \end{minipage}
    \hfill
    \begin{minipage}[b]{0.48\textwidth}
        \centering
        \vspace{-8mm} 
        \includegraphics[width=\textwidth, height=4.5cm, keepaspectratio=true]{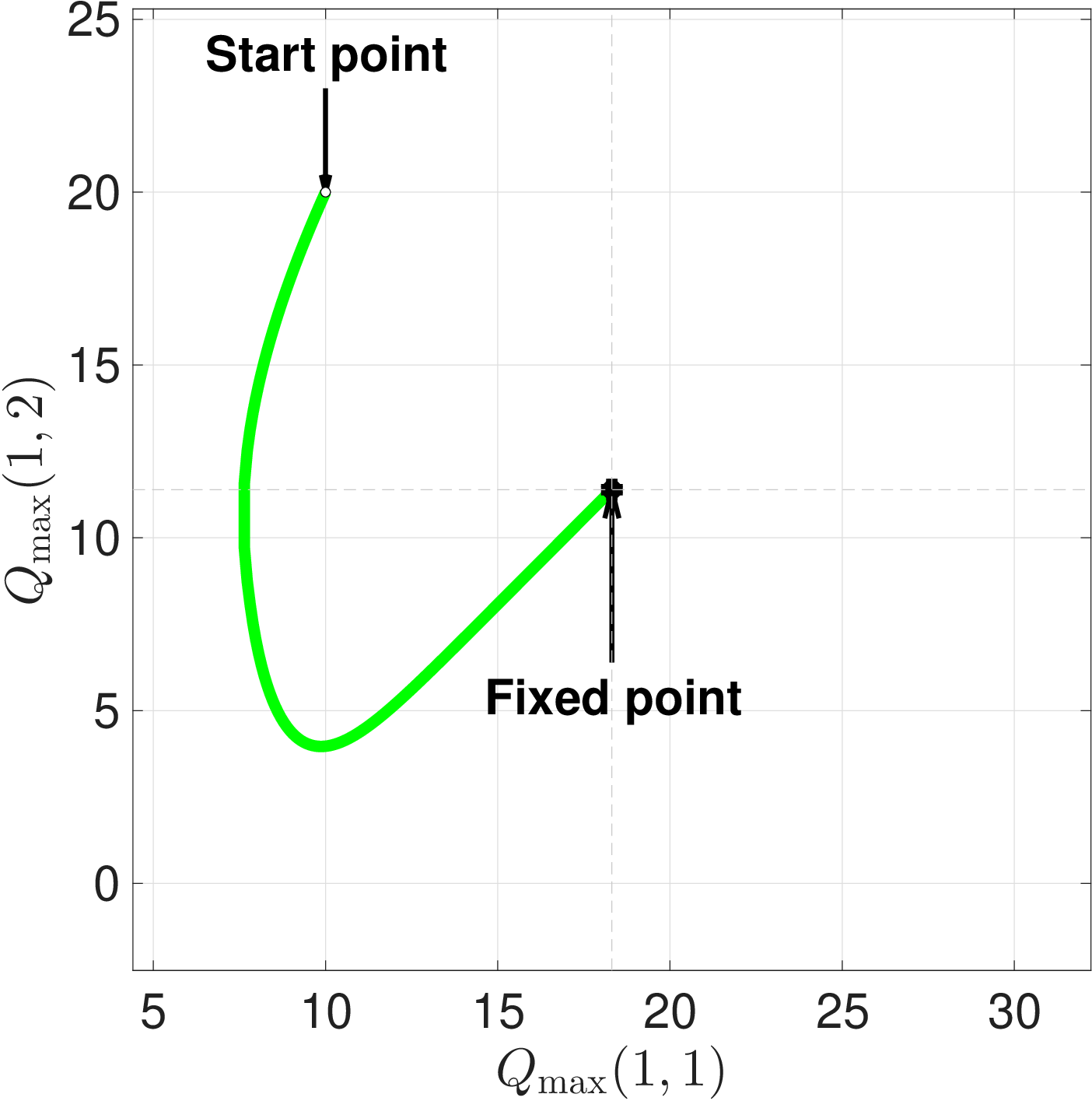}
    \end{minipage}

    \vspace{5mm}

    \begin{minipage}[b]{0.48\textwidth}
        \centering
        \includegraphics[width=\textwidth, height=4.5cm, keepaspectratio=false]{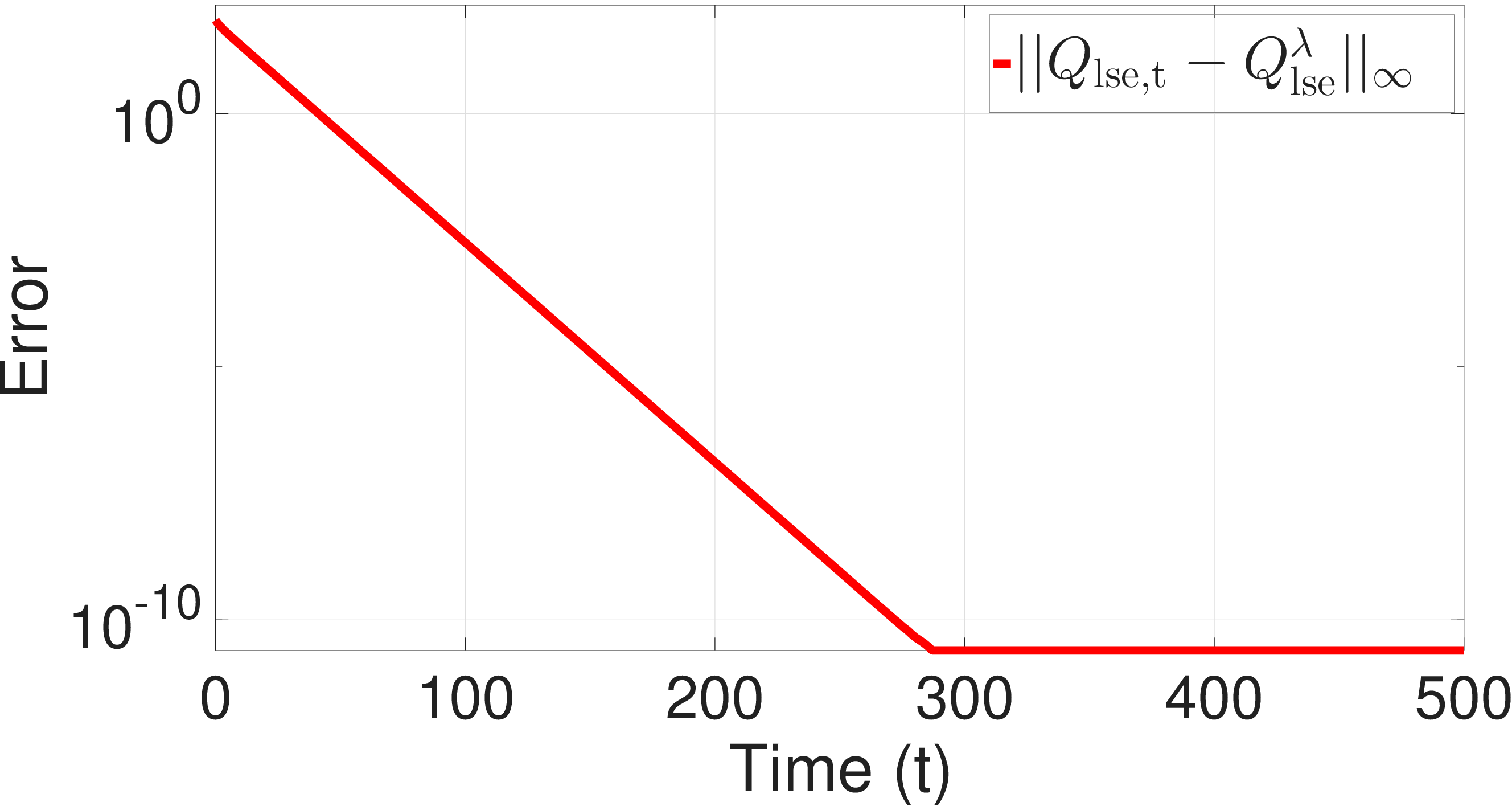}
    \end{minipage}
    \hfill
    \begin{minipage}[b]{0.48\textwidth}
        \centering
        \vspace{-8mm}
        \includegraphics[width=\textwidth, height=4.5cm, keepaspectratio=true]{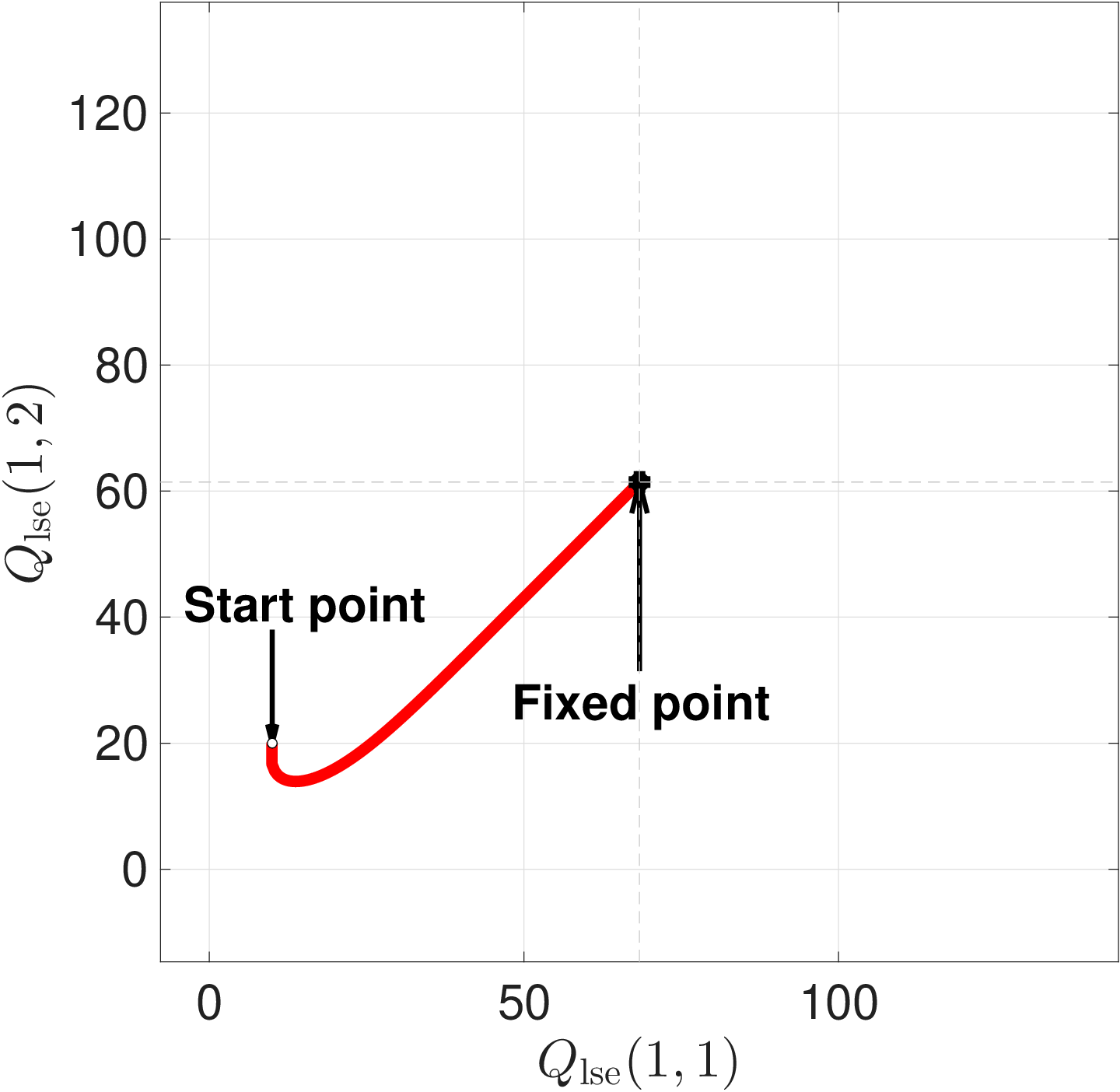}
    \end{minipage}

    \vspace{5mm}

    \begin{minipage}[b]{0.48\textwidth}
        \centering
        \includegraphics[width=\textwidth, height=4.5cm, keepaspectratio=false]{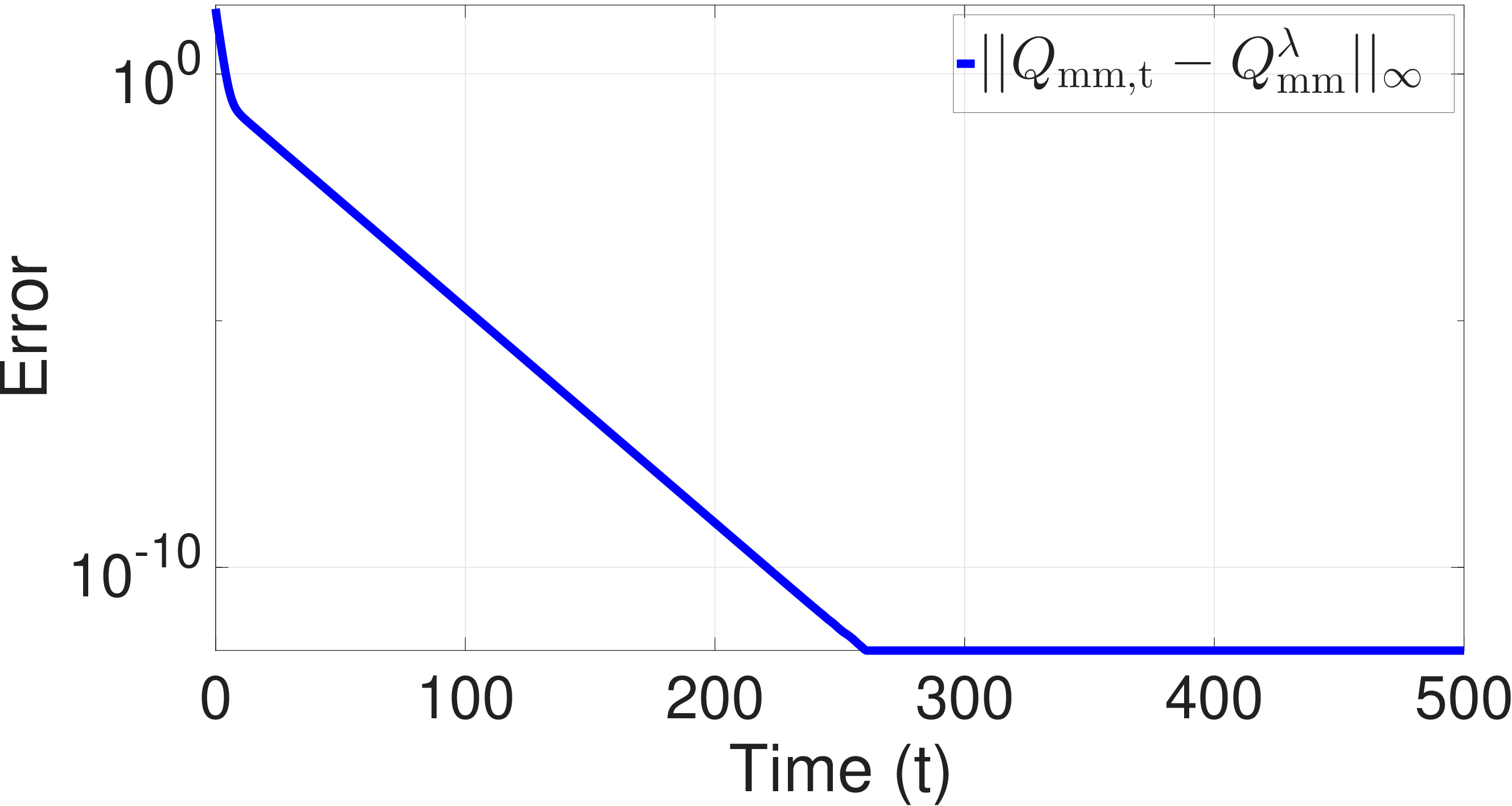}
    \end{minipage}
    \hfill
    \begin{minipage}[b]{0.48\textwidth}
        \centering
        \vspace{-8mm}
        \includegraphics[width=\textwidth, height=4.5cm, keepaspectratio=true]{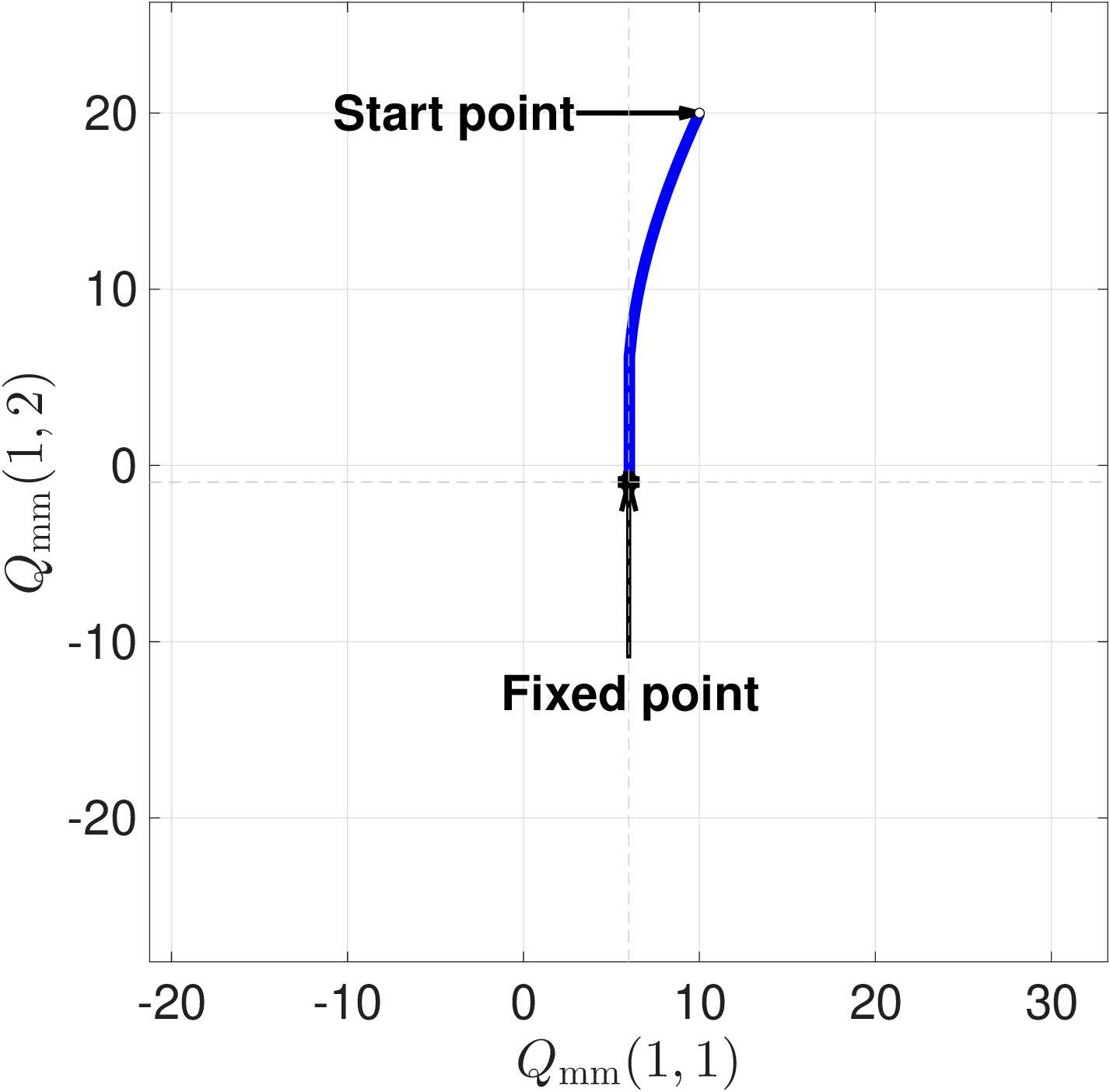}
    \end{minipage}

    \vspace{5mm}

    \begin{minipage}[b]{0.48\textwidth}
        \centering
        \includegraphics[width=\textwidth, height=4.5cm, keepaspectratio=false]{figures/bz_err.eps}
        \vspace{3mm}
        \centerline{\small (a) Error convergence curves}
    \end{minipage}
    \hfill
    \begin{minipage}[b]{0.48\textwidth}
        \centering
        \vspace{-8mm}
        \includegraphics[width=\textwidth, height=4.5cm, keepaspectratio=true]{figures/bz_init_multi_ph.eps}
        \vspace{4mm}
        \centerline{\small (b) Phase plane trajectories}
    \end{minipage}

    \vspace{2mm}
    \caption{Empirical stability analysis under different operators. Panel (a) shows the error curves, and panel (b) shows the phase-plane trajectories toward their respective fixed points.}
    \label{fig:3}
\end{figure}

\begin{figure}[!htbp]
    \centering

    \begin{minipage}[b]{0.48\textwidth}
        \centering
        \includegraphics[width=\textwidth, height=4.5cm, keepaspectratio=false]{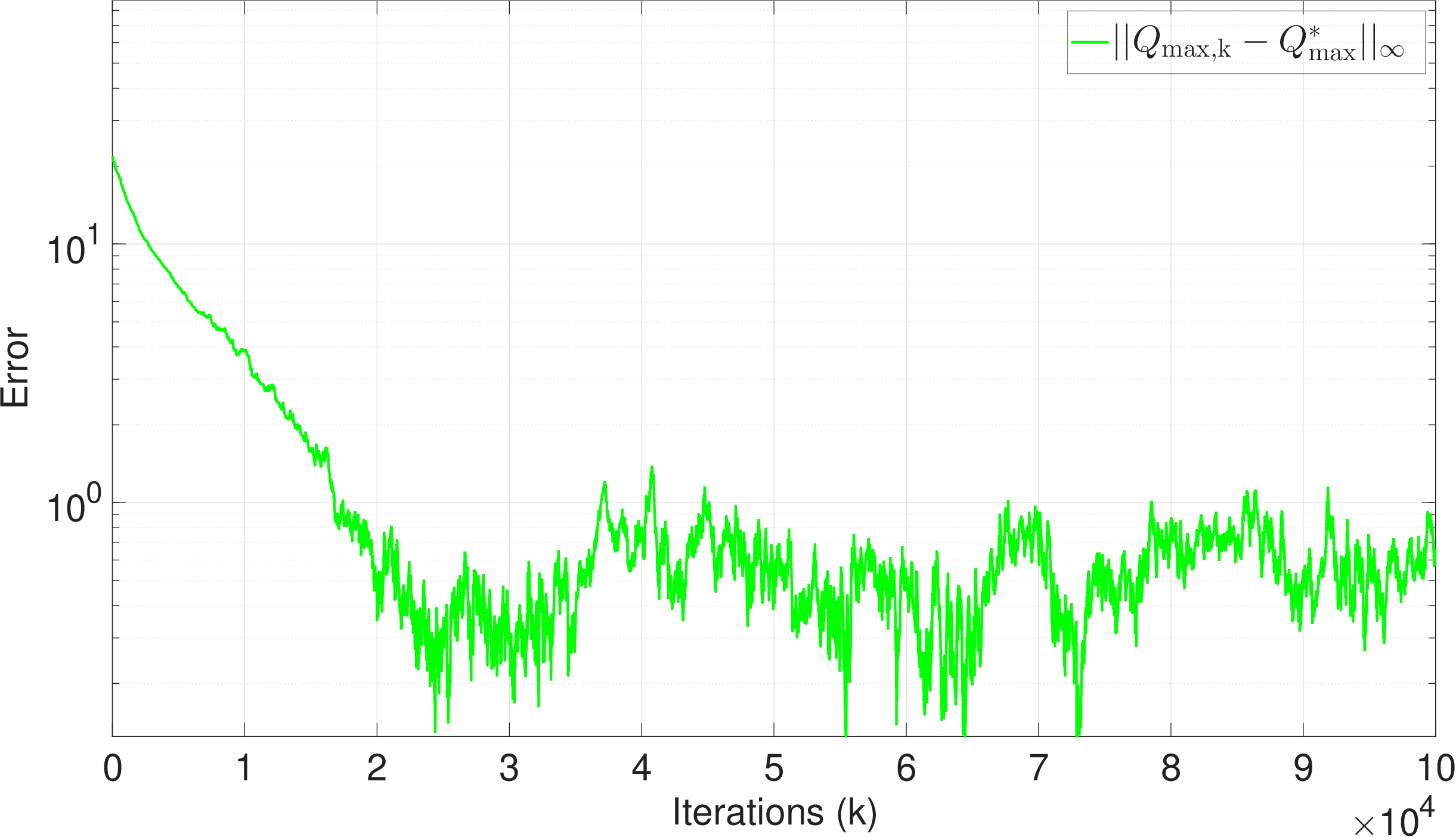}
    \end{minipage}
    \hfill
    \begin{minipage}[b]{0.48\textwidth}
        \centering
        \vspace{-8mm} 
        \includegraphics[width=\textwidth, height=4.5cm, keepaspectratio=true]{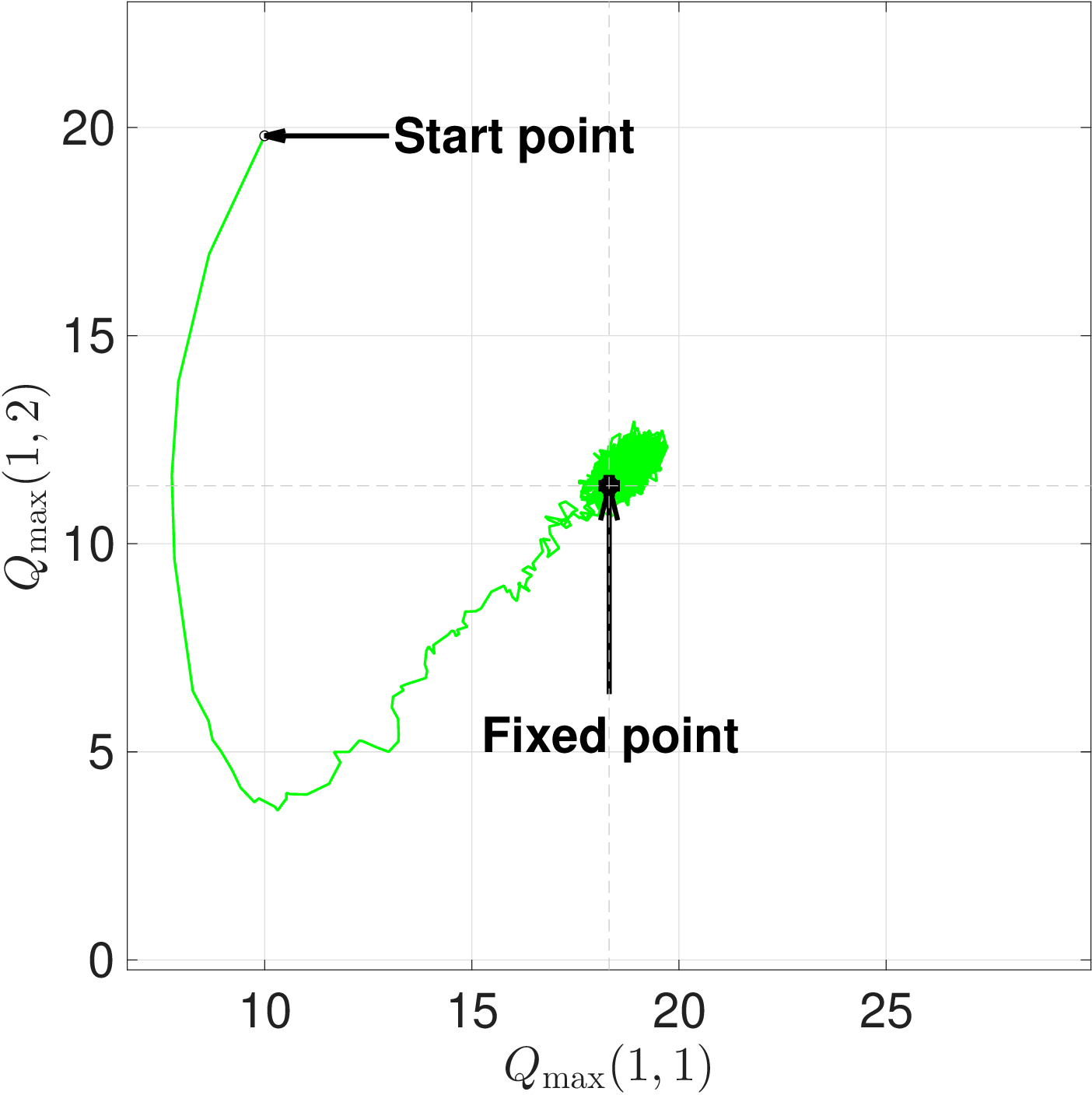}
    \end{minipage}

    \vspace{5mm}

    \begin{minipage}[b]{0.48\textwidth}
        \centering
        \includegraphics[width=\textwidth, height=4.5cm, keepaspectratio=false]{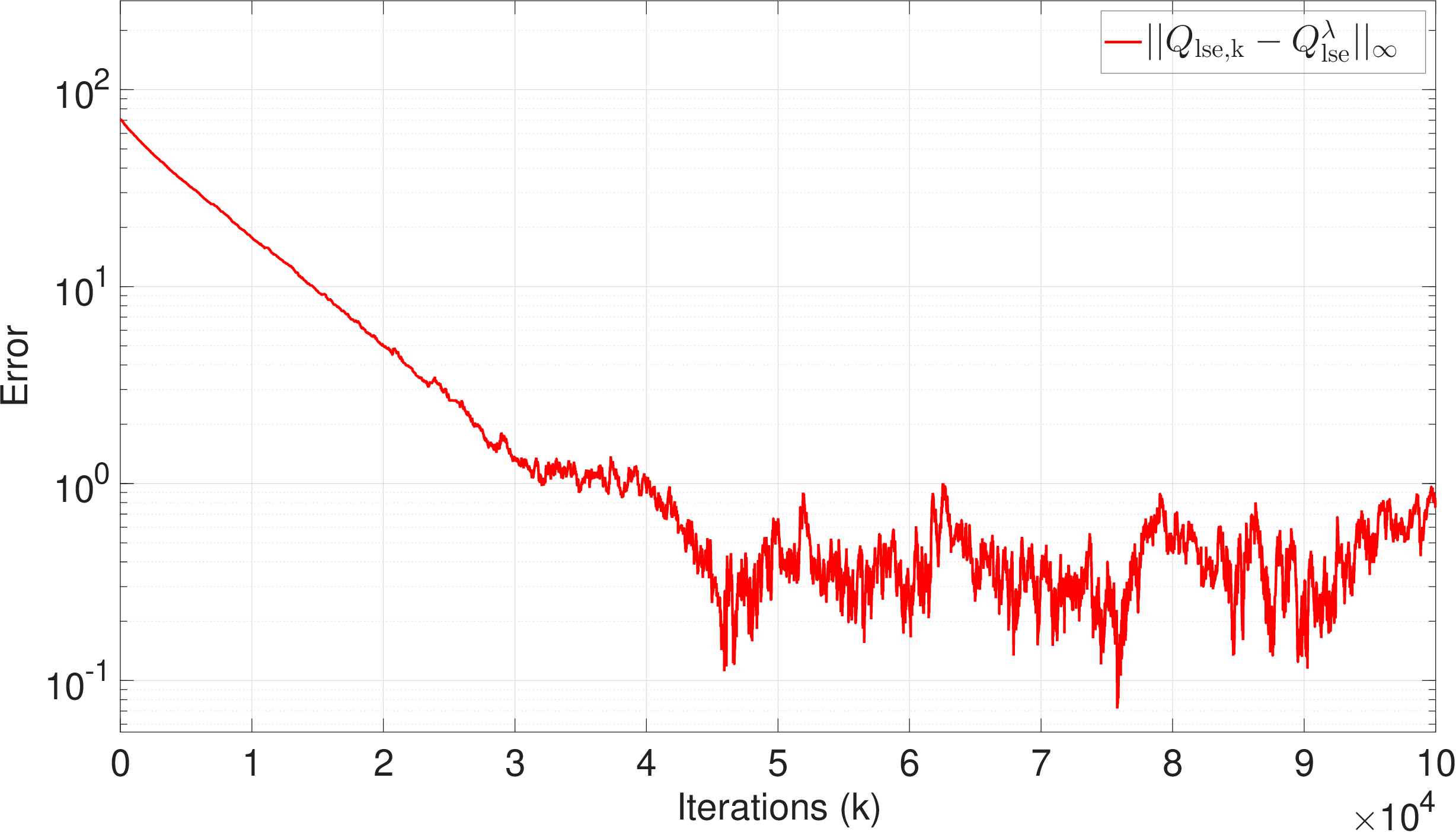}
    \end{minipage}
    \hfill
    \begin{minipage}[b]{0.48\textwidth}
        \centering
        \vspace{-8mm}
        \includegraphics[width=\textwidth, height=4.5cm, keepaspectratio=true]{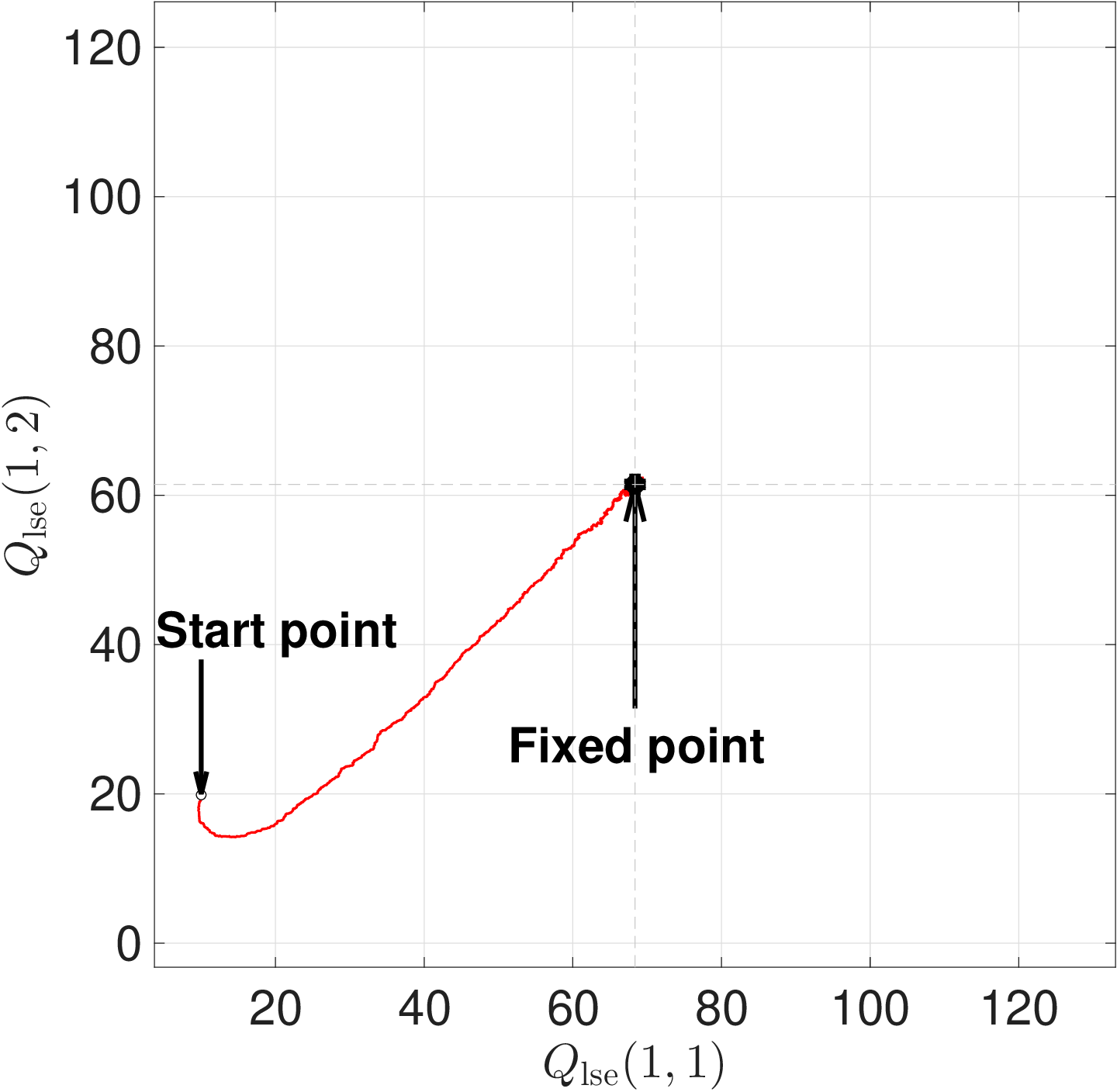}
    \end{minipage}

    \vspace{5mm}

    \begin{minipage}[b]{0.48\textwidth}
        \centering
        \includegraphics[width=\textwidth, height=4.5cm, keepaspectratio=false]{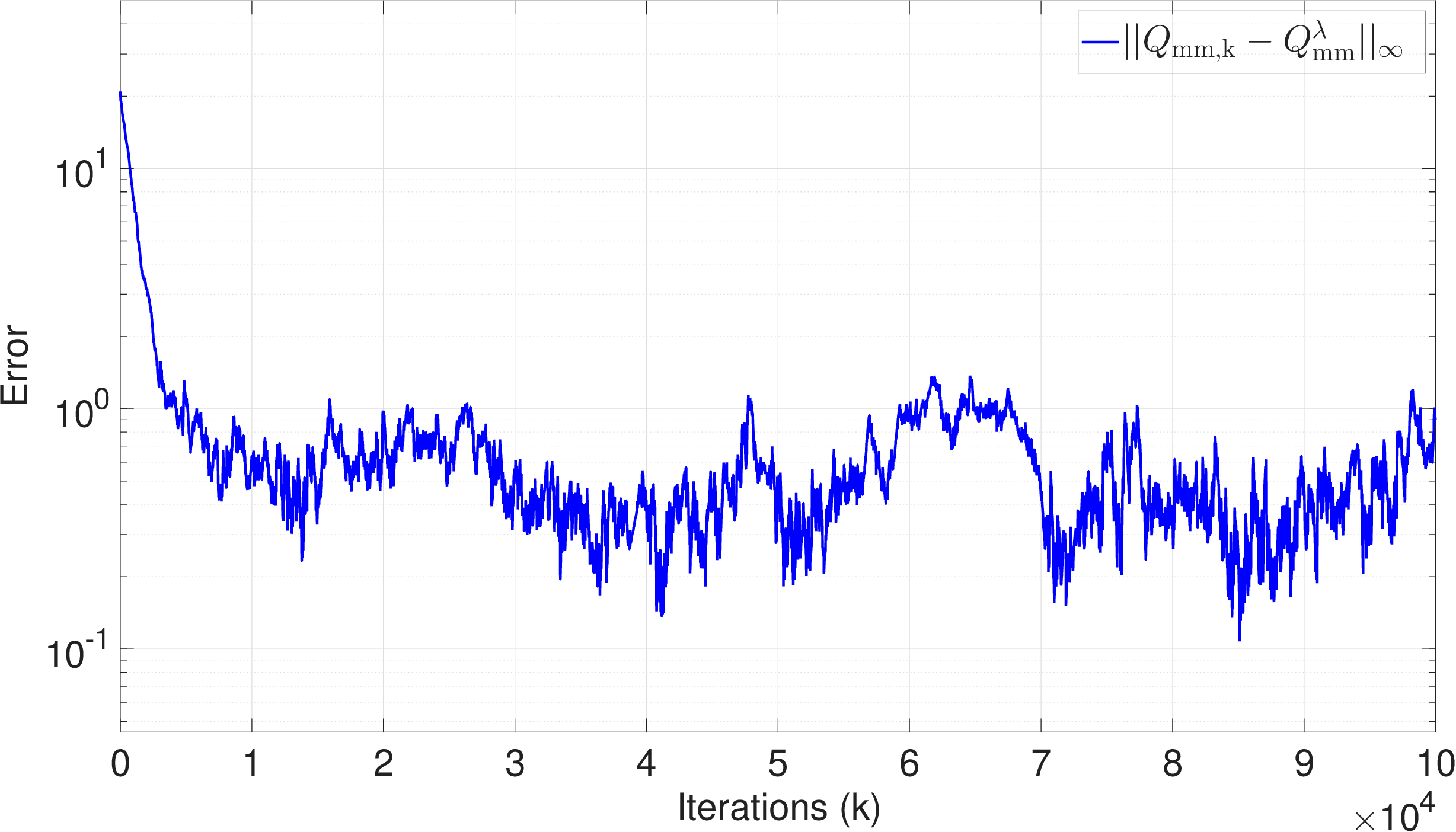}
    \end{minipage}
    \hfill
    \begin{minipage}[b]{0.48\textwidth}
        \centering
        \vspace{-8mm}
        \includegraphics[width=\textwidth, height=4.5cm, keepaspectratio=true]{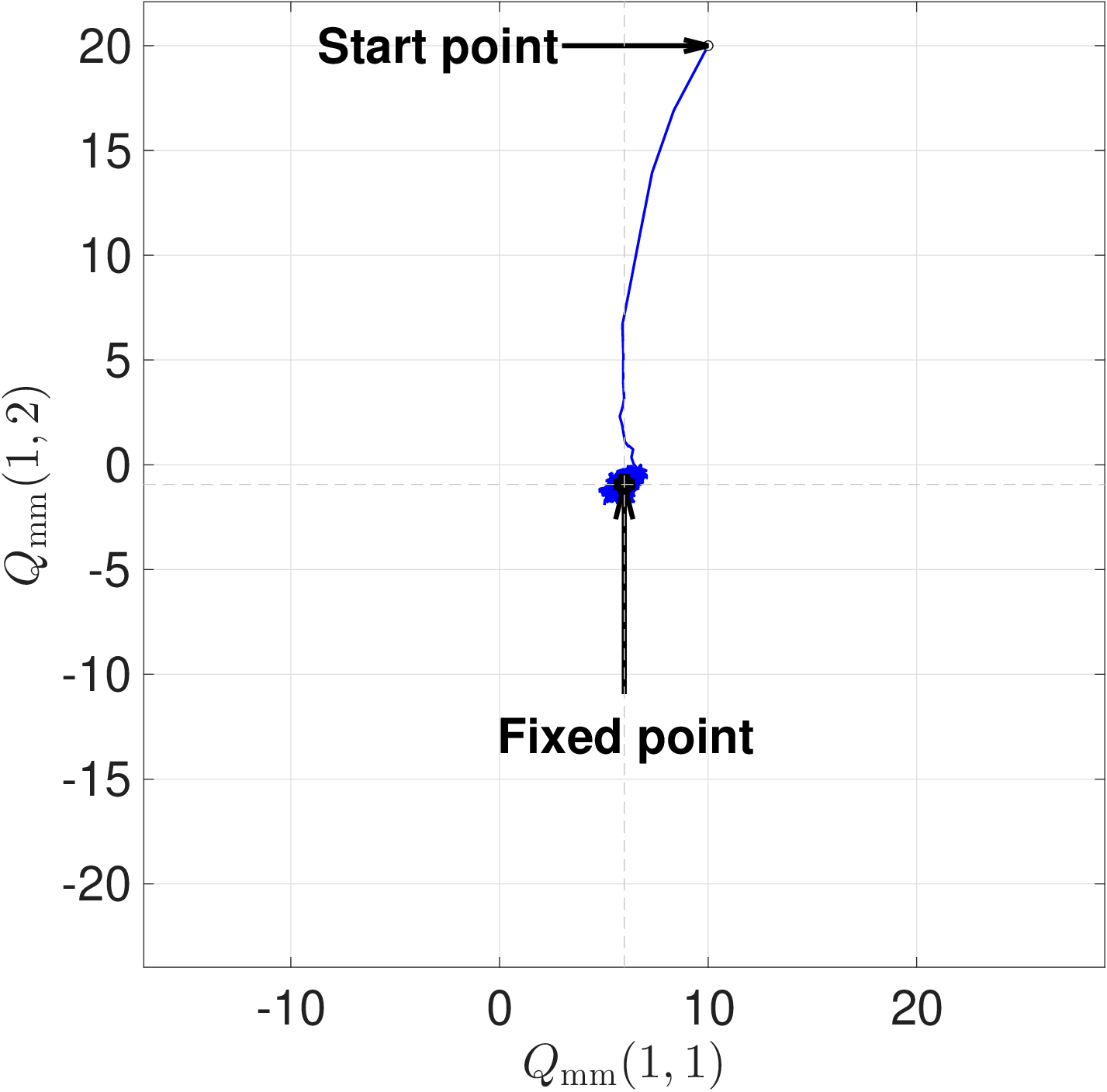}
    \end{minipage}

    \vspace{5mm}

    \begin{minipage}[b]{0.48\textwidth}
        \centering
        \includegraphics[width=\textwidth, height=4.5cm, keepaspectratio=false]{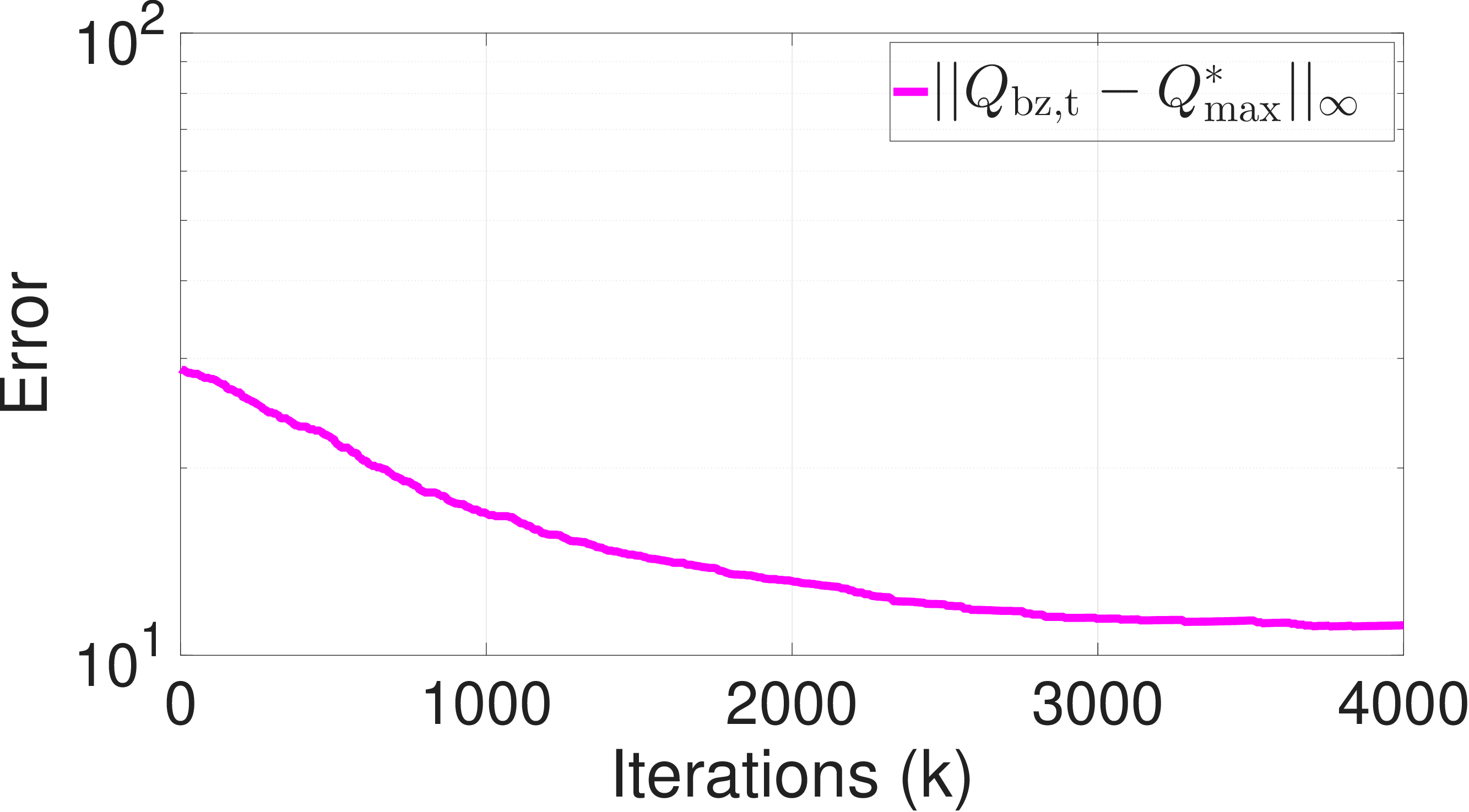}
        \vspace{3mm}
        \centerline{\small (a) Error convergence curves}
    \end{minipage}
    \hfill
    \begin{minipage}[b]{0.48\textwidth}
        \centering
        \vspace{-8mm}
        \includegraphics[width=\textwidth, height=4.5cm, keepaspectratio=true]{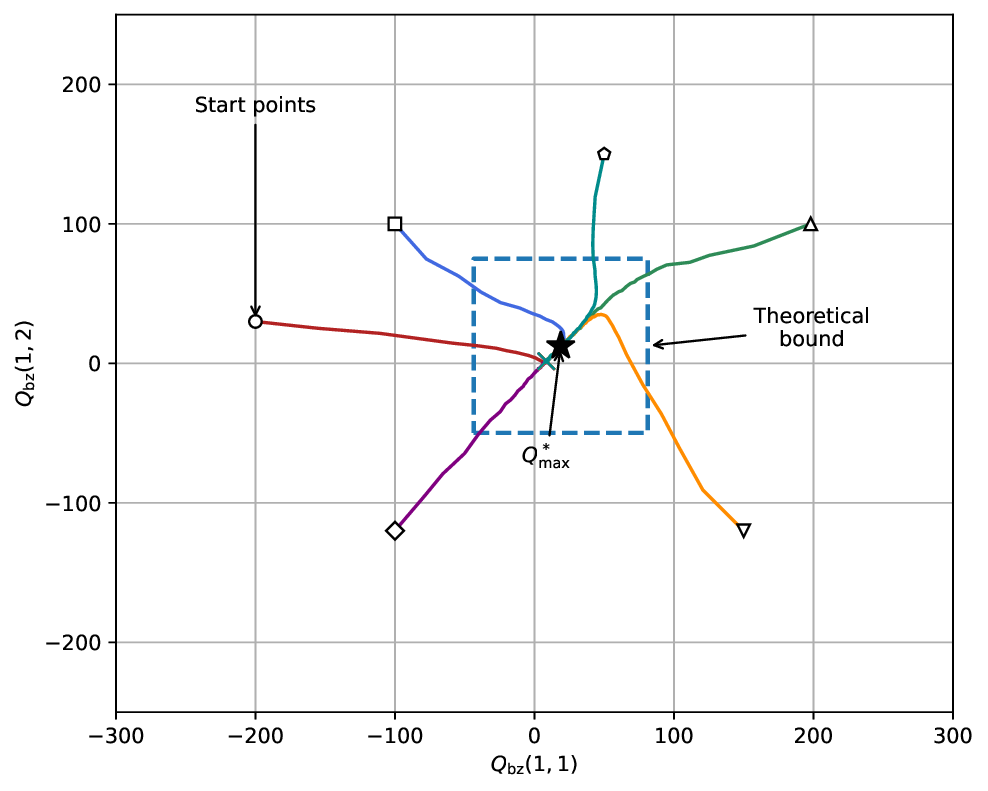}
        \vspace{4mm}
        \centerline{\small (b) Phase plane trajectories}
    \end{minipage}

    \vspace{2mm}
    \caption{Empirical stability analysis under different operators in RL settings. Panel (a) shows the error curves, and panel (b) shows the phase-plane trajectories toward their respective fixed points.}
    \label{fig:4}
\end{figure}

In this section, we present simulation results corresponding to~\cref{fig:3} and~\cref{fig:4}. We consider a simple MDP with $|{\cal S}|=4$, $|{\cal A}|=2$, $\gamma = 0.9$, $R = [0, 5, 0, -5, 0, 5, 0, -5]^\top$, and
\[P_{1} = \begin{bmatrix} 0.1 & 0.8 & 0.0 & 0.1 \\ 0.1 & 0.1 & 0.8 & 0.0 \\ 0.0 & 0.1 & 0.1 & 0.8 \\ 0.8 & 0.0 & 0.1 & 0.1 \end{bmatrix}, \quad
P_{2} = \begin{bmatrix} 0.1 & 0.1 & 0.0 & 0.8 \\ 0.8 & 0.1 & 0.1 & 0.0 \\ 0.0 & 0.8 & 0.1 & 0.1 \\ 0.1 & 0.0 & 0.8 & 0.1 \end{bmatrix},\]
where $P_1$ and $P_2$ represent the state transition probability matrices under actions $a=1$ and $a=2$, respectively.
The temperature parameter for the LSE, mellowmax, and Boltzmann operators is set to $\lambda=10$. For the continuous-time simulations, we use the synchronous mean field with $D=I$.

The continuous-time ODE results are shown in~\cref{fig:3}, whose first three rows correspond to the standard max, LSE, and mellowmax operators. In each error plot, $\|Q_t-Q_e\|_{\infty}$ decays approximately linearly on a logarithmic scale, consistent with the exponential convergence predicted by~\cref{thm:stability5}. The phase portraits in the right column use a common initial point and illustrate convergence toward the respective fixed points.
We note that the trajectory plots are obtained by projecting the dynamics onto a two-dimensional plane for visualization, since the true $Q_t$ is eight-dimensional.
In the fourth row, we focus on the Boltzmann operator. Trajectories from different initial points are shown to converge toward a ball $\mathcal{H}$ around $Q_{\max}^*$, which is obtained theoretically in~\cref{thm:norm-Q-boltzmann}. The dashed square shown in the figure is the two-dimensional coordinate-plane cross-section of the actual set $\cal H$.
Regardless of the initial points, the trajectories consistently remain around the theoretical error ball $\mathcal{H}$ defined in~\cref{thm:norm-Q-boltzmann}.

\Cref{fig:4} presents the results for the corresponding RL versions with a constant step-size $\alpha = 0.01$ over $100,000$ iterations. The simulation follows a single Markov trajectory, whereas the convergence theorems assume i.i.d.\ generative-model samples. In addition, the constant step-size does not satisfy the diminishing step-size condition in~\eqref{eq:step-size-rule}. Therefore, these simulations are illustrative and are not direct numerical instances of~\cref{thm:Q-learning-convergence,thm:Q-learning-convergence2}. Over the simulated horizon, the first three rows (max, LSE, and mellowmax) show trajectories approaching neighborhoods of their respective fixed points.
We note that the trajectory plots are obtained by projecting the dynamics onto a two-dimensional plane for visualization, since the true $Q_k$ is eight-dimensional.
In the fourth row, the Boltzmann operator exhibits an initial decay followed by a persistent residual. This behavior is qualitatively consistent with convergence toward a neighborhood of $Q_{\max}^*$, but the constant-step experiment does not establish the probability-one set-convergence conclusion of~\cref{thm:Q-learning-convergence2}. The dashed square shown in the figure is the two-dimensional coordinate-plane cross-section of the actual set $\cal H$.
These observations provide qualitative illustrations of the ODE-based stability mechanisms in discrete-time stochastic implementations.

\newpage

\end{document}